%% file: main.tex
\documentclass{article}

\usepackage{iclr2027_conference,times}

\usepackage[utf8]{inputenc}
\usepackage[T1]{fontenc}
\usepackage{hyperref}
\usepackage{url}
\usepackage{booktabs}
\usepackage{amsfonts}
\usepackage{amsmath}
\usepackage{nicefrac}
\usepackage{microtype}
\usepackage{xcolor}
\usepackage{graphicx}
\usepackage{enumitem}
\usepackage{caption}
\usepackage{subcaption}
\usepackage{tabularx}
\usepackage{longtable}
\usepackage{makecell}
\usepackage{amssymb}
\usepackage{float}
\usepackage{placeins}
\usepackage{colortbl}
\usepackage{multirow}
\input{table_style}
\usepackage{arxiv_preprint}

\renewcommand{\topfraction}{0.88}
\renewcommand{\textfraction}{0.12}
\renewcommand{\floatpagefraction}{0.7}

\definecolor{latentblue}{HTML}{E6F2FF}

\title{Population-Scalable Multi-Agent World \\
Modeling across Scene Geometries}

\author{
\textbf{Renjie Zhao}\textsuperscript{1,3,$\ast$}\quad
\textbf{Yuxiang Wu}\textsuperscript{1,3,$\ast$}\quad
\textbf{Mingyu Zhang}\textsuperscript{1,3,4,$\diamond$}\quad
\textbf{Jiaxin Li}\textsuperscript{1,3,4,$\diamond$}
\\[4pt]
\textbf{Sisi Li}\textsuperscript{2}\quad
\textbf{He Li}\textsuperscript{2}\quad
\textbf{Yimin Sheng}\textsuperscript{2}\quad
\textbf{Tianxi Tan}\textsuperscript{1,3,4}
\\[4pt]
\textbf{Zhenkai Zhang}\textsuperscript{3}\quad
\textbf{Jiao Liang}\textsuperscript{2}\quad
\textbf{Jianyi Zhu}\textsuperscript{2}\quad
\textbf{Yong-Lu Li}\textsuperscript{1,3,4,$\dagger$}
\\[7pt]
\textsuperscript{1}RhOS.ai\qquad
\textsuperscript{2}Ophilus.AI
\\[3pt]
\textsuperscript{3}Shanghai Jiao Tong University\qquad
\textsuperscript{4}Shanghai Innovation Institute
\\[5pt]
\textsuperscript{$\ast$}Equal contribution.\quad
\textsuperscript{$\diamond$}Project leaders.\quad
\textsuperscript{$\dagger$}Corresponding author.
\\[4pt]
\href{mailto:mingyuzhang@rhos.ai}{\texttt{mingyuzhang@rhos.ai}}\qquad
\href{mailto:yonglu_li@rhos.ai}{\texttt{yonglu\_li@rhos.ai}}
\\[5pt]
Project page: \url{https://rhos.ai/research/khora}
\\[2pt]
Online demo: \url{https://ophilus.ai/khora}
}

\hypersetup{
  hidelinks,
  pdftitle={Population-Scalable Multi-Agent World Modeling across Scene Geometries},
  pdfauthor={Renjie Zhao, Yuxiang Wu, Mingyu Zhang, Jiaxin Li, Sisi Li, He Li, Yimin Sheng, Tianxi Tan, Zhenkai Zhang, Jiao Liang, Jianyi Zhu, Yong-Lu Li},
  pdfsubject={Public preprint},
  pdfkeywords={world models, multi-agent generation, shared world state, mesh-guided generation}
}

\begin{document}
\raggedbottom

\maketitle

\input{sec/0_abstract}

\input{sec/1_intro}

\input{sec/2_related}

\input{sec/3_preliminary}

\input{sec/4_method}

\input{sec/5_exp}

\input{sec/6_conclusion}


\FloatBarrier
\bibliographystyle{plainnat}
\bibliography{references}

\newpage
\appendix
\input{sec/appendix}

\end{document}

%% file: sec/0_abstract.tex
\begin{abstract}

World models have recently achieved impressive progress in visual prediction and interactive generation, yet existing multi-agent systems are typically tied to predefined agent configurations and scene-specific environments. We propose \textbf{Khora}, a scalable multi-agent world model that supports both inference-time population expansion and generalization across geometrically distinct environments. Khora decouples shared world-state evolution from visual rendering: an action-conditioned state model maintains dynamic agents in a persistent Spatio-temporal Board (\emph{STBoard}), while each observation is generated independently through a projection-based rendering interface. The renderer is trained with one target view per sample and reused across agents at inference without retraining. Crucially, scene geometry is provided to the renderer as an explicit mesh condition, allowing the same trained model to render observations that conform to different input environments without changing the underlying world dynamics or model architecture. This design enables Khora to scale along two complementary dimensions: many agents can interact within the same evolving world, while the same model can render observations across many different mesh-defined environments. Experiments demonstrate multi-agent consistency, action-conditioned interaction, generalization to unseen meshes, and scalability to large agent populations.

\end{abstract}

\begin{figure}[h]
  \centering
   \includegraphics[width=\linewidth,trim=0 58bp 0 8bp,clip]{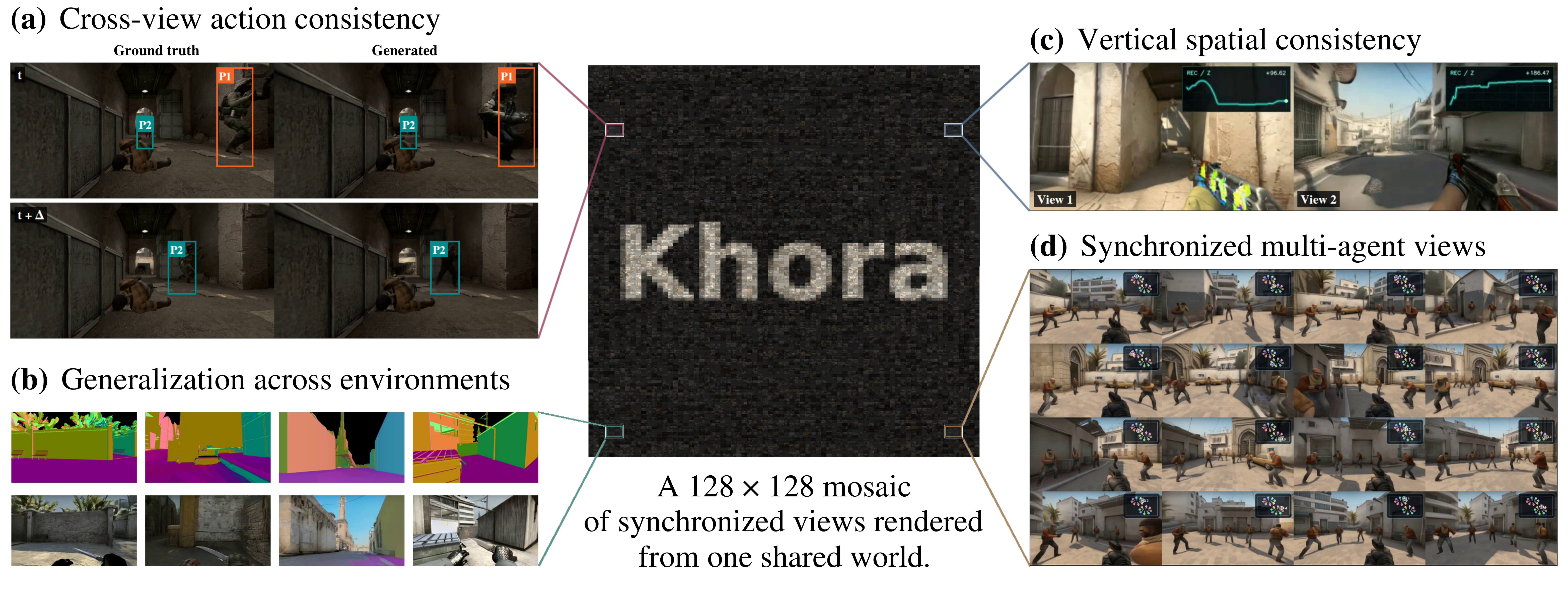}
  \caption{\textbf{Khora} renders observations from a shared world state. Its renderer, trained with one target view per sample, supports additional agents without retraining and follows mesh-defined scene layouts.}
  \label{fig:rhos}
\end{figure}

%% file: sec/1_intro.tex
\section{Introduction}

\input{sec/figures_2_and_3}

Imagine a multiplayer battlefield where players act simultaneously and observe the same evolving world from different perspectives. A world model must do more than generate plausible egocentric videos: it must maintain a coherent, continuously evolving shared state consistent across agents, viewpoints, and actions. Recent advances in action-conditioned generation \citep{hafner2023dreamerv3,wang2023drivedreamerrealworlddrivenworldmodels,brooks2024genie} and multi-agent modeling \citep{savva2026solaris,wu2026multiworld,hu2026mira} highlight these requirements. Yet a scalable simulator must allow agents to enter and leave at inference time without retraining or changing the model interface.

Existing approaches enforce consistency through cross-agent interaction within video generators \citep{savva2026solaris,liu2026gammaworld,hu2026mira,hu2026metaworld} or shared representations and external memories \citep{po2026multigen,odyssey2026agora1}. Predefined view configurations can limit expansion, while direct view coupling places interaction costs inside the neural generator. \textbf{The governing rules of a physical world are population agnostic}: interactions depend on the entities present, but transition rules need not change with their number. We therefore treat population expansion as an inference-time operation on shared world state, rather than a change to agent-specific model components.

We propose \textbf{Khora}, a scalable multi-agent world model that decouples world-state evolution from visual rendering. Initialized from initial observations and a coarse scene prior, its action-conditioned model predicts world state and updates a persistent \emph{STBoard} containing scene memory and dynamic entity states. Khora transforms relevant entities into the target camera coordinate frame and rasterizes them into a fixed-dimensional spatial condition. The renderer is trained with one target view per sample and reuses the same weights across views at inference; additional agents require only new state entries and rendering queries, without changing the renderer architecture or retraining.

This separation also supports mesh-guided scene expansion: a supplied mesh specifies scene geometry, allowing the same trained model to generate imagery following new spatial layouts. We demonstrate this capability with externally authored meshes.

Controlled qualitative evaluations and an interactive system prototype illustrate cross-view action consistency, persistence under partial occlusion, coherent rendering from synchronized viewpoints, and operation with agent populations not observed during training.

Our contributions are summarized as follows:

\begin{enumerate}[leftmargin=1.5em]
  \item We formulate \textbf{inference-time population scalability} as an objective for multi-agent world models, requiring support for changing agent populations without architectural modification or retraining.
  \item We introduce Khora, a scalable world model architecture that decouples world-state evolution from visual rendering. It combines a persistent shared STBoard, an action-conditioned pose and state transition model, and a projection-based per-view rendering interface.
  \item We demonstrate that Khora directly supports multi-view generation at inference time through additional view queries, without architectural modification or retraining. It also maintains visual quality, multi-view consistency and action consistency. It achieves scalable real-time simulation with dominant rendering cost scaling linearly with the number of agents.
\end{enumerate}

\newcommand{\yes}{\(\checkmark\)}
\newcommand{\partialcap}{\(\triangle\)}
\newcommand{\no}{--}

\begin{table}[t]
\centering
\caption{
\textbf{Comparison of representative multi-agent video world models.}
Prior methods are ordered by their first arXiv submission date.
\(\checkmark\) indicates explicit support, \(\triangle\) partial or limited support,
and -- indicates that the capability is not explicitly established.
The complexity shown for MIRA is inferred from full spatial attention over tiled views,
not measured scaling beyond its four-player configuration.
}
\label{tab:capability_comparison}
\small
\setlength{\tabcolsep}{5pt}
\renewcommand{\arraystretch}{1.15}

\begin{tabularx}{\linewidth}{Lcccc}
\toprule
\tableheadleft{Method} & \tablehead{Persistent\\shared state} & \tablehead{Dynamic\\population} & \tablehead{Unseen-count\\generalization} & \tablehead{Population-time\\scaling} \\
\midrule

Solaris~\citep{savva2026solaris} & \no & \no & \no & \(O(N^2)\) \\

\rowcolor{tableStripe}
MultiGen~\citep{po2026multigen} & \yes & \partialcap & \no & \(O(N)\) \\

MultiWorld~\citep{wu2026multiworld} & \partialcap & \yes & \partialcap & \(O(N^2)\) \\

\rowcolor{tableStripe}
Gamma-World~\citep{liu2026gammaworld} & \no & \yes & \partialcap & \(O(N)\) \\

MIRA~\citep{hu2026mira} & \partialcap & \no & \no & \(O(N^2)\) \\

\rowcolor{latentblue}
\textbf{Khora (ours)} & \yes & \yes & \yes & \(\mathbf{\approx O(N)}\) \\

\bottomrule
\end{tabularx}
\end{table}

%% file: sec/figures_2_and_3.tex
\begin{figure}[t]
  \centering
  \includegraphics[width=\linewidth]{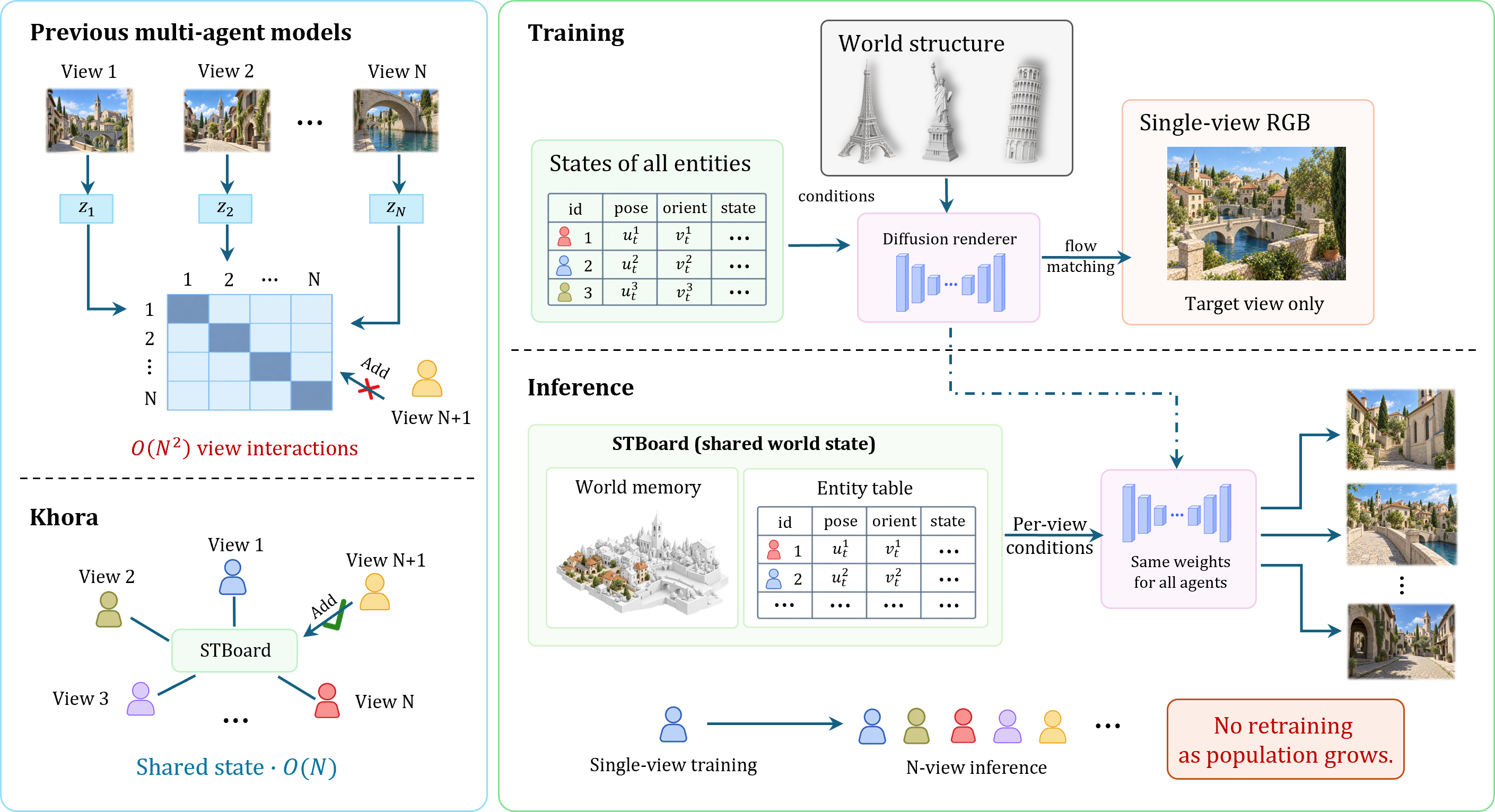}
  \caption{\textbf{Pairwise view coupling versus shared-state rendering.} Khora projects a persistent STBoard (Spatio-temporal Board) into view-specific conditions and reuses a shared renderer trained with one target view per sample for multi-view inference without retraining. New meshes provide geometry priors at inference.}
  \label{fig:conceptual_comparison}
\end{figure}

%% file: sec/2_related.tex
\section{Related Works}

\paragraph{World models and interactive generation.}
World models predict environment dynamics for planning and control \citep{ha2018worldmodels,hafner2023dreamerv3}. Diffusion and transformer generators \citep{ho2020ddpm,rombach2022latent,dosovitskiy2021vit,peebles2023dit} underpin action-conditioned video simulation in Genie, DIAMOND, AdaWorld, and Vid2World \citep{brooks2024genie,alonso2024diamond,gao2025adaworld,huang2025vid2world}.

\paragraph{Multi-agent video world models.}
Solaris \citep{savva2026solaris} couples two-player views through cross-view attention; MultiGen \citep{po2026multigen} maintains external memory for editable multiplayer worlds. MultiWorld \citep{wu2026multiworld} uses a global state encoder, while Gamma-World \citep{liu2026gammaworld} uses permutation-symmetric encodings and sparse hub attention. MIRA \citep{hu2026mira} predicts four fixed-order views with tiled-latent attention and rolling history. MetaWorld \citep{hu2026metaworld} derives multi-agent supervision from monocular videos and couples views through inter-branch attention. Khora trains its renderer with one target view per sample and supports population changes through fixed-dimensional queries of evolving shared state (Table~\ref{tab:capability_comparison}).

\paragraph{Persistent scene memory.}
WorldMem \citep{xiao2025worldmem} stores frames with poses and timestamps. Context-as-Memory \citep{yu2025contextmemory} retrieves frames by field-of-view overlap; VMem \citep{li2025vmem} indexes views with surfels. Spatia \citep{zhao2025spatia} updates a persistent point cloud through visual SLAM. WorldPlay \citep{sun2025worldplay} reconstructs historical context, while RELIC \citep{hong2025relic} retains compressed, camera-aware latent memory for real-time generation. Khora combines persistent appearance memory with an action-updated entity table and fixed-dimensional rendering queries, targeting inference-time population changes.

\paragraph{Geometry-guided image and scene generation.}
GQN \citep{eslami2018gqn} separates scene representation from viewpoint-conditioned generation. ControlNet \citep{zhang2023controlnet} provides spatial conditioning, and SceneCraft \citep{hu2024scenecraft} projects semantic layouts into conditioning maps. GEN3C \citep{ren2025gen3c} renders a growing 3D cache into target-view conditions; WorldMesh \citep{schneider2026worldmesh} and DAR \citep{chen2026dar} use mesh-derived conditions for image and video synthesis. Khora uses a fixed mesh scaffold with updated appearance and entity states to condition independently queried views; its unseen-mesh examples evaluate layout-conditioned generation.

\paragraph{Evaluation of visual world models.}
Image and video quality metrics \citep{wang2004ssim,zhang2018lpips,heusel2018fid,unterthiner2019fvd} do not directly establish multi-agent consistency. We complement them with human ratings, GT-referenced actor metrics, and geometric-and-color scene proxies related to MetaWorld \citep{hu2026metaworld}, distinguishing reference fidelity from agreement between generated views.

%% file: sec/3_preliminary.tex
\section{Preliminaries}




Consider a world with a variable set of agents $I_t$. At inference time, the model receives a coarse static world prior $G$, represented by a scene mesh, and a collection of initial observations $O_0=\{o_0^i\}_{i\in I_0}$. At each subsequent time step $t$, the model receives a set of actions $A_t=\{a_t^i\}_{i\in I_t}$. The initial observations and static prior are used to initialize the world memory $M$ and the initial agent poses:
\begin{align}
    M,\{p_0^i\}_{i\in I_0}=E(G,O_0).
\end{align}
During autoregressive rollout, no ground-truth future observations or poses are provided. The transition model updates shared state and predicts the next poses from current state and agent actions:
\begin{align}
    S_{t+1},\{p_{t+1}^i\}_{i\in I_t}=F_\theta(S_t,\{p_t^i,a_t^i\}_{i\in I_t};G).
\end{align}
Each requested observation is then independently decoded from the updated shared state:
\begin{align}
    o_{t+1}^i=R(p_{t+1}^i,S_{t+1},G).
\end{align}
This formulation emphasizes three properties of a multi-agent world model. First, \textbf{persistence}: the shared state preserves scene and entity information across agents, viewpoints, and temporary occlusions. Second, \textbf{update}: agent actions modify both the predicted agent poses and the shared dynamic state. Third, \textbf{scalability}: observations are generated through independent view queries whose interface does not assume a predefined number of agents.

Our primary objective is inference-time population scalability: supporting changes in the active agent set without modifying or retraining the transition or rendering architecture. We additionally study mesh-guided scene expansion, in which the geometry prior \(G\) is replaced across rollouts by a new scene geometry \(G'\). The model is then initialized for the new environment and generates observations conditioned on new structure. This setting evaluates whether the generated views follow the supplied geometry and remain consistent across views. Population scalability concerns changes in the entities inhabiting a world, whereas scene expansion concerns changes in the spatial structure of the world.

%% file: sec/4_method.tex
\section{Method}

Khora decouples world-state evolution from view synthesis (Figure~\ref{fig:pipeline}). Initialized from observations and a coarse scene prior, the STBoard maintains scene geometry, appearance memory, and dynamic entities (Section~\ref{sec:blackboard}). Actions update agent poses and entity states (Section~\ref{sec:transition}); a shared renderer combines view-specific state queries with visual history, then back-projects generated appearance into scene memory (Section~\ref{sec:renderer}). Agent populations and requested views can change during inference. We use a coarse mesh to specify scene structure separately from appearance synthesis, a strategy also explored in mesh-conditioned generation and rendering \citep{schneider2026worldmesh,chen2026dar}. Its geometry provides a common reference for view projection and geometric checks of proposed agent motion.

\input{sec/figure_4}


\subsection{STBoard}
\label{sec:blackboard}
Multi-agent models may organize representations around predefined
observers~\citep{savva2026solaris,hu2026mira}, tying the rendering interface
to the participating population.

Khora maintains a world-centric shared representation called the Spatio-temporal Board (\emph{STBoard}). It associates a fixed scene geometry \(G\) with a persistent appearance memory \(M\) and a dynamic entity table \(D_t\). The geometry provides the spatial scaffold of the environment. The appearance memory stores visual information associated with this scaffold, while the entity table represents active entities through attributes such as identity, pose, orientation, and task-dependent state.

The geometry remains fixed within a rollout, whereas the appearance memory and dynamic entity states are updated through different mechanisms. Agent actions drive updates to the entity table. Generated observations are back-projected onto the mesh to update the appearance memory, making accumulated scene appearance available to subsequent rendering queries. The initialization of scene memory and agent states uses the scene prior and available initial observations.

Each agent occupies an entity-table entry rather than a fixed view slot.
Adding an agent requires a new entry and rendering queries, not a change
to the renderer architecture.


\subsection{Action-conditioned World Evolution}
\label{sec:transition}

Khora updates poses and shared state before rendering, so agent interactions
are represented explicitly rather than inferred solely from image history.

For each agent, we form a deterministic proposal
$\hat p_{t+1}^i=K(p_t^i,a_t^i)$ using the known action-to-motion mapping $K$.
A learned residual corrects this proposal using the shared world state and
static scene context:
\begin{equation}
    p_{t+1}^i=\hat p_{t+1}^i+\Delta(S_t,G,\hat p_{t+1}^i,a_t^i).
\end{equation}
The kinematic term captures control-induced translation and rotation, while
the learned residual accounts for state-dependent effects such as collisions,
terrain, and interactions with other agents. The refined poses and actions
then update the shared dynamic state before rendering.


\subsection{Geometry-conditioned Rendering and Appearance Update}
\label{sec:renderer}
Khora projects shared 3D agent states into the target camera view, providing
explicit spatial conditions rather than requiring the renderer to infer
multi-agent configurations from latent tokens.

\textbf{View-conditioned spatial queries.} For a target camera pose $p_k$,
Khora queries the fixed geometry $G$ and current appearance memory $M_k$
for a scene condition $B_k$. Dynamic states at world time $t$ are
transformed into the same camera frame and rasterized into an entity
condition $C_k$:
\begin{equation}
\begin{aligned}
    B_k &= \mathcal P_{\mathrm{scene}}(G,M_k,p_k), \\
    C_k &= \mathcal P_{\mathrm{entity}}(S_t,p_k)
    \in\mathbb R^{H\times W\times d}.
\end{aligned}
\end{equation}
The entity condition has fixed dimensions as the population changes:
additional entities modify its content without adding input slots or
agent-specific renderer components.

\textbf{Autoregressive latent generation.} The renderer combines the spatial conditions with the target view's visual history. A causal Wan-VAE~\citep{wan2025} encodes the initial observations into latent context. Conditioned on this context and the shared-world query, a diffusion transformer~\citep{peebles2023dit} generates the next video latent block:
\begin{equation}
\hat z_k
\sim
p_\phi\!\left(z\mid H_k,B_k,C_k,p_k\right),
\end{equation}
where $H_k$ is the latent history. The video decoder produces
$\hat o_k=\mathcal D_{\mathrm{VAE}}(\hat z_k)$, and each prediction updates
the history for the next step
(Appendix~\ref{app:renderer}, Figure~\ref{fig:renderer_details}); all views reuse
the same renderer. The appendix also details windowed RGB decoding.

\textbf{Shared memory across views.} Latent history supplies per-view temporal
context, while the entity table and appearance memory share agent states and
scene information across views. Communication occurs through STBoard updates,
not direct exchange between observation streams inside the neural renderer.

\textbf{Computational cost.} Let $N_a$ denote the number of active agents and $N_v$ the number of requested views. In view-coupled generators~\citep{savva2026solaris,hu2026mira}, dense cross-agent communication inside the neural renderer can require $O(N_aN_v)$ expensive neural interactions, which becomes $O(N^2)$ when every agent receives a rendered view. Khora invokes the neural renderer independently for each requested view, resulting in $O(N_v)$ renderer evaluations. Constructing the conditioning maps still requires up to $O(N_aN_v)$ entity-to-view projections. However, the practical cost can be expressed as
\begin{align}
    T_{\text{Khora}}\approx N_vC_{\text{render}}+N_aN_vC_{\text{proj}},
\end{align}
where $C_{\text{render}}\gg C_{\text{proj}}$.
For up to 80 agents, projection and rasterization account for less than 8\% of total inference latency.

\subsection{Training Data and Inference Protocol}
Khora is trained on Counter-Strike: Global Offensive (CS:GO) trajectories collected on a single map, Dust2, with aligned visual observations, actions, and structured state supervision.

The source episodes of the training set include up to 10 players, while each indexed training segment contains 2--4 player views. We sample exclusively from the valid temporal intervals specified in the training index. These intervals contain 44,486,627 player-view frames at 32 FPS, totaling 411.9 hours of aggregate player-view video.

A coarse world prior (mesh) initializes the shared scene representation. At inference time, Khora receives this prior, initial observations, and subsequent actions, and autoregressively generates future poses, dynamic states, and observations.

%% file: sec/figure_4.tex
\begin{figure}[t]
  \centering
  \includegraphics[width=\linewidth]{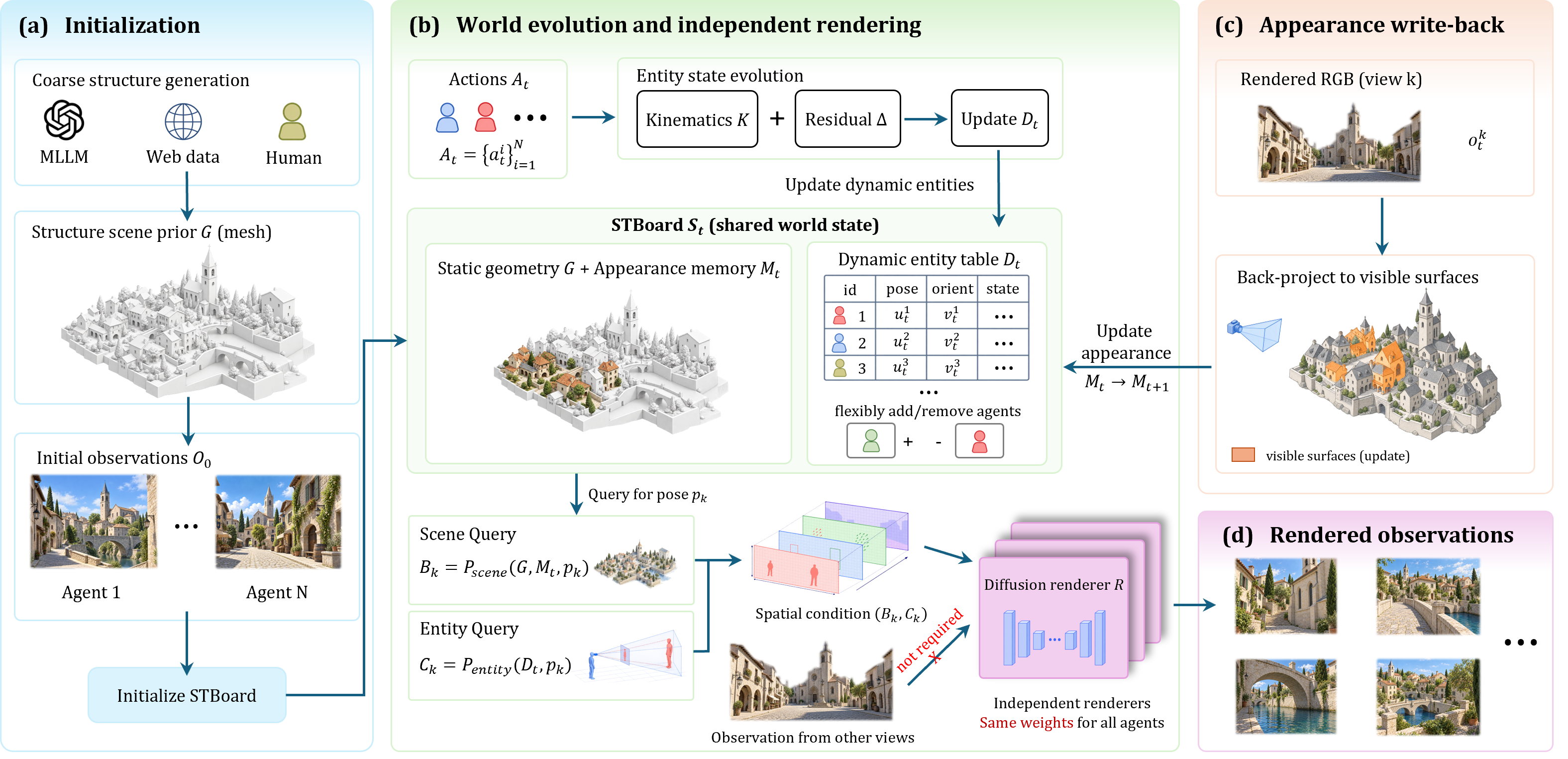}
  \caption{\textbf{Architecture of Khora.} Initial observations and a coarse scene prior initialize the shared STBoard (Spatio-temporal Board). Actions update the dynamic entity table through kinematic and residual transitions. Each requested view queries the shared state for pose- and scene-conditioned inputs, which are decoded by a shared-weight diffusion renderer. Generated RGB observations are back-projected onto visible surfaces to update the persistent appearance memory, while agents can be added or removed without retraining.}
  \label{fig:pipeline}
\end{figure}

%% file: sec/5_exp.tex
\section{Experiments}

\input{sec/figures_5_to_7}


We evaluate visual quality, multi-agent consistency, computational scaling, dynamic populations, and mesh-guided generalization. External methods are ordered by first public release, followed by the Khora ablation and full model.

\textbf{Khora w/o Cross-Agent State} masks other-agent state conditions at inference while retaining scene memory, ego conditions, and visual history. It uses the same renderer checkpoint and inference settings without retraining; visual history may still contain other agents.



\subsection{Visual Quality}

We assess visual fidelity using paired metrics (PSNR, SSIM~\citep{wang2004ssim}, and LPIPS~\citep{zhang2018lpips}) and distribution-level metrics (FID~\citep{heusel2018fid} over individual frames and FVD~\citep{unterthiner2019fvd} over complete generated clips).

Paired metrics compare generated and synchronized GT frames, averaging over agents and timesteps. Scores use a common resolution but retain model-specific temporal windows. Evaluation splits, sample counts, and scoring windows are detailed in Appendix~\ref{app:setup}.

\begin{table}[t]
\centering
\caption{\textbf{Quantitative comparison of visual quality.}
Scores use a common resolution; temporal windows and available evaluation cohorts differ (Appendix~\ref{app:setup}).}
\label{tab:visual_quality}
\small
\setlength{\tabcolsep}{4pt}
\begin{tabularx}{\linewidth}{LN{99.9999}N{9.9999}N{9.9999}N{999.9999}N{999.9999}}
\toprule
\tableheadleft{Method} & \tablehead{PSNR $\uparrow$} & \tablehead{SSIM $\uparrow$} & \tablehead{LPIPS $\downarrow$} & \tablehead{FID $\downarrow$} & \tablehead{FVD $\downarrow$} \\
\midrule
\multicolumn{6}{l}{\textit{2 Views (10,000 cases / 20,000 videos)}} \\
Solaris \citep{savva2026solaris} & 20.1465 & 0.5273 & 0.1835 & \bestscore{4.3726} & \bestscore{11.4263} \\
\rowcolor{tableStripe}
Gamma-World~\citep{liu2026gammaworld} & 15.5611 & 0.4332 & 0.5607 & 53.9303 & 297.1115 \\
MIRA~\citep{hu2026mira} & 21.7290 & 0.5973 & 0.3927 & 16.0886 & 231.9198 \\
\rowcolor{tableStripe}
Khora w/o Cross-Agent State & \secondscore{26.0427} & \secondscore{0.7125} & \secondscore{0.1826} & 8.7262 & 51.6657 \\
\rowcolor{latentblue}
\textbf{Khora} & \bestscore{26.2425} & \bestscore{0.7130} & \bestscore{0.1789} & \secondscore{7.7098} & \secondscore{36.5529} \\
\midrule
\multicolumn{6}{l}{\textit{4 Views (10,000 cases / 40,000 videos)}\hspace{0.6em}{\scriptsize\itshape No comparable results for Solaris (configuration limits).}} \\
Gamma-World~\citep{liu2026gammaworld} & 16.9473 & 0.5025 & 0.4856 & 55.8759 & 272.2558 \\
MIRA~\citep{hu2026mira} & 20.8771 & 0.6072 & 0.3570 & 16.3253 & 198.0576 \\
\rowcolor{tableStripe}
Khora w/o Cross-Agent State & \secondscore{24.9196} & \secondscore{0.7017} & \secondscore{0.1851} & \secondscore{11.3010} & \secondscore{73.4917} \\
\rowcolor{latentblue}
\textbf{Khora} & \bestscore{25.2825} & \bestscore{0.7020} & \bestscore{0.1772} & \bestscore{7.9821} & \bestscore{38.6849} \\
\midrule
\multicolumn{6}{l}{\textit{8 Views (159 cases / 1,272 videos)}\hspace{0.6em}{\scriptsize\itshape No comparable results for Solaris, Gamma-World, or MIRA (configuration limits).}} \\
\rowcolor{tableStripe}
Khora w/o Cross-Agent State & 26.2282 & 0.7573 & 0.1544 & 30.1706 & 104.9082 \\
\rowcolor{latentblue}
\textbf{Khora} & \bestscore{26.5775} & \bestscore{0.7573} & \bestscore{0.1477} & \bestscore{27.0311} & \bestscore{53.6148} \\
\bottomrule
\end{tabularx}
\par\vspace{8pt}
\parbox{\linewidth}{\footnotesize\emph{Note.} All external baselines completed all required training stages on our dataset. Checkpoints, evaluation settings, and sample counts are detailed in Appendix~\ref{app:baseline_protocol}.}
\end{table}

Table~\ref{tab:visual_quality} reports per-view quality across population sizes.
Khora improves over its input ablation on most metrics while retaining a
population-independent rendering interface. Appendix~\ref{app:setup} details
checkpoints, preprocessing, and evaluation scope.

\subsection{Multi-Agent Consistency}

We evaluate multi-agent consistency using human ratings on a five-point scale
and scene-level point-cloud geometry and color scores inspired by MetaWorld's CVC formulation
\citep{hu2026metaworld}. Specifically, \textbf{CVC-XView} evaluates the similarity of reconstructed
generated views; \textbf{CVC-GT} compares each generated view with its GT,
then averages over views. These proxies do not establish actor synchronization.
All five methods use the same human-rating protocol
(Appendix~\ref{app:user_study}); the evaluation sets and temporal windows for
the automatic metrics are detailed in Appendix~\ref{app:geometry_metric}.

\begin{table}[t]
\centering
\caption{\textbf{Human evaluation and automatic world consistency.} Human ratings use a five-point scale; CVC scores lie in $[0,1]$. Higher is better. CVC-XView is evaluated on 10,000 common cases across all five models. CVC-GT and CVC-XView use different evaluation cohorts and temporal windows; Appendix~\ref{app:geometry_metric} details both protocols and their component scores.}
\label{tab:multi_agent_consistency}
\small
\setlength{\tabcolsep}{4pt}
\begin{tabularx}{\linewidth}{LN{9.999}N{9.999}N{9.999}N{9.999999}N{9.999999}}
\toprule
& \multicolumn{3}{c}{\textbf{Human evaluation}} & \multicolumn{2}{c}{\textbf{Automatic world scores}} \\
\cmidrule(lr){2-4}\cmidrule(l){5-6}
\tableheadleft{Method} & \tablehead{Action $\uparrow$} & \tablehead{World $\uparrow$} & \tablehead{Overall $\uparrow$} & \tablehead{CVC-GT $\uparrow$} & \tablehead{CVC-XView $\uparrow$} \\
\midrule
Solaris~\citep{savva2026solaris} & 3.000 & \secondscore{4.105} & 3.553 & 0.918310 & 0.836662 \\
\rowcolor{tableStripe}
Gamma-World~\citep{liu2026gammaworld} & 2.027 & 2.786 & 2.406 & 0.897661 & 0.836490 \\
MIRA~\citep{hu2026mira} & \secondscore{3.170} & 4.009 & \secondscore{3.589} & 0.922486 & 0.830889 \\
\rowcolor{tableStripe}
Khora w/o Cross-Agent State & 2.075 & 4.065 & 3.070 & \secondscore{0.931900} & \secondscore{0.838207} \\
\rowcolor{latentblue}
\textbf{Khora} & \bestscore{3.570} & \bestscore{4.213} & \bestscore{3.892} & \bestscore{0.932200} & \bestscore{0.838773} \\
\bottomrule
\end{tabularx}
\end{table}

Khora receives the highest human ratings (Table~\ref{tab:multi_agent_consistency}).
Its small CVC-XView gain over the ablation is descriptive, not evidence of
synchronized character motion.

\paragraph{GT-referenced cross-view action consistency.}
We measure the other player's location, image motion, pose, and appearance
against GT in one ego view per case. GT-only eligibility checks select 1,643
of 10,000 held-out evaluation cases.
Khora uses supplied trajectories and state conditions: this evaluates
state-conditioned rendering, not autonomous dynamics or direct agreement
between generated views (Appendix~\ref{app:action_protocol}).

\input{sec/action_main_table}

Table~\ref{tab:cross_view_action} separates two physical windows because
Solaris and MIRA have no common purely generated interval. With checkpoint
and inference settings fixed, removing other-agent conditions degrades
localization, motion, pose, and appearance in both windows. This supports
their contribution to visible actor fidelity, not long-horizon dynamics.
Full metrics and paired confidence intervals appear in
Appendices~\ref{app:action_full} and~\ref{app:action_uncertainty}.


\subsection{Scalability and Real-Time Performance}

Figure~\ref{fig:scalability_runtime} reports measured scaling from 1--80 agents:
latency rises from 107.16\,ms to 116.73\,ms per rollout step and per-view FPS
falls from 37.33 to 34.27, with view synthesis dominant. Peak per-GPU VRAM is
nearly constant; distributed aggregate throughput rises from 37.3 to 2741.3
view-fps. Projections beyond 80 agents are extrapolations (Appendix~\ref{app:runtime}).

\runtimefigure

\subsection{Inference-Time Dynamic Population}

We finally evaluate whether the active population can change within a single autoregressive rollout. The sequence begins with two agents. Additional agents are introduced at later timesteps, increasing the population from two to four and then from four to eight. Several agents subsequently leave the environment, reducing the population again. Newly introduced agents are initialized from their first observations and appended to the dynamic entity table, while departing agents are removed from the active set. The transition model, STBoard interface, and neural renderer remain unchanged throughout the rollout.

\dynamicpopulationfigure

Figure~\ref{fig:dynamic_population} illustrates that Khora can accommodate population changes during an ongoing simulation. Existing agents preserve their trajectories and observations when new agents are introduced, while the new agents become visible from relevant viewpoints after being inserted into the shared state. This experiment demonstrates that population scalability is an inference-time capability rather than a fixed configuration determined during training.

\subsection{Mesh-guided Generalization to Unseen Environments}
\label{sec:mesh_generalization}

We examine whether Khora can generate scene imagery conditioned on geometry not encountered during training. We supply previously unseen meshes as structural priors and use the same trained model
weights as in the main experiments, without additional weight updates. Figure~\ref{fig:unseen_mesh} compares the input mesh
structures with the corresponding generated observations from matched camera viewpoints. These examples show that Khora synthesizes visual appearance that follows the supplied spatial
layouts, demonstrating mesh-guided scene generalization beyond the training environments.

\unseenmeshfigure

%% file: sec/figures_5_to_7.tex
\newcommand{\runtimefigure}{%
\begin{figure}[t]
    \centering
    \includegraphics[width=\textwidth]{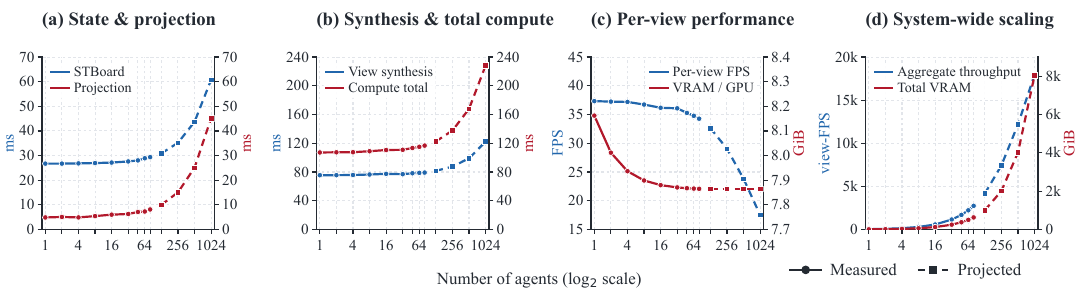}
    \caption{\textbf{Runtime scalability with increasing agent population.}
    Blue and red denote the left- and right-axis metrics in each panel, respectively.
    Solid circles are measurements (1--80 agents); dashed squares are projections
    (128--1024 agents). Curves are drawn separately across the measured/projected
    boundary. Aggregate throughput equals agent count times per-view FPS;
    each requested view is synthesized on its respective GPU.}
    \label{fig:scalability_runtime}
\end{figure}
}

\definecolor{agentred}{RGB}{205,65,65}
\definecolor{agentblue}{RGB}{55,105,190}
\definecolor{agentgreen}{RGB}{55,155,75}
\definecolor{agentyellow}{RGB}{170,125,15}

\definecolor{statebg}{RGB}{248,248,248}
\definecolor{stateborder}{RGB}{210,210,210}
\definecolor{statetext}{RGB}{105,111,117}

\newlength{\dynamicviewwidth}
\newlength{\dynamicviewheight}
\setlength{\dynamicviewwidth}{0.20625\textwidth}
\setlength{\dynamicviewheight}{0.1160773\textwidth}

\newcommand{\compactview}[1]{%
\raisebox{-0.5\height}{\includegraphics[width=\dynamicviewwidth]{#1}}%
}

\newcommand{\stagecell}[2]{%
\raisebox{-0.5\height}{%
\parbox[b][\dynamicviewheight][c]{0.140\textwidth}{%
\raggedright{\scriptsize\bfseries\mbox{#1}}\\[2pt]
{\scriptsize #2}%
}%
}%
}

\newcommand{\joincell}[1]{%
\raisebox{1.7pt}{\raisebox{-0.5\height}{%
\fcolorbox{stateborder}{statebg}{%
\parbox[b][\dimexpr\dynamicviewheight-2\fboxsep-2\fboxrule\relax][c]{\dimexpr\dynamicviewwidth-2\fboxsep-2\fboxrule\relax}{%
\centering
{\normalsize\textcolor{#1}{\(\oplus\)}}\\[1pt]
{\scriptsize\bfseries\textcolor{statetext}{Not yet active}}
}}%
}}%
}

\newcommand{\leavecell}[1]{%
\raisebox{1.7pt}{\raisebox{-0.5\height}{%
\fcolorbox{stateborder}{statebg}{%
\parbox[b][\dimexpr\dynamicviewheight-2\fboxsep-2\fboxrule\relax][c]{\dimexpr\dynamicviewwidth-2\fboxsep-2\fboxrule\relax}{%
\centering
{\normalsize\textcolor{#1}{\(\ominus\)}}\\[1pt]
{\scriptsize\bfseries\textcolor{statetext}{Removed}}
}}%
}}%
}

\newcommand{\dynamicpopulationfigure}{%
\begin{figure}[t]
\centering

\begin{tabular}{@{}c@{\hspace{2pt}}c@{\hspace{2pt}}c@{\hspace{2pt}}c@{\hspace{2pt}}c@{}}

& \textbf{\textcolor{agentred}{Agent 1}} &
\textbf{\textcolor{agentblue}{Agent 2}} &
\textbf{\textcolor{agentgreen}{Agent 3}} &
\textbf{\textcolor{agentyellow}{Agent 4}}
\\[2pt]

\stagecell{Initial}{$\boldsymbol{t}_{0}\;(N=2)$} &
\compactview{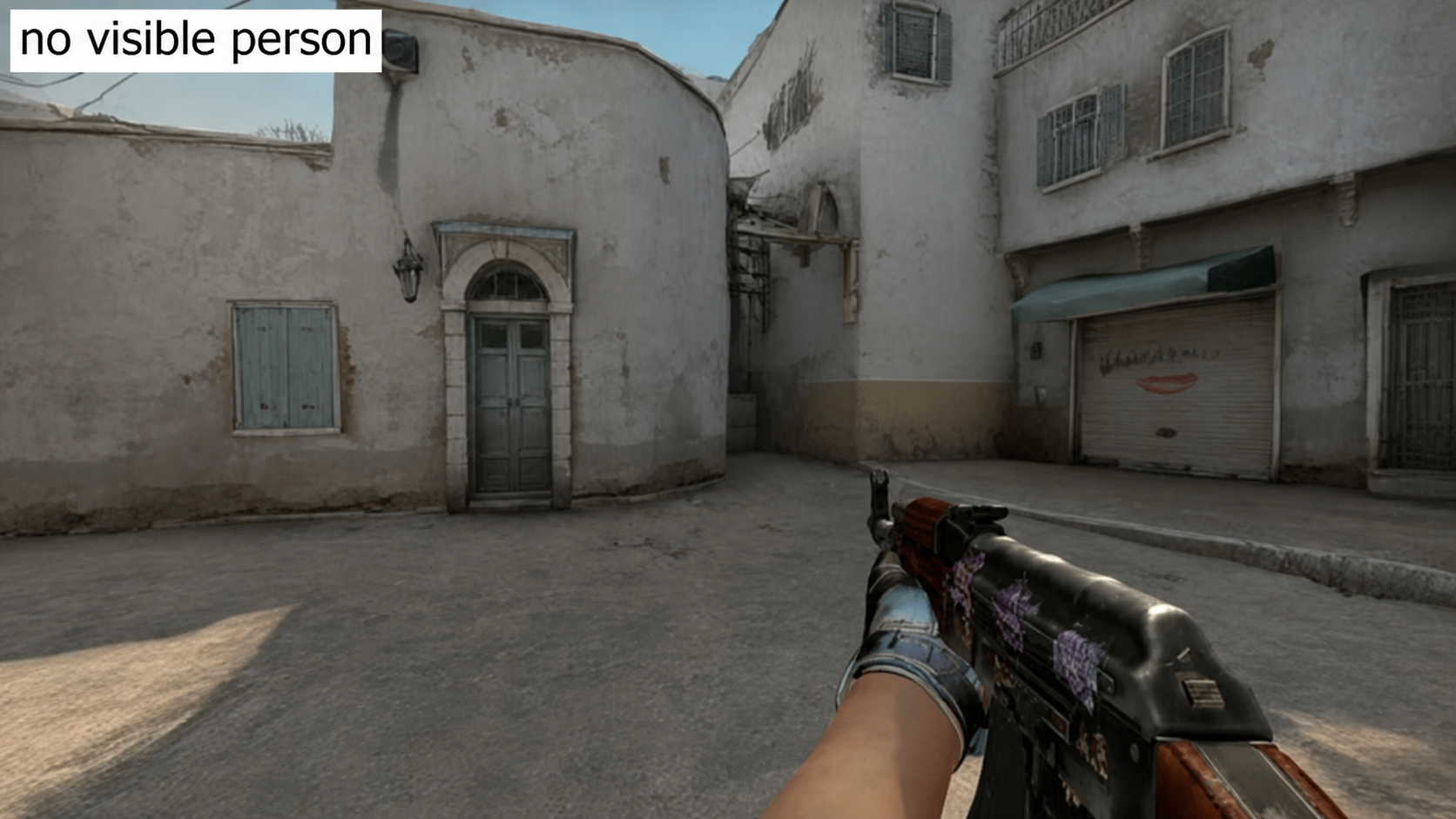} &
\compactview{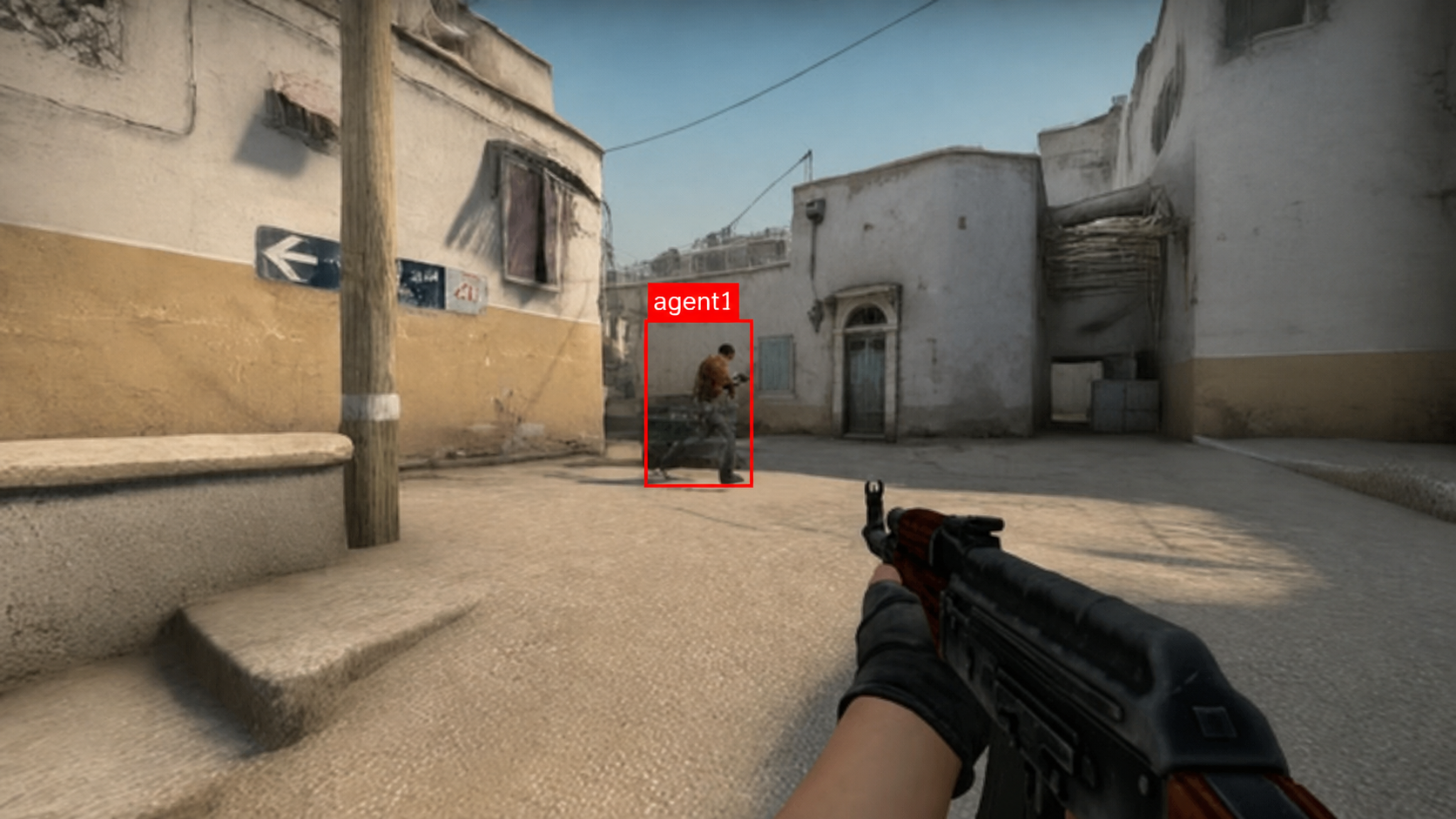} &
\joincell{agentgreen} &
\joincell{agentyellow}
\\\noalign{\vskip 3pt}

\stagecell{Agents 3, 4 join}{$\boldsymbol{t}_{1}\;(N:2\rightarrow4)$} &
\compactview{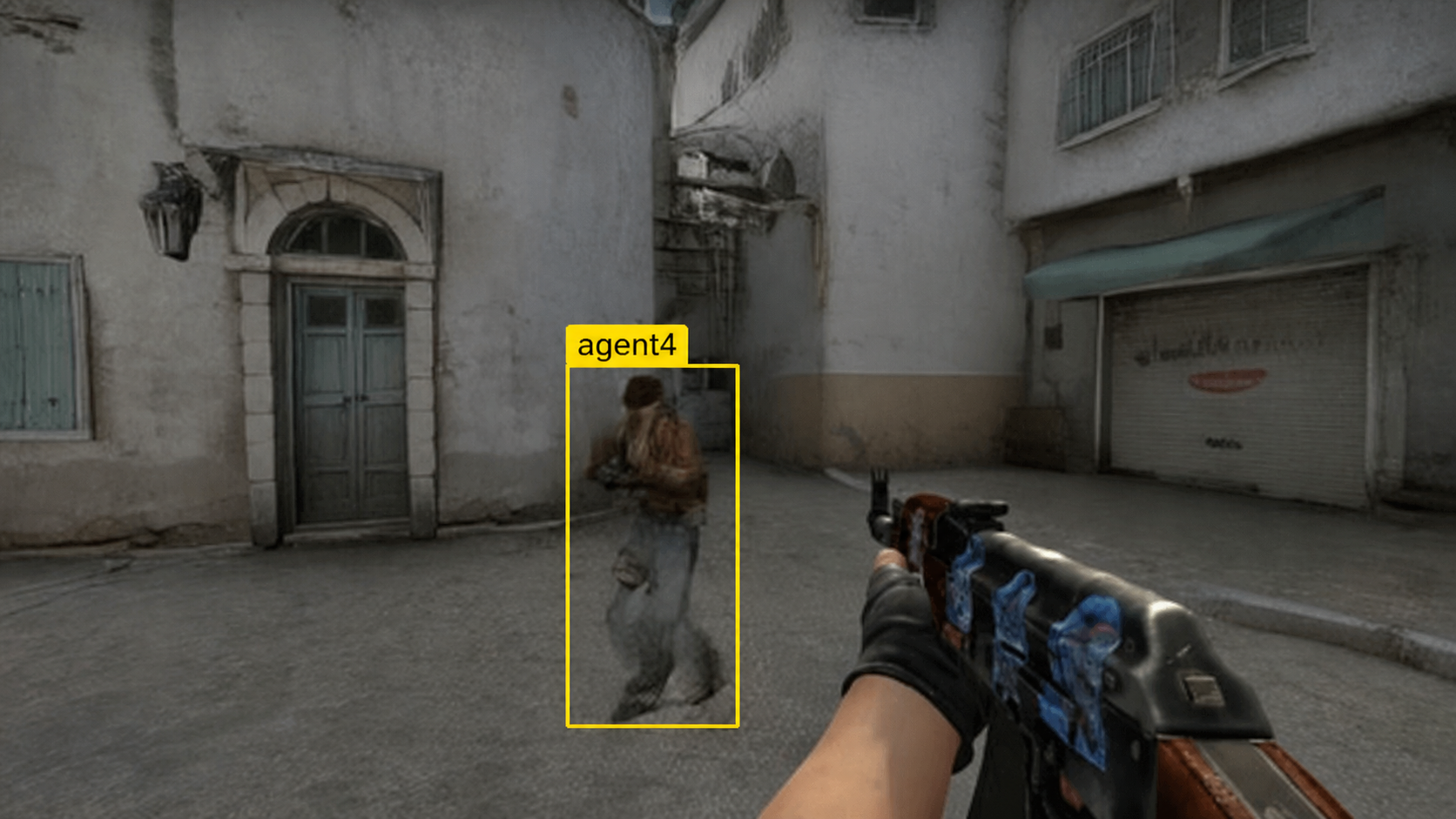} &
\compactview{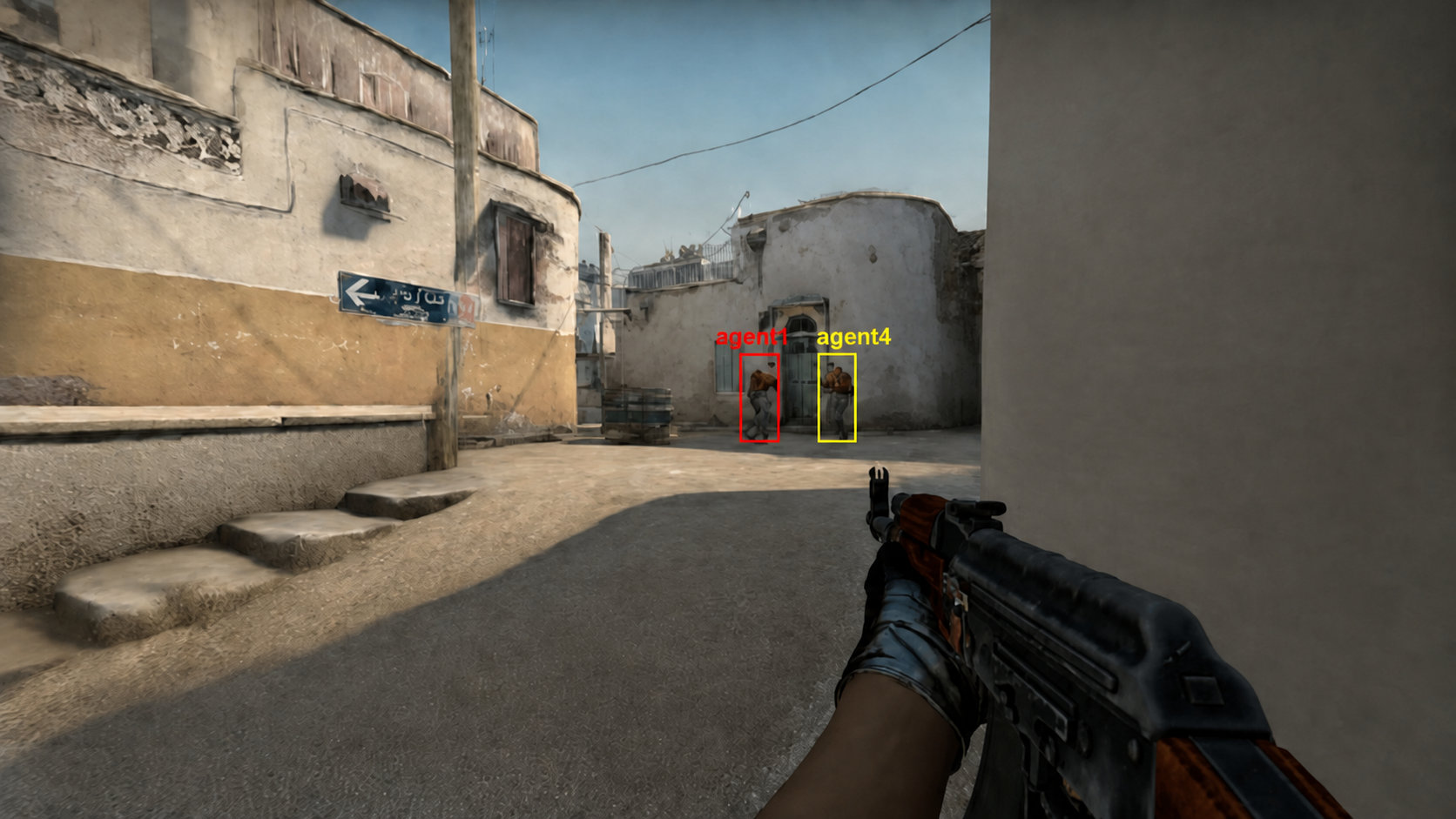} &
\compactview{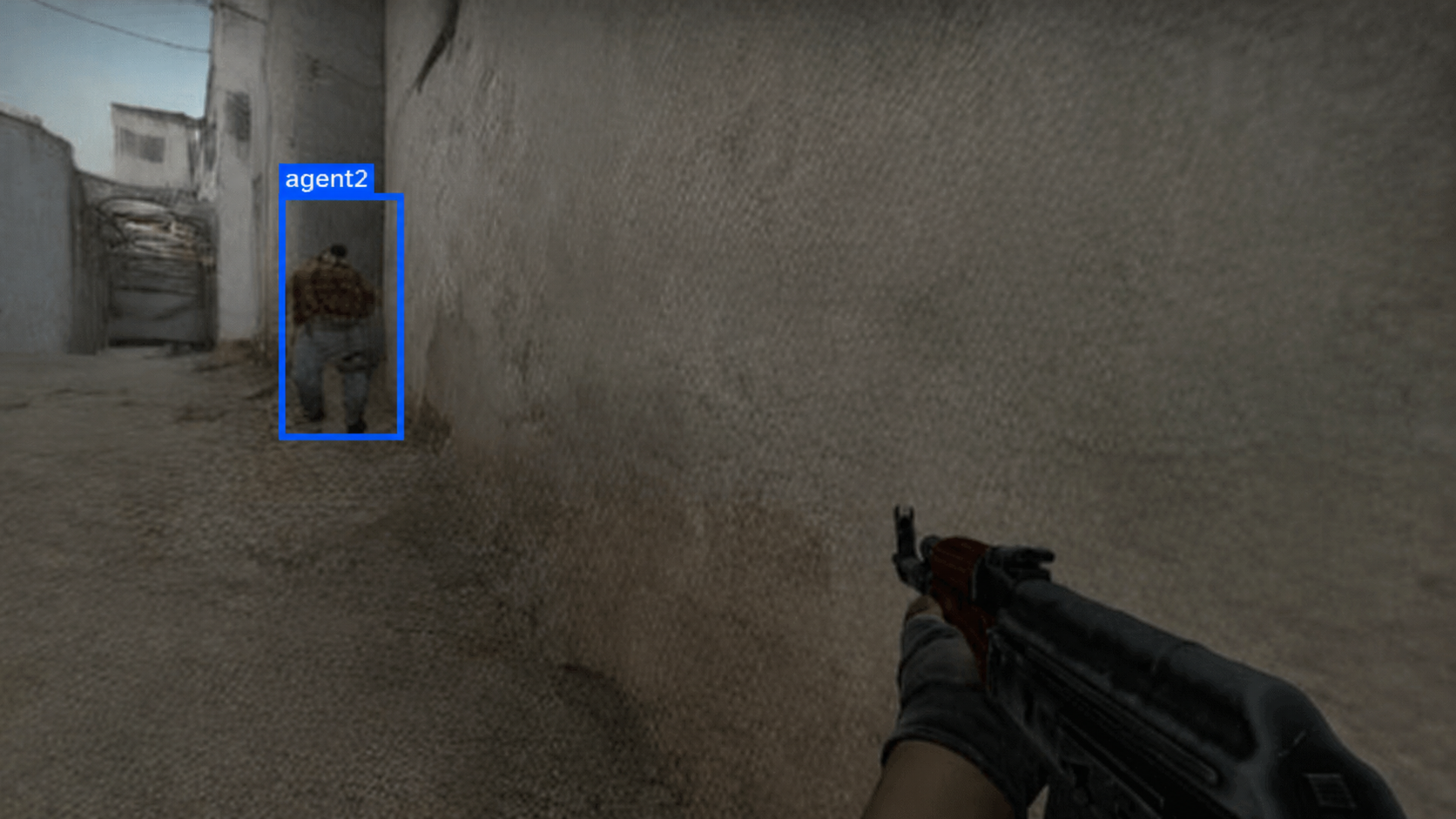} &
\compactview{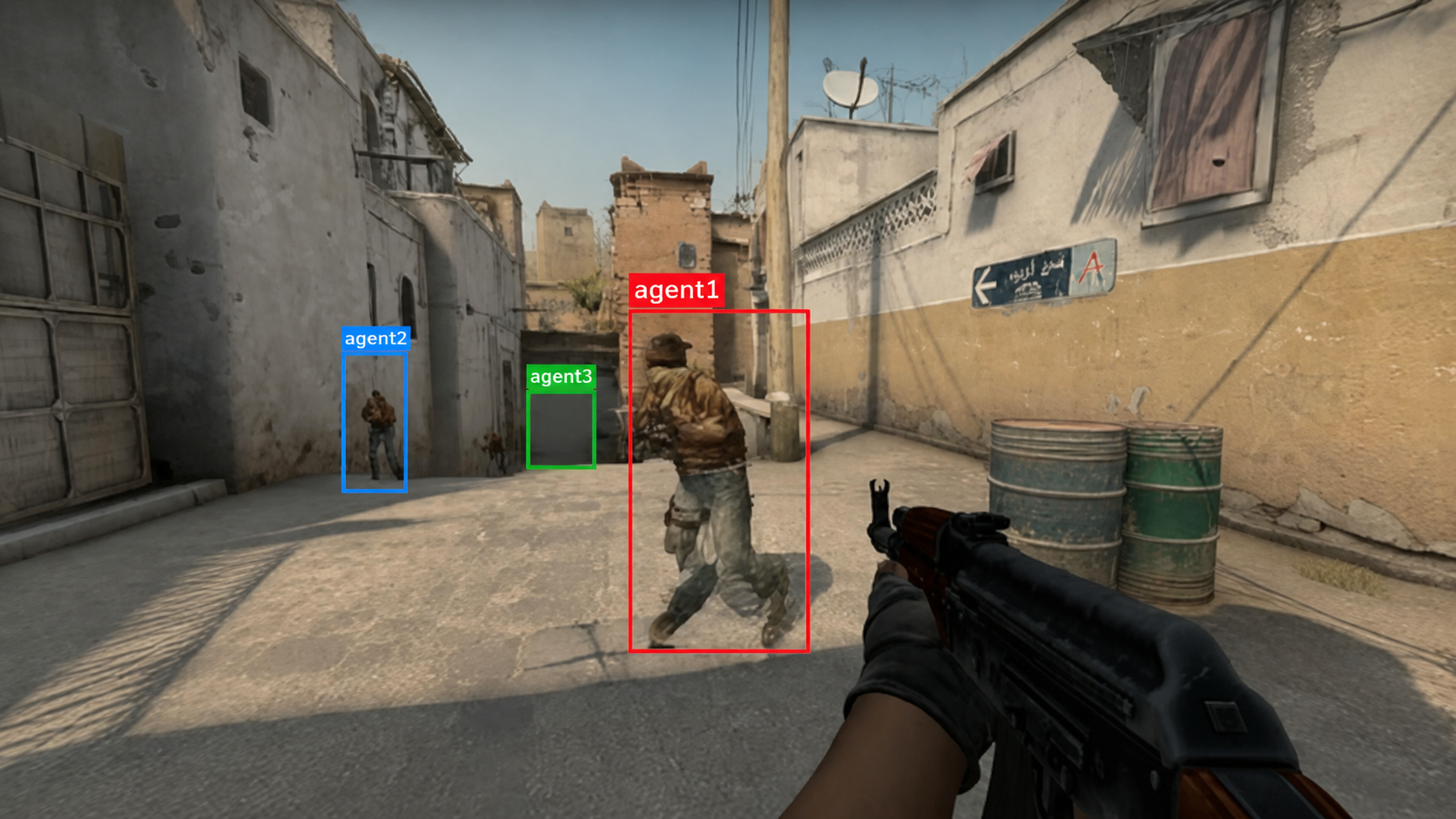}
\\\noalign{\vskip 3pt}

\stagecell{Agents 1, 2 leave}{$\boldsymbol{t}_{2}\;(N:4\rightarrow2)$} &
\leavecell{agentred} &
\leavecell{agentblue} &
\compactview{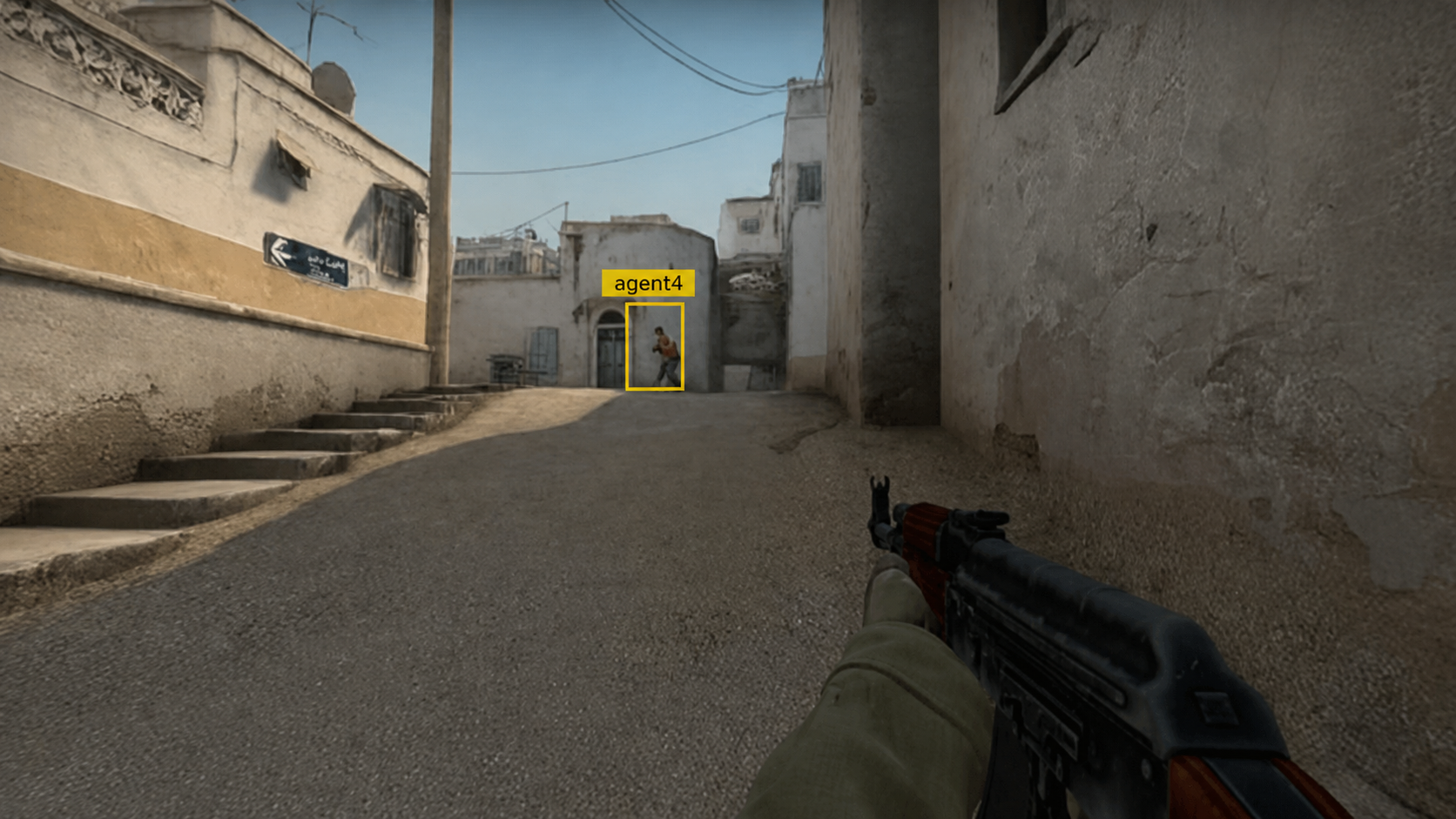} &
\compactview{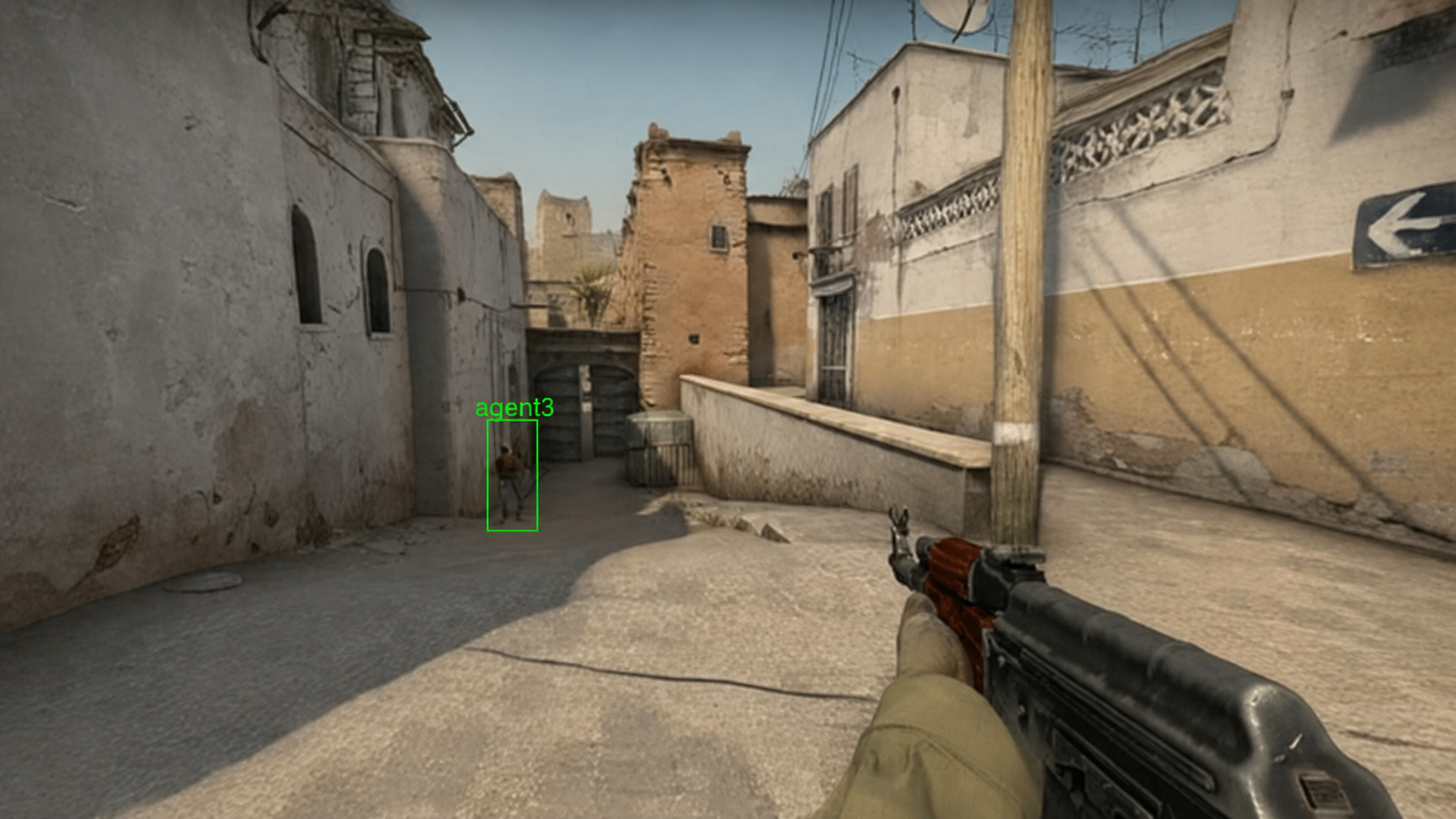}

\end{tabular}

\caption{
\textbf{Dynamic population within a single rollout.}
Agents enter and leave while the shared world persists; model weights and the
transition and rendering architecture remain unchanged.
}
\label{fig:dynamic_population}
\end{figure}
}

\newcommand{\unseenmeshfigure}{%
\begin{figure}[t]
    \centering

    \newcommand{\meshrowlabel}[1]{%
        \parbox[c]{0.12\textwidth}{\raggedright\small\bfseries ##1}%
    }
    \begin{tabular}{@{}c@{\hspace{0.0075\textwidth}}c@{\hspace{0.0075\textwidth}}c@{\hspace{0.0075\textwidth}}c@{\hspace{0.0075\textwidth}}c@{}}
        & \small\textbf{Case 1} & \small\textbf{Case 2} &
        \small\textbf{Case 3} & \small\textbf{Case 4} \\[3pt]
        \meshrowlabel{Input mesh} &
        \begin{minipage}[c][0.11953125\textwidth][c]{0.2125\textwidth}%
            \centering\includegraphics[width=\linewidth]{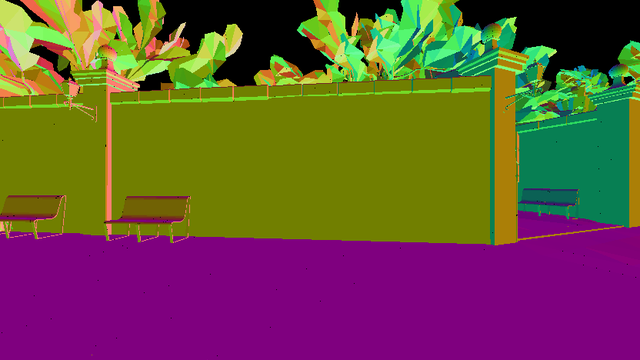}%
        \end{minipage} &
        \begin{minipage}[c][0.11953125\textwidth][c]{0.2125\textwidth}%
            \centering\includegraphics[width=\linewidth]{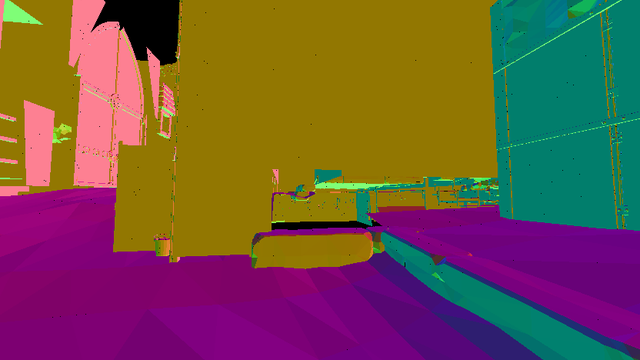}%
        \end{minipage} &
        \begin{minipage}[c][0.11953125\textwidth][c]{0.2125\textwidth}%
            \centering\includegraphics[width=\linewidth]{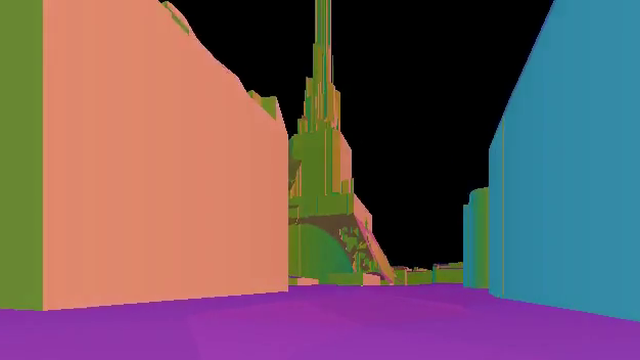}%
        \end{minipage} &
        \begin{minipage}[c][0.11953125\textwidth][c]{0.2125\textwidth}%
            \centering\includegraphics[width=\linewidth]{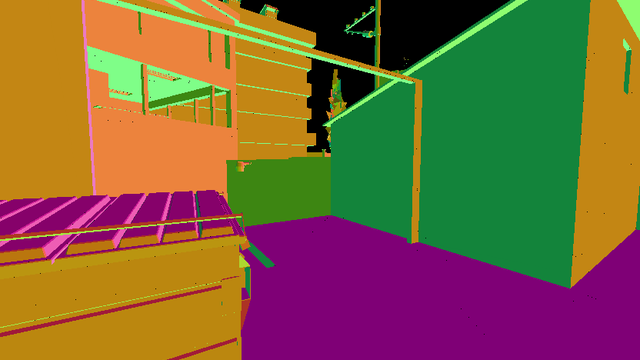}%
        \end{minipage}
        \\\noalign{\vskip 3pt}

        \meshrowlabel{Generated} &
        \begin{minipage}[c][0.11953125\textwidth][c]{0.2125\textwidth}%
            \centering\includegraphics[width=\linewidth]{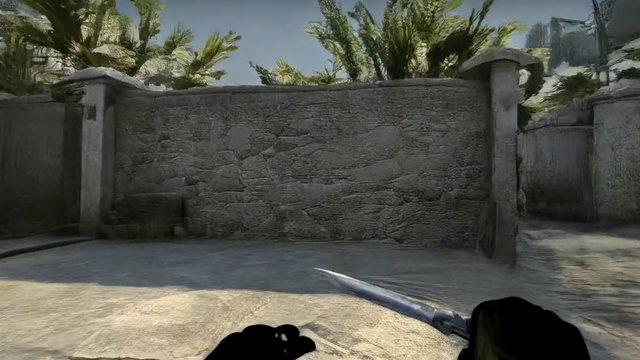}%
        \end{minipage} &
        \begin{minipage}[c][0.11953125\textwidth][c]{0.2125\textwidth}%
            \centering\includegraphics[width=\linewidth]{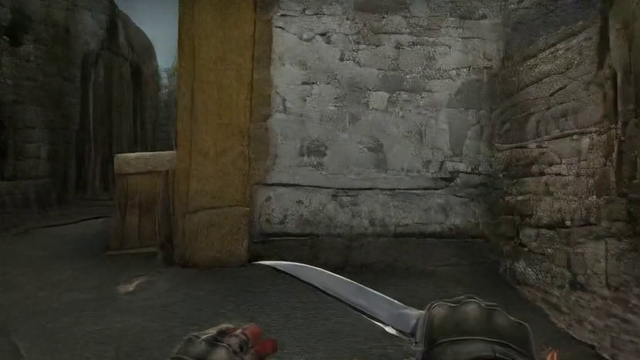}%
        \end{minipage} &
        \begin{minipage}[c][0.11953125\textwidth][c]{0.2125\textwidth}%
            \centering\includegraphics[width=\linewidth]{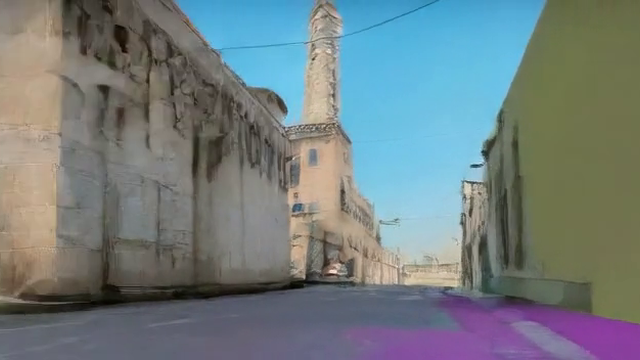}%
        \end{minipage} &
        \begin{minipage}[c][0.11953125\textwidth][c]{0.2125\textwidth}%
            \centering\includegraphics[width=\linewidth]{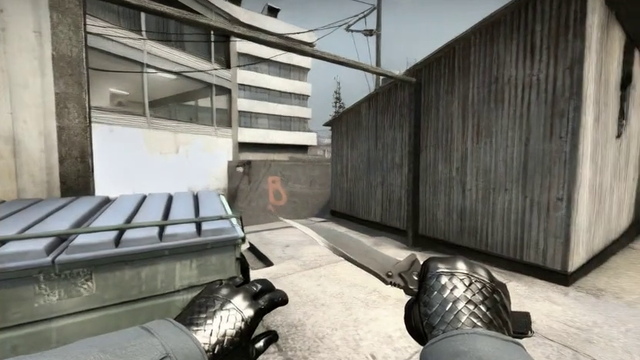}%
        \end{minipage}
    \end{tabular}

    \caption{
        \textbf{Mesh-guided generation in unseen environments.} Each column pairs an unseen mesh (top) with the corresponding generated observation (bottom). The same trained weights are used without additional updates; the generated views follow the supplied scene geometry.
    }
    \label{fig:unseen_mesh}
\end{figure}
}

%% file: sec/action_main_table.tex
\begin{table}[t]
\centering
\caption{\textbf{Cross-view action consistency with a GT reference.} Each panel scores the same 1,643 eligible cases at matched physical frames. Conditioning histories differ across model families; Khora and its ablation share the checkpoint and inference settings. PCK uses a threshold of 20\% of GT box height; FG denotes fixed GT foreground. MAE is lower-is-better.}
\label{tab:cross_view_action}
\small
\setlength{\tabcolsep}{3pt}
\begin{tabularx}{\linewidth}{LN{9.9999}N{9.9999}N{9.9999}N{9.9999}N{9.9999}N{9.9999}}
\toprule
\tableheadleft{Method} & \tablehead{F1 $\uparrow$} & \tablehead{Recall $\uparrow$} & \tablehead{Direction $\uparrow$} & \tablehead{PCK $\uparrow$} & \tablehead{FG-SSIM $\uparrow$} & \tablehead{FG-MAE $\downarrow$} \\
\midrule
\multicolumn{7}{l}{\textit{Relative frames 13--32 (zero-based)}} \\
Solaris~\citep{savva2026solaris} & 0.1683 & 0.1611 & \secondscore{0.5267} & \secondscore{0.2111} & 0.1803 & 0.1647 \\
\rowcolor{tableStripe}
Gamma-World~\citep{liu2026gammaworld} & 0.0619 & 0.0525 & 0.1963 & 0.0500 & 0.1568 & 0.2000 \\
Khora w/o Cross-Agent State & \secondscore{0.2842} & \secondscore{0.1850} & 0.0654 & 0.1723 & \secondscore{0.2487} & \secondscore{0.1417} \\
\rowcolor{latentblue}
\textbf{Khora} & \bestscore{0.9041} & \bestscore{0.8721} & \bestscore{0.7121} & \bestscore{0.7438} & \bestscore{0.3910} & \bestscore{0.0834} \\
\midrule
\multicolumn{7}{l}{\textit{Relative frames 38--57 (zero-based)}} \\
Gamma-World~\citep{liu2026gammaworld} & 0.0195 & 0.0151 & 0.0921 & 0.0181 & 0.1476 & 0.2261 \\
\rowcolor{tableStripe}
MIRA~\citep{hu2026mira} & \secondscore{0.4525} & \secondscore{0.4299} & \secondscore{0.4507} & \secondscore{0.4129} & \secondscore{0.2555} & \secondscore{0.1268} \\
Khora w/o Cross-Agent State & 0.0087 & 0.0047 & 0.0043 & 0.0045 & 0.2201 & 0.1625 \\
\rowcolor{latentblue}
\textbf{Khora} & \bestscore{0.9024} & \bestscore{0.8645} & \bestscore{0.6929} & \bestscore{0.7385} & \bestscore{0.3609} & \bestscore{0.0868} \\
\bottomrule
\end{tabularx}
\end{table}

%% file: sec/6_conclusion.tex
\section{Discussion \& Limitations}

Khora requires authored geometry, reliable memory, and rendering resources.
Mesh-guided results do not establish dynamics generalization; motion,
collisions, action consequences, structural accuracy, and long-horizon
consistency require broader evaluation. Appearance back-projection can
contaminate memory with generation or alignment errors under occlusion or
conflicting observations. Rendering costs grow with views; projection costs
up to $O(N_aN_v)$, with latency depending on population, scene representation,
scheduling, and hardware.

Future work includes automatic geometry acquisition, closed-loop evaluation,
confidence-aware updates, cross-view conflict handling, visibility-aware
projection, and efficient rendering.

\section{Conclusion}

We present Khora, a multi-agent world model that separates shared-state evolution from per-view rendering through a persistent STBoard, enabling dynamic populations without architectural changes or retraining. Experiments assess visual quality, multi-agent consistency, population flexibility, and computational scaling, while qualitative results demonstrate generation conditioned on unseen meshes with unchanged model weights. We hope this work provides a practical step toward scalable open-world simulators capable of supporting large numbers of interacting agents, and encourages future research on world-centric representations and scalable interactive world models.

%% file: sec/appendix.tex
\begingroup
\captionsetup{font=normalsize,skip=11pt}
\setcounter{secnumdepth}{3}
\setcounter{topnumber}{3}
\setcounter{bottomnumber}{0}
\setcounter{totalnumber}{3}
\renewcommand{\topfraction}{0.95}
\renewcommand{\textfraction}{0}
\renewcommand{\floatpagefraction}{0.5}
\newcommand{\appneedspace}[1]{%
  \par
  \ifdim\pagegoal<\maxdimen
    \ifdim\dimexpr\pagegoal-\pagetotal\relax<#1\relax
      \newpage
    \fi
  \fi
}

\section*{Appendix Guide}
The appendix details the renderer, evaluation protocols, and additional results.

\begin{tabularx}{\linewidth}{lLc}
\toprule
\tableheadleft{Section} & \tableheadleft{Topic} & \tablehead{Page} \\
\midrule
\ref{app:renderer} & \hyperref[app:renderer]{Design of diffusion renderer} & \pageref{app:renderer} \\
\rowcolor{tableStripe}
\ref{app:setup} & \hyperref[app:setup]{Experimental setup and visual quality} & \pageref{app:setup} \\
\ref{app:human} & \hyperref[app:human]{Human evaluation} & \pageref{app:human} \\
\rowcolor{tableStripe}
\ref{app:geometry_metric} & \hyperref[app:geometry_metric]{Cross-view world consistency} & \pageref{app:geometry_metric} \\
\ref{app:action} & \hyperref[app:action]{Cross-view action consistency} & \pageref{app:action} \\
\rowcolor{tableStripe}
\ref{app:runtime} & \hyperref[app:runtime]{Runtime and scalability} & \pageref{app:runtime} \\
\ref{app:visual_coverage} & \hyperref[app:visual_coverage]{Additional qualitative results} & \pageref{app:visual_coverage} \\
\bottomrule
\end{tabularx}

Arrows indicate preferred metric directions. Bold and underlined values mark
the best and second-best results within each comparison group; two-method
comparisons highlight only the better value. Counts, uncertainty statistics,
and diagnostic controls are not ranked.

\input{sec/appendix_renderer}
\FloatBarrier
\input{sec/appendix_setup}

\FloatBarrier
\input{sec/appendix_human}

\input{sec/appendix_world}

\FloatBarrier
\input{sec/action_appendix}

\FloatBarrier
\input{sec/evidence_runtime}

\FloatBarrier
\input{sec/evidence_qualitative}
\FloatBarrier
\endgroup

%% file: sec/appendix_renderer.tex
\section{Design of diffusion renderer}
\label{app:renderer}

\begin{figure}[tp]
    \centering
    \includegraphics[width=\linewidth]
        {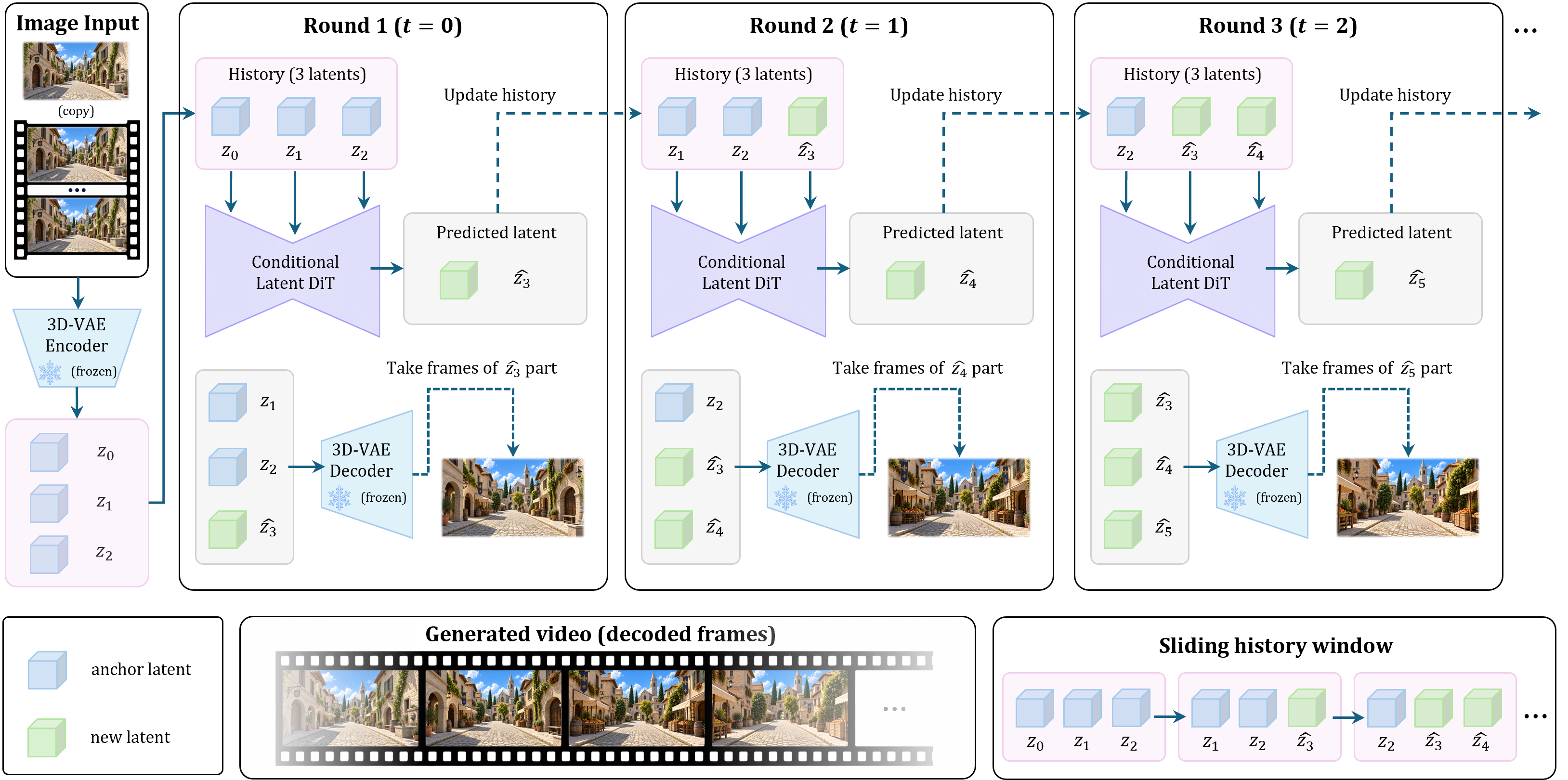}
    \caption{
        \textbf{Autoregressive latent rendering in Khora.} A frozen 3D-VAE encoder maps the initial RGB context into video latents. At each generation step, a conditional latent DiT predicts one new latent from a history of three latents. The prediction is appended to the history after removing its oldest element. The frozen 3D-VAE decoder processes the updated three-latent window. Only the four RGB frames corresponding to the newly predicted latent frame are retained. Blue blocks denote latents encoded from the initial observations, and green blocks denote generated latents. The same DiT and VAE parameters are reused throughout the rollout.
    }
    \label{fig:renderer_details}
\end{figure}

Khora renders each requested view using a conditional latent diffusion transformer \citep{peebles2023dit} and a frozen 3D video VAE \citep{wan2025}. Figure~\ref{fig:renderer_details} illustrates the autoregressive process. The renderer maintains a sliding history of three video latents, predicts one additional latent per generation step, and decodes the updated latent window into RGB observations. All views share the same renderer parameters, while maintaining their own temporal histories.

\paragraph{Latent representation and initialization.}
Let $\mathcal{E}$ and $\mathcal{D}$ denote the frozen 3D-VAE encoder and decoder, respectively. The VAE has a temporal compression factor of four: each newly predicted latent frame corresponds to four RGB frames. The initial RGB context is encoded into video latents, and the first prediction uses the three-latent history

\begin{equation}
    H_2 = [z_0,z_1,z_2].
\end{equation}

Each $z_j$ denotes a temporal latent frame containing spatial features, rather than an individual spatial token. Both the encoder and decoder remain frozen during renderer training.

\paragraph{Shared-world conditioning.}
At each generation step, the DiT receives the latent history together with the target-view condition obtained from the shared world representation. We denote this condition by $c_{n+1}$; it comprises the scene and entity conditions associated with the target camera query, as described in the main method.

The scene condition conveys geometry and available appearance memory, while the entity condition describes the projected dynamic state. The history supplies temporal visual context, and the shared-world condition specifies the scene configuration to be rendered.

\paragraph{Next-latent prediction.}
The renderer generates one new temporal latent conditioned on the preceding three latents and the target-view condition:

\begin{equation}
    \hat z_{n+1}
    \sim
    p_{\phi}\!\left(
        z_{n+1}\mid H_n,c_{n+1}
    \right),
    \qquad n\geq 2,
\end{equation}

where $p_{\phi}$ is implemented by the conditional latent DiT.

For compact notation, define
\begin{equation}
    \bar z_j =
    \begin{cases}
        z_j, & j\in\{0,1,2\},\\
        \hat z_j, & j\geq 3.
    \end{cases}
\end{equation}

The history at step $n$ is then
\begin{equation}
    H_n =
    [\bar z_{n-2},\bar z_{n-1},\bar z_n].
\end{equation}
Thus, the first prediction is conditioned on encoded
observations, whereas subsequent predictions progressively depend on generated latents.

\paragraph{Sliding-window history update.}
After $\hat z_{n+1}$ is predicted, the oldest latent is removed and the new latent is appended:
\begin{equation}
    H_{n+1}
    =
    [\bar z_{n-1},\bar z_n,\hat z_{n+1}].
\end{equation}

The first three updates are
\begin{align}
    [z_0,z_1,z_2]
    &\longrightarrow \hat z_3,\\
    [z_1,z_2,\hat z_3]
    &\longrightarrow \hat z_4,\\
    [z_2,\hat z_3,\hat z_4]
    &\longrightarrow \hat z_5.
\end{align}

After these updates, the history consists entirely of generated latents. No future ground-truth RGB observations are introduced during rollout. Predicted latents are passed directly to the next history window; decoding and re-encoding the generated RGB output are not required for this history update.

\paragraph{Windowed RGB decoding.}
After a new latent frame is predicted, the frozen decoder processes the updated three-latent window:
\begin{equation}
    \widetilde O_{n+1}
    =
    \mathcal{D}\!\left(
        [\bar z_{n-1},\bar z_n,\hat z_{n+1}]
    \right).
\end{equation}
Only the four RGB frames corresponding to the final latent frame $\hat z_{n+1}$ are retained:
\begin{equation}
    \hat O_{n+1}
    =
    \operatorname{Tail}_{4}\!\left(\widetilde O_{n+1}\right),
\end{equation}
where $\operatorname{Tail}_{4}$ selects the last four decoded frames. The preceding two latent frames provide temporal context for decoding, and their corresponding RGB outputs are discarded. Each autoregressive step therefore appends four new frames to the output sequence, without duplicating frames from overlapping decoding windows. The predicted latent is also retained directly in the history for the next generation step.

\paragraph{Temporal history and shared appearance memory.}
The latent history and the STBoard appearance memory serve different purposes. The three-latent history is local to an observation stream and supplies recent temporal context. The appearance memory is associated with the shared scene geometry and can retain information beyond this local window. Decoded observations are back-projected onto the mesh to update the appearance memory, as described in the main method. Subsequent view queries can access the updated scene information through their conditions. This appearance update is distinct from the direct latent-history update.

\paragraph{Extension to multiple views.}
For multiple requested views, the same autoregressive procedure is applied with a separate history $H_n^i$ and target-view condition $c_{n+1}^i$ for each view $i$:
\begin{equation}
    \hat z_{n+1}^i
    \sim
    p_{\phi}\!\left(
        z_{n+1}^i\mid H_n^i,c_{n+1}^i
    \right).
\end{equation}
The renderer parameters and three-latent history length do not change with the number of views. Additional views require additional histories and rendering queries, while scene and entity information are obtained from the shared STBoard. The fixed history bounds the temporal context processed by each rendering query; it does not impose a constant memory cost on the complete multi-view system.

%% file: sec/appendix_setup.tex
\section{Experimental Setup and Visual Quality}
\label{app:setup}

\subsection{Baseline Checkpoints and Preprocessing}
\label{app:baseline_protocol}

For Solaris~\citep{savva2026solaris}, the CS:GO adaptation uses the
Dust2 Stage 2 checkpoint at training iteration 120,000. Its multiplayer configuration
jointly generates two synchronized views with 50 denoising steps. Each view
uses one RGB conditioning frame and 32 subsequent generated frames; the
checkpoint is evaluated only in the two-view setting. Input observations
are resized with aspect-ratio-preserving letterboxing to
360$\times$640. Saved predictions have resolution 352$\times$640, and the
common-window visual-quality scorer resizes them to
360$\times$640 using bilinear interpolation. Dimensions here are height
$\times$ width.

For Solaris, the conditioning observation is physical frame 0 and the
first generated frame corresponds to physical frame 1. Its native
evaluation covers physical frames 1--32; the common-window evaluation
uses frames 13--32 on 10,000 two-view cases.

For Gamma-World~\citep{liu2026gammaworld}, we use the Stage 2 checkpoint at
training iteration 10,000. Its native 320$\times$480 output is resized to the
common 360$\times$640 scoring resolution. The reported values are computed on
the complete 10,000-case two-view evaluation set over frames 38--79.

For the four-view Gamma-World evaluation, we use the same Stage 2 checkpoint
with four agent slots. We evaluate 10,000 cases (40,000 videos), scoring
frames 38--79 after one RGB conditioning frame per view. GT
preprocessing follows the four-view pipeline: resize to 288$\times$512,
normalize and quantize through bfloat16, then bilinearly resize to
360$\times$640.

For MIRA~\citep{hu2026mira}, we use the two-view checkpoint at training step
15,000 for the two-view baseline and the corresponding four-view checkpoint at
training step 3,000 for the four-view baseline. Both evaluations use frames
38--79 after conditioning on 38 RGB frames. MIRA's native 288$\times$512
decoded output is resized to the common 360$\times$640 scoring resolution, and
the paired ground truth is resized from 480$\times$832. MIRA does not provide
an 8-view result in this experiment.

\input{sec/evidence_quality}

%% file: sec/evidence_quality.tex
\subsection{Visual Quality: Evaluation Protocol and Results}
\label{app:quality_sources}

Table~\ref{tab:quality_exact} reports the detailed Khora and input-ablation
results underlying Table~\ref{tab:visual_quality}. The two- and four-view
settings each contain 10,000 cases (20,000 and 40,000 videos); the eight-view
setting contains 159 cases (1,272 videos). All six rows score the 68-frame
suffix, source frames 13--80. Evaluation samples are held out from model
training. Khora scores frames 13--80 after
13 RGB context frames, while MIRA scores frames 38--79 after 38 context
frames. Gamma-World also scores frames 38--79, with one RGB context frame.
Scoring windows therefore differ across models.

\input{sec/evidence_quality_table}

\paragraph{Feature and frame conventions.}
PSNR averages frame-level RGB scores in dB; SSIM~\citep{wang2004ssim} uses an $11\times11$
Gaussian kernel with $\sigma=1.5$; LPIPS v0.1~\citep{zhang2018lpips} uses AlexNet~\citep{krizhevsky2012alexnet}. FID~\citep{heusel2018fid} uses
2,048-dimensional torch-fidelity~\citep{obukhov2020torchfidelity} Inception~\citep{szegedy2016inception} features and one deterministically
selected frame per video. Its local suffix index is
$(1000003s+10007v)\bmod68$, with sample index $s$ and view index $v$.
FVD~\citep{unterthiner2019fvd} uses the 400-dimensional output of the TF-Hub Kinetics-400 I3D
model~\citep{carreira2017i3d}, with local indices
$[0,4,9,13,18,22,27,31,36,40,45,49,54,58,63,67]$.
This feature definition is part of the metric: values from other I3D
representations cannot be compared as if they used the same evaluator.

\begin{figure}[tp]
\centering
\includegraphics[width=\linewidth]{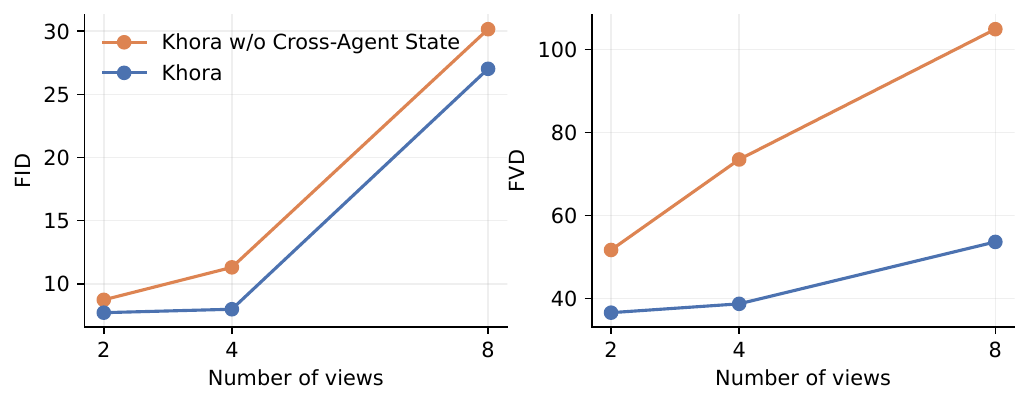}
\caption{\textbf{Visual quality across the available view settings.}
The six points in each panel come directly from Table~\ref{tab:quality_exact}.
Lower is better. Lines connect separate cohorts and are descriptive;
particularly, the eight-view cohort has only 159 cases. No uncertainty bars
are inferred from aggregate FID/FVD values.}
\label{fig:quality_views}
\end{figure}

\paragraph{Detailed baseline results.}
Table~\ref{tab:baseline_exact} reports the Gamma-World and MIRA visual-quality
results to six decimal places. These score 42
frames, 38--79, at $640\times360$, unlike Khora's 68-frame suffix.
Gamma-World conditions on one RGB frame and MIRA on 38. Their FID selects
one frame per video with the analogous deterministic rule modulo 42.
MIRA's two-view evaluation directly resizes decoded GT frames to
$640\times360$ using bilinear interpolation. Its four-view evaluation
first resizes GT frames to $832\times480$, normalizes them to $[-1,1]$
in bfloat16, and then bilinearly resizes them to $640\times360$ before
rounding to 8-bit RGB. These preprocessing differences should be considered
when comparing the two settings.

\input{sec/evidence_baseline_quality}

%% file: sec/evidence_quality_table.tex
\begin{table}[tp]
\centering\setlength{\tabcolsep}{3pt}
\caption{\textbf{Detailed Khora visual-quality results.} Khora and Khora w/o Cross-Agent State share their view-specific cohort. PSNR is in dB; six decimals are shown. The 8-view cohort is substantially smaller.}
\label{tab:quality_exact}
\begin{spreadtabular}{8}{ccN{99999}N{99.999999}N{9.999999}N{9.999999}N{99.999999}N{999.999999}}
\toprule
\tablehead{Views} & \tablehead{Model} & \tablehead{Cases} & \tablehead{PSNR $\uparrow$} & \tablehead{SSIM $\uparrow$} & \tablehead{LPIPS $\downarrow$} & \tablehead{FID $\downarrow$} & \tablehead{FVD $\downarrow$} \\
\midrule
2 & \makecell[l]{Khora w/o\\Cross-Agent State} & 10000 & 26.042689 & 0.712535 & 0.182607 & 8.726227 & 51.665660 \\
\rowcolor{tableStripe}
2 & Khora & 10000 & \bestscore{26.242469} & \bestscore{0.713042} & \bestscore{0.178877} & \bestscore{7.709788} & \bestscore{36.552903} \\
4 & \makecell[l]{Khora w/o\\Cross-Agent State} & 10000 & 24.919565 & 0.701700 & 0.185143 & 11.301036 & 73.491736 \\
\rowcolor{tableStripe}
4 & Khora & 10000 & \bestscore{25.282468} & \bestscore{0.702034} & \bestscore{0.177186} & \bestscore{7.982100} & \bestscore{38.684905} \\
8 & \makecell[l]{Khora w/o\\Cross-Agent State} & 159 & 26.228233 & 0.757253 & 0.154418 & 30.170570 & 104.908220 \\
\rowcolor{tableStripe}
8 & Khora & 159 & \bestscore{26.577453} & \bestscore{0.757305} & \bestscore{0.147736} & \bestscore{27.031112} & \bestscore{53.614786} \\
\bottomrule
\end{spreadtabular}
\end{table}

%% file: sec/evidence_baseline_quality.tex
\begin{table}[tp]
\centering
\caption{\textbf{Detailed baseline visual-quality results.} All rows score source frames 38--79 on 10,000 cases (20,000 videos for two views; 40,000 for four views). PSNR is in dB. These horizons differ from the Khora rows above.}
\label{tab:baseline_exact}
\begin{spreadtabular}{7}{lcN{99.999999}N{9.999999}N{9.999999}N{999.999999}N{999.999999}}
\toprule
\tableheadleft{Method} & \tablehead{Views} & \tablehead{PSNR $\uparrow$} & \tablehead{SSIM $\uparrow$} & \tablehead{LPIPS $\downarrow$} & \tablehead{FID $\downarrow$} & \tablehead{FVD $\downarrow$} \\
\midrule
Gamma-World~\citep{liu2026gammaworld} & 2 & 15.561124 & 0.433187 & 0.560724 & 53.930308 & 297.111511 \\
\rowcolor{tableStripe}
Gamma-World~\citep{liu2026gammaworld} & 4 & 16.947278 & 0.502547 & 0.485552 & 55.875949 & 272.255839 \\
MIRA~\citep{hu2026mira} & 2 & 21.728992 & 0.597260 & 0.392728 & 16.088566 & 231.919817 \\
\rowcolor{tableStripe}
MIRA~\citep{hu2026mira} & 4 & 20.877136 & 0.607222 & 0.356992 & 16.325321 & 198.057638 \\
\bottomrule
\end{spreadtabular}
\end{table}

%% file: sec/appendix_human.tex
\section{Human Evaluation}
\label{app:human}

\subsection{User-Study Protocol}
\label{app:user_study}

The rating interfaces present synchronized multi-view clips in randomized,
model-anonymized order. Participants rate each assigned sample on a five-point
scale, and scores are averaged over the ratings in the cohorts specified below.
All five methods follow the same rating protocol, including the five-point
scale, model-anonymized presentation, and scoring criteria, providing a fair
comparison under shared evaluation criteria. Participant and case counts
for each evaluation cohort are reported below.
The two criteria are defined as follows:

\paragraph{Cross-view action consistency.}
Raters judge whether an action performed by one agent produces temporally and
spatially compatible consequences in the other observations. They consider
actor motion, interaction events, and event timing across synchronized views.
Low scores correspond to contradictions such as an entity moving in one view
but remaining stationary in another, or the same event occurring at different
times.

\paragraph{Cross-view world consistency.}
Raters judge whether shared visible regions agree on static scene geometry,
fixed objects, materials, and spatial relations. Normal differences in
perspective, occlusion, lighting, and field of view are not penalized. Dynamic
agents and their actions are assessed under action consistency instead.

\paragraph{Paired Khora study.}
For Khora and its without-cross-agent-state ablation, each participant
received 150 randomized,
model-blinded samples: 25 source cases for each combination of two model
variants and three view counts (2, 4, and 8). Within each view count, the same
25 source cases were assigned for both variants. The interface synchronized
playback, seeking, and playback speed across the generated views; raters
compared these views with one another, rather than against a GT video.

\subsection{Complete Ratings and Distributions}

Table~\ref{tab:human_consistency_full} reports the human-rating scores
summarized in Table~\ref{tab:multi_agent_consistency}. Rating distributions,
participant and sample counts, and aggregation conventions are detailed
below. External baselines follow first-public-release order.

\begin{table}[htbp]
\centering
\caption{\textbf{Human consistency ratings matching the main table.} Scores use a five-point scale; higher is better.}
\label{tab:human_consistency_full}
\setlength{\tabcolsep}{5pt}
\begin{tabularx}{\linewidth}{LN{9.999}N{9.999}N{9.999}}
\toprule
\tableheadleft{Method} & \tablehead{Action $\uparrow$} & \tablehead{World $\uparrow$} & \tablehead{Overall $\uparrow$} \\
\midrule
Solaris~\citep{savva2026solaris} & 3.000 & \secondscore{4.105} & 3.553 \\
\rowcolor{tableStripe}
Gamma-World~\citep{liu2026gammaworld} & 2.027 & 2.786 & 2.406 \\
MIRA~\citep{hu2026mira} & \secondscore{3.170} & 4.009 & \secondscore{3.589} \\
\rowcolor{tableStripe}
Khora w/o Cross-Agent State & 2.075 & 4.065 & 3.070 \\
\textbf{Khora} & \bestscore{3.570} & \bestscore{4.213} & \bestscore{3.892} \\
\bottomrule
\end{tabularx}
\end{table}

\input{sec/evidence_human}

\subsection{Khora and Ablation: Paired Human Evaluation}
\label{app:human_khora_records}
\input{sec/evidence_human_khora}

%% file: sec/evidence_human.tex
\begin{figure}[t]
\centering
\includegraphics[width=\linewidth]{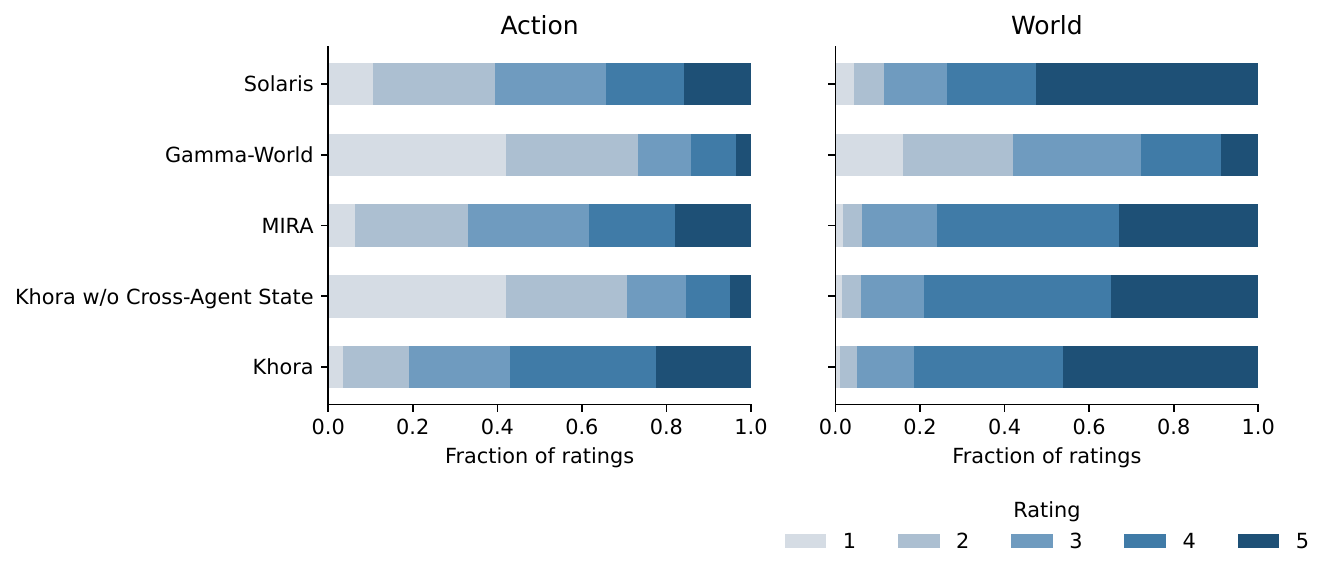}
\caption{\textbf{Distribution of available human ratings.} Each bar sums
to one within a model and criterion. Darker segments denote higher ratings
on the five-point scale. Counts and descriptive means are given in
Table~\ref{tab:human_distribution}; the Khora view-count breakdown appears
in Table~\ref{tab:human_khora_distribution}.
All methods use the same five-point rating protocol. Solaris, Gamma-World,
and MIRA have 114, 112, and 112 ratings from 12 participants; Khora and its
ablation each have 600 ratings from eight participants who completed all
assignments. The panels do not imply independent ratings or statistical significance.}
\label{fig:human_distributions}
\end{figure}

\paragraph{Human-rating distributions.}
The baseline evaluation includes 338 submitted ratings
from 12 participants: 114 for Solaris and 112 each for Gamma-World
and MIRA, covering ten source cases per model. Table~\ref{tab:human_distribution}
and Figure~\ref{fig:human_distributions} include these records together with
600 ratings each for Khora and its ablation, whose participant selection and
view-count breakdown are described below. Only submitted ratings are included. These are
repeated observations within participants and cases; an ordinary
independent-sample standard error would be inappropriate.

\input{sec/evidence_human_table}

Table~\ref{tab:human_distribution} reports criterion means and standard
deviations to six decimal places. The main table rounds each criterion mean
to three decimals and computes Overall from the unrounded criterion means.
Khora and ablation ratings and sample counts are further broken down by
view count in Section~\ref{app:human_khora_records}.

%% file: sec/evidence_human_table.tex
\begin{table}[tp]
\centering\setlength{\tabcolsep}{3pt}
\caption{\textbf{Human-rating distributions for all five models.} Baselines have 12 raters and 10 source cases each. For Khora and its ablation, eight participants completed all assignments, yielding 600 ratings per model, pooled equally over 2, 4, and 8 views. Columns 1--5 give exact counts; SD is descriptive. Mean ranks compare methods within each criterion.}
\label{tab:human_distribution}
\begin{tabularx}{\linewidth}{LcN{999}N{9.999999}N{9.999999}N{999}N{999}N{999}N{999}N{999}}
\toprule
\tableheadleft{Method} & \tablehead{Criterion} & \tablehead{$n$} & \tablehead{Mean $\uparrow$} & \tablehead{SD} & \tablehead{1} & \tablehead{2} & \tablehead{3} & \tablehead{4} & \tablehead{5} \\
\midrule
Solaris~\citep{savva2026solaris} & Action & 114 & 3.000000 & 1.240896 & 12 & 33 & 30 & 21 & 18 \\
\rowcolor{tableStripe}
Solaris~\citep{savva2026solaris} & World & 114 & \secondscore{4.105263} & 1.162606 & 5 & 8 & 17 & 24 & 60 \\
Gamma-World~\citep{liu2026gammaworld} & Action & 112 & 2.026786 & 1.142621 & 47 & 35 & 14 & 12 & 4 \\
\rowcolor{tableStripe}
Gamma-World~\citep{liu2026gammaworld} & World & 112 & 2.785714 & 1.188750 & 18 & 29 & 34 & 21 & 10 \\
MIRA~\citep{hu2026mira} & Action & 112 & \secondscore{3.169643} & 1.192230 & 7 & 30 & 32 & 23 & 20 \\
\rowcolor{tableStripe}
MIRA~\citep{hu2026mira} & World & 112 & 4.008929 & 0.925081 & 2 & 5 & 20 & 48 & 37 \\
\makecell[l]{Khora w/o\\Cross-Agent State} & Action & 600 & 2.075000 & 1.192369 & 253 & 171 & 84 & 62 & 30 \\
\rowcolor{tableStripe}
\makecell[l]{Khora w/o\\Cross-Agent State} & World & 600 & 4.065000 & 0.899327 & 9 & 27 & 89 & 266 & 209 \\
Khora & Action & 600 & \bestscore{3.570000} & 1.103235 & 21 & 93 & 144 & 207 & 135 \\
\rowcolor{tableStripe}
Khora & World & 600 & \bestscore{4.213333} & 0.901393 & 7 & 23 & 82 & 211 & 277 \\
\bottomrule
\end{tabularx}
\end{table}

%% file: sec/evidence_human_khora.tex
\paragraph{Participants and evaluation cohort.}
Fourteen participants submitted 1,273 ratings. The analysis uses the eight
participants who completed all 150 assignments, yielding 1,200 ratings:
600 per model, with 200 per view count. The other six participants submitted
73 ratings in total; these partial responses are excluded from the statistics
below.
Each rating contains both an Action and a World score.

The completed assignments cover 198 distinct two-view source cases, 196
four-view cases, and 113 eight-view cases, identically paired across model
variants. Thus, 600 ratings per model correspond to 507 distinct
view-count/source-case combinations, with some cases rated by multiple
participants. Table~\ref{tab:human_distribution} reports pooled statistics
for all five models, while Table~\ref{tab:human_khora_distribution} gives
Khora and ablation category counts, means, and sample standard deviations
by view count. Their pooled means weight the three view counts equally
because each contributes 200 ratings.

\input{sec/evidence_human_khora_table}

\paragraph{Results and aggregation.}
Khora has Action $3.570000$ and World $4.213333$, which round to $3.570$
and $4.213$. Their arithmetic mean is $3.891667$, reported as $3.892$.
The ablation has Action $2.075000$ and World $4.065000$, giving Overall
$3.070000$.
Figure~\ref{fig:human_distributions} visualizes the same 600 ratings
per model. Participants and source cases contribute repeated observations,
so the descriptive SDs and rating distributions do not establish statistical
significance or an independent-sample confidence interval.

%% file: sec/evidence_human_khora_table.tex
\begin{table}[tp]
\centering\setlength{\tabcolsep}{3pt}
\caption{\textbf{Khora human ratings by view count.} Eight participants who completed all assignments collectively contribute $n=200$ ratings per model and view count. Columns 1--5 give exact counts; SD is descriptive. Bold marks the higher mean within each view count and criterion.}
\label{tab:human_khora_distribution}
\begin{tabularx}{\linewidth}{LccN{9.999999}N{9.999999}N{999}N{999}N{999}N{999}N{999}}
\toprule
\tableheadleft{Method} & \tablehead{Views} & \tablehead{Criterion} & \tablehead{Mean $\uparrow$} & \tablehead{SD} & \tablehead{1} & \tablehead{2} & \tablehead{3} & \tablehead{4} & \tablehead{5} \\
\midrule
\makecell[l]{Khora w/o\\Cross-Agent State} & 2 & Action & 2.095000 & 1.250517 & 87 & 54 & 25 & 21 & 13 \\
\rowcolor{tableStripe}
\makecell[l]{Khora w/o\\Cross-Agent State} & 2 & World & 4.060000 & 1.015656 & 5 & 12 & 31 & 70 & 82 \\
\makecell[l]{Khora w/o\\Cross-Agent State} & 4 & Action & 2.085000 & 1.263231 & 87 & 58 & 21 & 19 & 15 \\
\rowcolor{tableStripe}
\makecell[l]{Khora w/o\\Cross-Agent State} & 4 & World & 4.085000 & 0.813069 & 0 & 9 & 31 & 94 & 66 \\
\makecell[l]{Khora w/o\\Cross-Agent State} & 8 & Action & 2.045000 & 1.057624 & 79 & 59 & 38 & 22 & 2 \\
\rowcolor{tableStripe}
\makecell[l]{Khora w/o\\Cross-Agent State} & 8 & World & 4.050000 & 0.860933 & 4 & 6 & 27 & 102 & 61 \\
\midrule
Khora & 2 & Action & \bestscore{3.825000} & 1.136147 & 11 & 15 & 38 & 70 & 66 \\
\rowcolor{tableStripe}
Khora & 2 & World & \bestscore{4.245000} & 0.989886 & 5 & 9 & 22 & 60 & 104 \\
Khora & 4 & Action & \bestscore{3.530000} & 1.036568 & 3 & 34 & 56 & 68 & 39 \\
\rowcolor{tableStripe}
Khora & 4 & World & \bestscore{4.185000} & 0.880141 & 1 & 8 & 32 & 71 & 88 \\
Khora & 8 & Action & \bestscore{3.355000} & 1.088531 & 7 & 44 & 50 & 69 & 30 \\
\rowcolor{tableStripe}
Khora & 8 & World & \bestscore{4.210000} & 0.830269 & 1 & 6 & 28 & 80 & 85 \\
\bottomrule
\end{tabularx}
\end{table}

%% file: sec/appendix_world.tex
\section{Cross-View World Consistency}
\label{app:geometry_metric}

Our scene-level scores combine point-cloud geometry and color, as in the
CVC formulation in MetaWorld's Appendix~B.2 \citep{hu2026metaworld}.
We refer specifically to that point-cloud definition, not a feature-similarity
score. The two protocols below specify our input pairings, reconstruction,
registration, and aggregation; neither is a reproduction of MetaWorld's
evaluation protocol.

\subsection{CVC-GT: Measurement Protocol}
\label{app:cvc_gt}

CVC-GT is a GT-referenced scene-level proxy: generated and GT videos are reconstructed
separately for each ego view, and the resulting point clouds are aligned and
compared before averaging the two view scores. It does not directly register
one generated view to another. Depth Anything V2 Small~\citep{yang2024depthanythingv2} estimates depth;
temporal ICP~\citep{besl1992icp} accumulates each sequence, and point-cloud alignment precedes
geometry and color comparison. The pipeline uses native output resolution,
a 32-frame scoring window sampled every four frames (eight sampled frames
per view), and image pixel stride eight. CVC, F-score, ColorSim, precision,
and recall are averaged over valid cases. Baseline evaluations exclude
videos rejected by their quality checks, so their means are conditional on
the reported valid sets.

Solaris uses 100/100 valid/attempted cases, Gamma-World uses
9,988/9,988, and MIRA uses 100/102. Khora and its ablation use a paired
1,000/1,000-case evaluation with the same checkpoint and five denoising
steps. Their sampled relative frames are 13, 17, 21, 25, 29, 33, 37, and 41.
Baseline scoring windows start at saved-video frame
1 for Solaris and Gamma-World and 38 for MIRA. The action evaluation uses
explicit source-frame mapping and separate comparison windows.

\subsection{CVC-GT: Complete Results}

Table~\ref{tab:native_cvc_full} reports all components of the geometry-aware
scores summarized in Table~\ref{tab:multi_agent_consistency}.

\begin{table}[tp]
\centering
\caption{\textbf{Complete CVC-GT results.} CVC and its components are GT-referenced geometric proxies with the model-specific protocols described above. Higher is better; six decimals are shown. ColorSim denotes color similarity.}
\label{tab:native_cvc_full}
\setlength{\tabcolsep}{3.2pt}
\begin{tabularx}{\linewidth}{LN{9.999999}N{9.999999}N{9.999999}N{9.999999}N{9.999999}}
\toprule
\tableheadleft{Method} & \tablehead{CVC-GT $\uparrow$} & \tablehead{F-score $\uparrow$} & \tablehead{ColorSim $\uparrow$} & \tablehead{Precision $\uparrow$} & \tablehead{Recall $\uparrow$} \\
\midrule
Solaris~\citep{savva2026solaris} & 0.918310 & 0.940273 & 0.896348 & 0.956918 & 0.931426 \\
\rowcolor{tableStripe}
Gamma-World~\citep{liu2026gammaworld} & 0.897661 & 0.927937 & 0.867384 & 0.940621 & 0.923025 \\
MIRA~\citep{hu2026mira} & 0.922486 & 0.942951 & 0.902021 & 0.952500 & 0.940217 \\
\rowcolor{tableStripe}
Khora w/o Cross-Agent State & \secondscore{0.931900} & \secondscore{0.959991} & \secondscore{0.903809} & \secondscore{0.963585} & \secondscore{0.959753} \\
\textbf{Khora} & \bestscore{0.932200} & \bestscore{0.960586} & \bestscore{0.903815} & \bestscore{0.964424} & \bestscore{0.960035} \\
\bottomrule
\end{tabularx}
\end{table}

\input{sec/appendix_xview}

%% file: sec/appendix_xview.tex
\subsection{CVC-XView: Generated-View Comparison}
\label{app:generated_world}

CVC-XView compares generated view 0 with generated view 1 from the same
case and model, without GT images, GT depth, or GT point clouds. It measures
scene-level agreement between reconstructed generated observations, whereas
CVC-GT measures agreement with a reference observation. Both retain the
geometric and color components of the CVC implementation; the reconstruction,
normalization, direction averaging, and cohorts differ, so their absolute
scores must not be pooled or interpreted as a controlled protocol ablation.

\paragraph{Evaluation cohort.}
All five models are evaluated on the same 10,000 held-out cases, with two
generated views per case. No cases are excluded based on generated image
quality or scores, including low-motion and spatially flat outputs. All
cases are successfully scored for every model, and results are averaged
over this complete cohort rather than the GT-filtered 1,643-case cohort
used for action evaluation.

\paragraph{Frame mapping and reconstruction.}
Each model uses a 32-frame generated window, sampled every four
frames, giving eight RGB frames per view. All inputs are resized to
$640\times360$. Physical source-frame indices are 2:4:30 for
Solaris~\citep{savva2026solaris}, 1:4:29 for
Gamma-World~\citep{liu2026gammaworld}, 38:4:66 for MIRA~\citep{hu2026mira},
and 13:4:41 for Khora and its input ablation; $a:d:b$ denotes start, stride,
and last sampled frame. The two views of a case use matching physical times,
but the model families do not share a common physical scoring interval.

Depth Anything V2 Small~\citep{yang2024depthanythingv2} predicts relative
disparity $q$. With $m$ the median positive disparity in that frame, we use
$d=m/\max(q,0.05m)$ and retain depths up to eight. Back-projection assumes
a $90^\circ$ horizontal field of view and samples image pixels with stride
eight. Each view is reconstructed independently by accumulating its sampled
frames using PCA-initialized, point-to-point ICP~\citep{besl1992icp}, with
correspondence distance 0.4 and at most 20 iterations. Each accumulated
cloud is centered at its bounding-box midpoint and divided by its largest
extent. The two clouds are then aligned using PCA initialization followed
by at most 20 iterations of point-to-point ICP, with correspondence distance 0.2.
This differs from the FPFH-based global registration and point-to-plane ICP
described in MetaWorld's Appendix~B.2 \citep{hu2026metaworld}; our scoring
also independently registers both directions, as specified below.

\paragraph{Components and aggregation.}
For each registration direction, let $A$ be the aligned source cloud and
$B$ the reference cloud. Precision is the fraction of points in $A$ whose
nearest neighbor in $B$ is within $\tau=0.05$; recall reverses the point sets.
Their harmonic mean is the F-score, set to zero when precision and recall
sum to zero. ColorSim averages
$1-\|c_A-c_B\|_2/\sqrt{3}$ over source-to-reference nearest-neighbor pairs
within $\tau$, with RGB colors in $[0,1]$; it is zero if no pair matches.
For each direction, $\mathrm{CVC}=\tfrac12(\mathrm{F\mbox{-}score}+\mathrm{ColorSim})$.
We independently register and score both directions, then average each
component over directions and equally over the common cases. All five
reported components lie in $[0,1]$. F-score is computed before aggregation,
so its reported mean need not equal the harmonic mean of mean precision
and mean recall.

\subsection{CVC-XView: Complete Results and Scope}
\label{app:cvc_xview_results}

Table~\ref{tab:cvc_xview_full} contains all five mean scores underlying the
CVC-XView column of Table~\ref{tab:multi_agent_consistency}. These are
descriptive means; differences do not by themselves establish statistical
significance.

\input{sec/cvc_xview_table}

\paragraph{Interpretation and specificity.}
Independent scale normalization and unconstrained registration can align
similar backgrounds while concealing incorrect pose, scale, or moving
entities. There is no verified true-overlap mask or semantic identity
matching. Consequently, this metric is a registration-based scene proxy;
it does not establish synchronized actions or character identity.
Table~\ref{tab:world_controls} reports a ten-case specificity
control: replacing view 1 with the next case's view 1 can still produce
high scores. These controls use a separate ten-case cohort and must not be
mistaken for the 10,000-case model means. The GT-referenced action
metrics in Appendix~\ref{app:action} provide complementary actor evidence.

\input{sec/evidence_world_controls}

%% file: sec/cvc_xview_table.tex
\begin{table}[tp]
\centering\setlength{\tabcolsep}{3pt}
\caption{\textbf{Complete CVC-XView results.} All five models are evaluated on the same 10,000 cases. Higher is better; six decimals are shown. Physical time windows differ across model families.}
\label{tab:cvc_xview_full}
\begin{tabularx}{\linewidth}{LN{9.999999}N{9.999999}N{9.999999}N{9.999999}N{9.999999}}
\toprule
\tableheadleft{Method} & \tablehead{CVC-XView $\uparrow$} & \tablehead{F-score $\uparrow$} & \tablehead{ColorSim $\uparrow$} & \tablehead{Precision $\uparrow$} & \tablehead{Recall $\uparrow$} \\
\midrule
Solaris~\citep{savva2026solaris} & 0.836662 & \secondscore{0.841934} & 0.831390 & \secondscore{0.904104} & \secondscore{0.810266} \\
\rowcolor{tableStripe}
Gamma-World~\citep{liu2026gammaworld} & 0.836490 & \bestscore{0.870085} & 0.802894 & \bestscore{0.919766} & \bestscore{0.843761} \\
MIRA~\citep{hu2026mira} & 0.830889 & 0.822776 & \bestscore{0.839002} & 0.893388 & 0.788351 \\
\rowcolor{tableStripe}
Khora w/o Cross-Agent State & \secondscore{0.838207} & 0.838615 & \secondscore{0.837798} & 0.902333 & 0.806227 \\
\textbf{Khora} & \bestscore{0.838773} & 0.839772 & 0.837774 & 0.903197 & 0.807770 \\
\bottomrule
\end{tabularx}
\end{table}

%% file: sec/evidence_world_controls.tex
\begin{table}[tp]
\centering\setlength{\tabcolsep}{4pt}
\caption{\textbf{Cross-case mismatch control for generated-view CVC.} Ten cases per model. Gamma-World~\citep{liu2026gammaworld} is compared with Khora. A high mismatched score exposes limited specificity of registration-based scene proxies.}
\label{tab:world_controls}
\begin{tabularx}{\linewidth}{LcN{99}N{9.999999}N{9.999999}N{9.999999}N{9.999999}N{9.999999}}
\toprule
\tableheadleft{Method} & \tablehead{Pairing} & \tablehead{$n$} & \tablehead{CVC $\uparrow$} & \tablehead{F-score $\uparrow$} & \tablehead{ColorSim $\uparrow$} & \tablehead{Precision $\uparrow$} & \tablehead{Recall $\uparrow$} \\
\midrule
Gamma-World & paired & 10 & 0.835207 & 0.871595 & 0.798820 & 0.927648 & 0.833106 \\
\rowcolor{tableStripe}
Gamma-World & mismatch & 10 & 0.797549 & 0.843013 & 0.752084 & 0.903676 & 0.802885 \\
\textbf{Khora} & paired & 10 & 0.850039 & 0.866512 & 0.833567 & 0.928074 & 0.838546 \\
\rowcolor{tableStripe}
\textbf{Khora} & mismatch & 10 & 0.832969 & 0.862446 & 0.803492 & 0.896035 & 0.843299 \\
\bottomrule
\end{tabularx}
\end{table}

%% file: sec/action_appendix.tex
\section{Cross-View Action Consistency}
\label{app:action}

\subsection{Evaluation Protocol and Scope}
\label{app:action_protocol}

\subsubsection{Evaluation object and generation conditions}
The quantitative action evaluation measures the other synchronized player's
observable location, image motion, body configuration, and appearance in one
generated ego view. Dataset identity and visibility associate the reference
player; the evaluator does not compare two generated player tracks directly.
This is a GT-referenced operationalization of cross-view action consistency.
It does not recognize jump, crouch, fire, or reload events, and source action
labels are never substituted for generated-video predictions.

Khora and the ablation use
the same renderer checkpoint, seed, five denoising steps, initial 13 RGB
frames, and video encoding. The ablation clears the non-ego actor geometry
and state condition maps and pads the non-ego player-model-ID map at
inference; scene conditions, ego conditions, and RGB history are retained.
Thus the history can still show other players. The ablated model is not
separately retrained. This evaluation uses recorded camera poses and supplied
state trajectories to measure state-conditioned rendering, rather than
autonomous prediction of future states.

\subsubsection{Cohort construction}
We apply the eligibility criteria to all 10,000 cases in the fixed
two-player evaluation set, which is disjoint from the training set.
One deterministic ego view is selected per case:
the first view passing the metadata and geometric checks is tested for GT
detectability, without falling back to the second view if that test fails.
Both players must be alive over frames 13--57, the ego camera must be
unscoped, the target must be visible at all 40 scored frames, and no third
player may be visible in those frames. The target's projected body axis must
remain in front of the camera, within the image margins, and have height
28--250 pixels at $640\times360$. Its ground-plane displacement between the
endpoints of each window must be at least eight source-world units. A GT
person detection associated with the projected target is required in every
frame of both windows. Selection does not use generated images or scores.

Of the 10,000 cases, 7,541 fail the metadata/geometric checks, 816 fail the
all-frame GT detection check, and 1,643 complete all available model/window
evaluations, with no processing errors. This detector-observable,
single-visible-target cohort is 16.43\% of the source cases. It does not
evaluate invisible-target hallucinations, multi-person identity switching,
or all source trajectories without selection. The selected cases come from
1,466 source episodes. Every model/window contains 32,860 scored frames.

\subsubsection{Time and spatial mapping}
All indices are zero-based physical source-frame indices at 32 frames/s,
not timestamps inferred from the encoded video's playback rate. A relative
frame $t$ maps to GT frame $b+t$, where $b$ is the case's start frame.
For Solaris, generated-frame index $t-1$ corresponds to physical frame
$t$; for the other models, generated-frame index $t$ corresponds to
physical frame $t$. We evaluate frames 13--32 for Solaris, Gamma-World,
Khora, and its ablation, and frames 38--57 for MIRA, Gamma-World, Khora, and
its ablation. Solaris ends before MIRA's 38-frame input prefix ends, so no
common purely generated interval exists across all five models. Both
windows contain 20 frames; they are short-window evaluations, not evidence
of long-horizon stability. Equal physical frames within a panel do not
equalize input history or state information across model families.

\begin{table}[tp]
\centering\caption{\textbf{Generation settings for action evaluation.}
Diffusion steps describe the evaluated videos; model-family algorithms differ.
Khora and its input ablation share all settings except other-agent conditions.}
\label{tab:action_video_configs}
\begin{tabularx}{\linewidth}{LN{99}N{99}cc}
\toprule
\tableheadleft{Method} & \tablehead{RGB\\prefix} & \tablehead{Diffusion\\steps} & \tablehead{Native $W\times H$} & \tablehead{Scored\\frames} \\
\midrule
Solaris~\citep{savva2026solaris} & 1 & 50 & $640\times352$ & 13--32 \\
\rowcolor{tableStripe}
Gamma-World~\citep{liu2026gammaworld} & 1 & 35 & $480\times320$ & Both windows \\
MIRA~\citep{hu2026mira} & 38 & 50 & $512\times288$ & 38--57 \\
\rowcolor{tableStripe}
Khora w/o Cross-Agent State & 13 & 5 & $640\times360$ & Both windows \\
\textbf{Khora} & 13 & 5 & $640\times360$ & Both windows \\
\bottomrule
\end{tabularx}
\end{table}

Scoring uses $640\times360$ images. Solaris is placed in rows 4--355 with
four pixels of vertical padding on either side; other native outputs and GT
are resized using area interpolation. GT scoring regions are clipped to the
shared support $0\leq x<640$, $4\leq y<356$. No spatial registration,
best-matching generated crop, or temporal warping is applied to action scores.

\begin{figure}[tp]
  \centering
  \includegraphics[width=\linewidth]{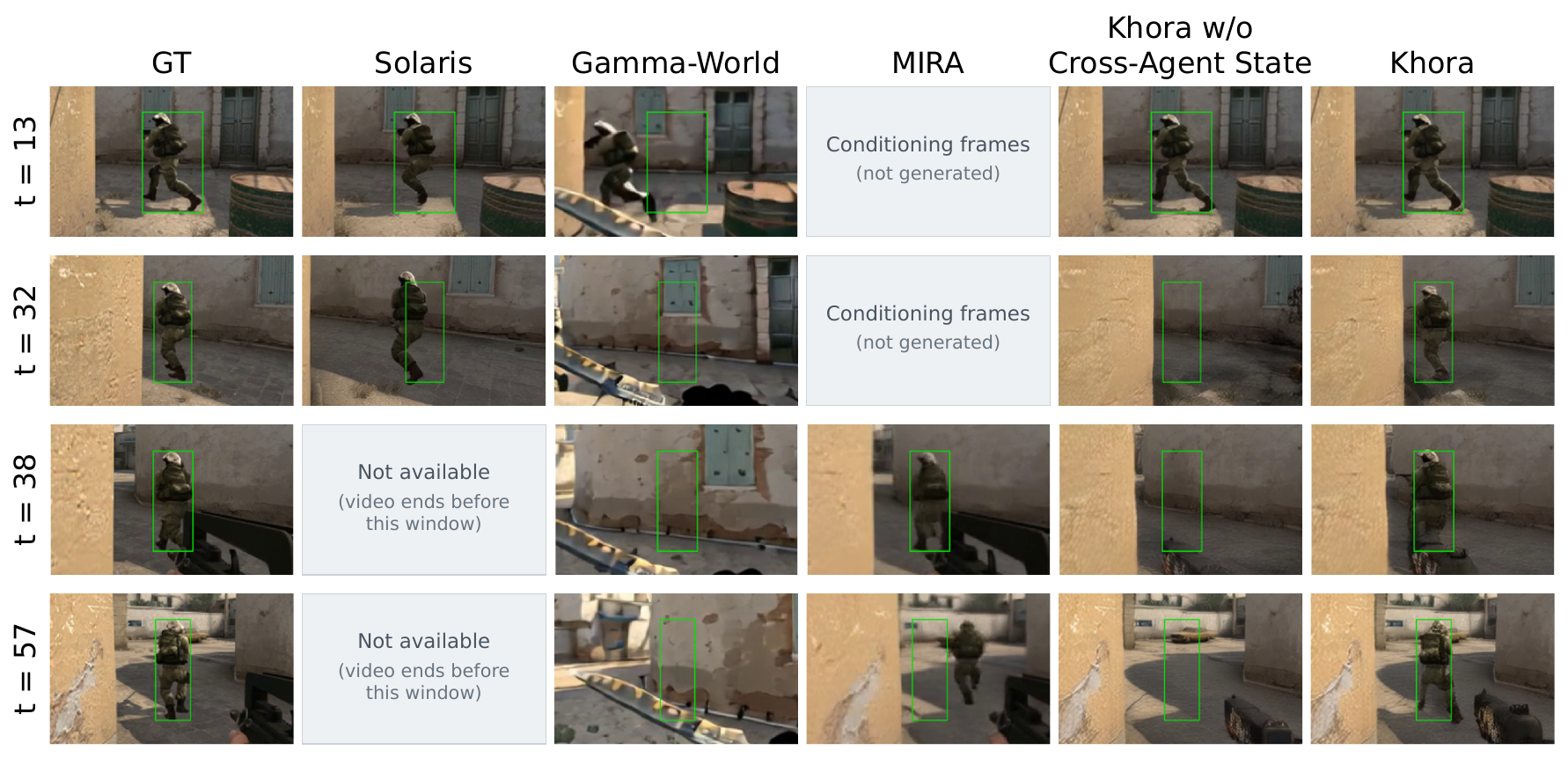}
  \caption{\textbf{Cross-view action consistency in fixed GT regions.}
  Case 79 is an example from the sampled inspection set. All methods
  use identical GT-defined crops and green reference boxes without generated
  recentering. Endpoint frames of each window are shown; numerical scores
  use all 20 frames at the original scoring resolution. Solaris is available
  only in the early window and MIRA only in the late window.
  All rows retain GT and five fixed model columns; gray cells identify
  conditioning frames or video coverage unavailable for generation scoring.}
  \label{fig:action_gt_regions}
\end{figure}

\subsubsection{GT pseudo-labels and detections}
We use Ultralytics 8.3.203~\citep{jocher2023ultralytics} with YOLOv8x detection, YOLOv8x-pose, and
YOLOv8x-seg weights, inference size 960, FP32, and batch size four. Dataset
player identity, world coordinates, rendering transform, and ego camera
parameters provide a projected body axis to associate GT detections.
The boxes, foreground masks, and 2D skeletons themselves are image-derived
pseudo-labels, not native renderer masks or annotated skeletons. Detection
confidence is 0.25 and NMS IoU is 0.6. The same fixed geometric filter is
used for GT and generated detections: width at least five pixels, height at
least 20 pixels, width/height at most 1.35, and exclusion of boxes with bottom
at least 350 and top at least 216 to suppress first-person arms.

For the foreground, we run segmentation only on GT. A person mask must have
confidence at least 0.25 and mask-box IoU at least 0.3 with the associated
GT detection. Its intersection with the fixed GT box must contain at least
32 pixels and cover at least 10\% of that box. Otherwise its foreground
score is unavailable; the box-region score is still computed. Valid
foreground coverage is 32,754/32,860 frames in the early window and
32,778/32,860 in the late window, identical for all methods within a window.

\begin{figure}[tp]
\centering
\includegraphics[width=\linewidth]{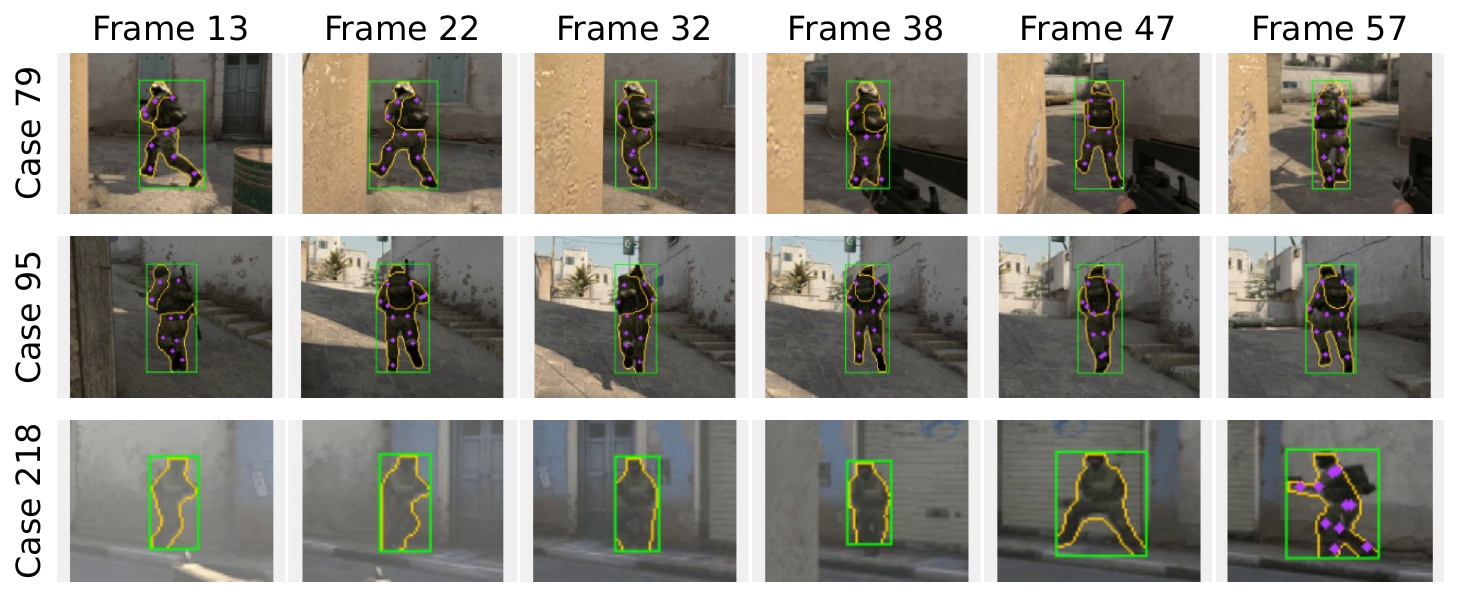}
\caption{\textbf{GT box, foreground, and pose diagnostics.} Cases 79, 95,
and 218 are sampled evaluation cases.
Green outlines denote the fixed GT boxes; yellow contours denote foreground
masks; magenta points denote body joints with confidence at least 0.5.
The same three frames from each window are shown for every case.
Crops are enlarged without altering the pseudo-labels.}
\label{fig:gt_region_diagnostics}
\end{figure}

Figure~\ref{fig:gt_region_diagnostics} shows three sampled cases with the image-derived reference
regions. The contours and keypoints illustrate the pseudo-label support
used by the evaluator, rather than manually verified segmentation or pose.

\subsubsection{Detection, position, and tracking}
For the single GT target, a frame contributes one TP if the largest retained
person-box IoU is at least 0.5. All other retained detections count as FP,
and an unmatched GT target counts as FN. Precision, recall, and F1 pool these
counts over cases. Mean best IoU includes missing frames with value zero.
The separate single-target tracker initializes using the nearest eligible
generated detection within $\max(60,h)$ pixels of the GT center, where $h$
is GT box height. Initialization can occur after the first frame; after the
first association it uses only the last generated box, with no GT updates
or interpolation of missing frames.

Let $D=\sqrt{640^2+360^2}$, and let $e_t$ be the center error divided by $D$.
Center PCK counts $e_t\leq0.05$, with missing tracks incorrect. Position AUC
is the mean of $\max(0,1-e_t/0.1)$, with missing tracks assigned zero. ADE
averages errors over available tracked frames in each case, and FDE uses
the last frame only when a track is available there. The word ``available''
does not imply a correct IoU match. Their missing-penalized counterparts
assign missing frames one image diagonal and cap available errors at that
value. Normalized errors are reported as percentages of $D$. Track coverage,
track breaks, and excess detections per frame accompany the conditional
errors to expose survivorship effects.

\subsubsection{Image motion and body pose}
Motion pairs are four frames apart. A frame pair is classified as moving if
the GT center displacement is at least three pixels. Direction is correct when the generated displacement
is also at least three pixels and its cosine with GT is at least 0.5;
missing tracks are incorrect. Direction accuracy pools correct counts over
GT-moving pairs, while image move/stop accuracy pools all pairs. The
available-track cosine is a separate conditional statistic. Motion includes
ego-camera motion and is not world-space trajectory accuracy.
Motion-pair coverage is the fraction of frame pairs with an available
generated track at both endpoints. Higher coverage indicates greater track
availability, not necessarily correct motion, and is read alongside direction
and move/stop accuracy. Tracking and generated-joint coverage are ranked as
model-dependent statistics; fixed GT box and foreground coverage describe
the evaluation regions and are not ranked.

Pose scoring uses COCO~\citep{lin2014coco} joints 5--16 (shoulders through ankles), retaining GT
joints with confidence at least 0.5. A generated joint must also have
confidence at least 0.5 and lie within $0.2h$ of the GT joint for absolute
PCK; unavailable predictions count as incorrect. PCK pools correct and
eligible GT joints. Root-normalized shape PCK separately subtracts each
pose's hip center and divides by its own pose-box height, with threshold
0.2; it requires confident GT hips and treats unavailable predicted hips as
incorrect. This supplementary score removes location and scale information.

\subsubsection{Fixed-region pixel metrics}
For a GT box or valid GT foreground $R_t$, we compute
\begin{equation}
  \mathrm{MAE}_t=\frac{1}{3|R_t|}\sum_{p\in R_t}\sum_{c=1}^3
  \frac{|I^{\mathrm{gen}}_{t,p,c}-I^{\mathrm{GT}}_{t,p,c}|}{255},
  \qquad S_t=1-\mathrm{MAE}_t.
\end{equation}
Table~\ref{tab:cross_view_action} reports foreground MAE (lower is better).
The complete tables and diagnostic controls also report the equivalent
similarity $S_t=1-\mathrm{MAE}_t$. Its bootstrap interval $[L,U]$
becomes $[1-U,1-L]$ for MAE; paired Khora-minus-ablation differences change
sign, with interval $[-U,-L]$.
The same GT-defined region is used for every generated model, including
frames with no detected generated person. MSE uses squared normalized RGB
differences, and PSNR is $\min(100,-10\log_{10}(\max(\mathrm{MSE},10^{-10})))$.
Exact-match frame counts are retained. SSIM~\citep{wang2004ssim} uses an $11\times11$ Gaussian
window with $\sigma=1.5$, population variance, and constants $0.01^2$ and
$0.03^2$. RGB-channel SSIM maps are averaged and pooled only inside the GT
region; the local SSIM window includes up to five context pixels outside it.
SSIM retains its raw range, including negative values.
The color score $1-\mathrm{MAE}$ is sensitive to
brightness and can be high even when the person is absent.

\subsubsection{Aggregation and uncertainty}
Scalar errors and pixel scores first average valid frames within a case and
then average valid cases equally. A PSNR aggregate therefore averages frame
PSNR values; it is not PSNR of the aggregate MSE. Detection P/R/F1, direction,
move/stop accuracy, pose PCK, and joint coverage pool their appropriate
counts. Counts are summed, except track breaks, which are a per-case mean.
The complete tables also report available-case denominators,
across-case sample standard deviations (with $n-1$ denominator), and
supplementary frame-weighted pixel similarity and SSIM. A case SD describes
the distribution of per-case ratios, even when the primary statistic pools
counts; it is not a standard error of that pooled statistic.

We compute 2,000 paired bootstrap~\citep{efron1979bootstrap} replicates (seed 20260925), sampling the
1,466 source episodes with replacement and retaining all selected cases
from each sampled episode. Each episode contributes at most four cases.
The same resamples are used for all models and windows. Ratio statistics
are recomputed from resampled counts. Tables~\ref{tab:action_ci}
and~\ref{tab:action_paired_ci} give percentile 95\% intervals for the main
scores and Khora-minus-ablation differences. These intervals quantify
sampling uncertainty conditional on this cohort; they do not account for
pseudo-label bias or model-training randomness.

\subsubsection{Metric controls and limitations}
Identity, missing-person, spatial-shift, time-reversal, and frozen-frame
controls check metric behavior. A black-image control can
still yield foreground $1-\mathrm{MAE}=0.864479$ while foreground
SSIM is 0.013803, illustrating why pixel similarity cannot establish actor
presence. The control is illustrative, not an aggregate baseline.
We also visually inspect 600 GT crops (three frames in each window for 100
cases sampled uniformly without replacement, seed 20260925), including box,
mask, and confident-joint overlays. A 20-case subset additionally supplies
160 generated midpoint crops across all available
model/window combinations. Small targets often lack confident body joints,
and segmentation boundaries can include equipment or omit fine limbs.
These visual diagnostics are not an independent human annotation study or
an estimate of annotator agreement.
Detector and pose-estimator domain bias remains,
especially for small, blurred, or stylized generated people. Semantic
jump/crouch/fire/reload accuracy, multi-target identity metrics, and
world-space motion accuracy are not computed and have no numerical entries.

\begin{figure}[tp]
  \centering
  \includegraphics[width=\linewidth]{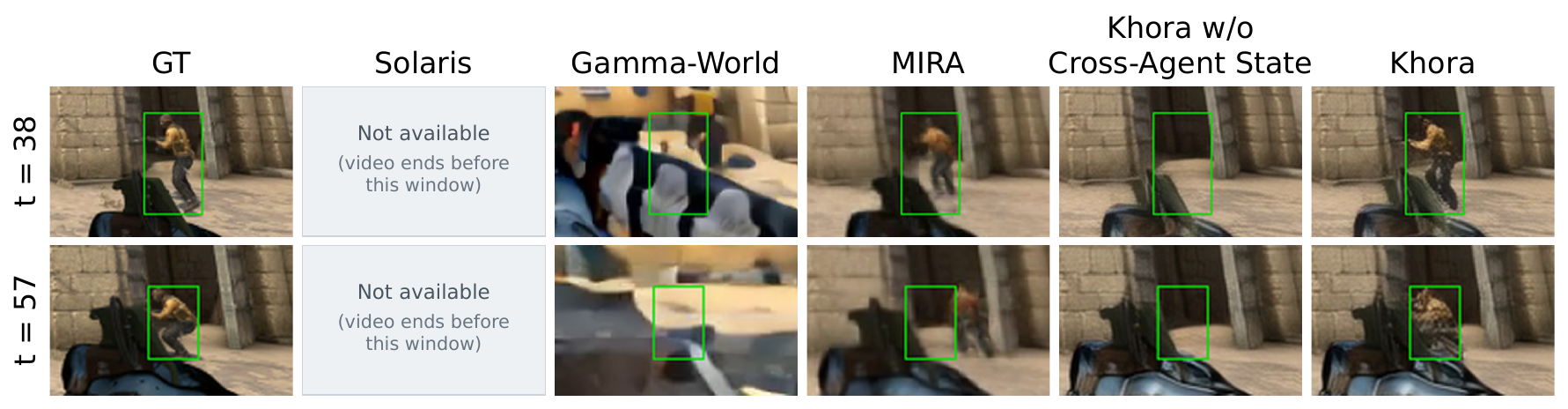}
  \caption{\textbf{A visible actor missed by the detector.}
  Case 3629 is selected because it has the lowest late-window Khora F1 score in
  the random inspection set. Despite zero detector F1, a player remains
  visible: the frame-38 confidence is 0.1445, below the 0.25 threshold.
  All methods use the same enlarged GT crop and green reference box.
  This diagnoses detector sensitivity, not verified actor absence.
  The six columns match Figure~\ref{fig:action_gt_regions}; the Solaris
  column is marked unavailable for this window.}
  \label{fig:action_detector_failure}
\end{figure}

The low-F1 example in Figure~\ref{fig:action_detector_failure} also exposes detector
false negatives on a visible generated player. To examine threshold
sensitivity, we recompute detection P/R/F1 on all 1,643 cases at confidence
thresholds 0.15, 0.25, 0.35, and 0.50, holding GT references and geometric
filters fixed. The within-window F1 ordering is unchanged at all four
thresholds. Khora F1 ranges from 0.888349 to 0.904063 in the early window
and 0.880531 to 0.902400 in the late window. This check does not eliminate
detector domain bias. Table~\ref{tab:action_thresholds} reports every count
and P/R/F1 value from this sensitivity analysis.

\input{sec/evidence_action_plots}

\FloatBarrier
\appneedspace{8\baselineskip}
\subsection{Complete Numerical Results}
\label{app:action_full}
Tables~\ref{tab:action_full_early} and~\ref{tab:action_full_late} report the
complete results across eight model/window combinations, including
supplementary scores, valid-case counts, and sample standard deviations.
Main-text values use four decimals; these full tables use six. All methods
share the same selected cases within each window. ``N/A'' denotes an
undefined statistic, never a zero imputed for a failed GT pseudo-label.
\input{sec/action_full_tables}

\FloatBarrier
\appneedspace{16\baselineskip}
\subsection{Uncertainty Estimates}
\label{app:action_uncertainty}
Table~\ref{tab:action_ci} reports the episode-cluster bootstrap intervals
for each model and scoring window; Table~\ref{tab:action_paired_ci} gives
the paired Khora-minus-ablation differences. The resampling protocol and
scope of these intervals are described in Section~\ref{app:action_protocol}.
\input{sec/action_ci_tables}

\FloatBarrier
\appneedspace{18\baselineskip}
\subsection{Detection-Threshold Sensitivity}
\label{app:action_threshold}
Table~\ref{tab:action_thresholds} gives the complete counts and detection
scores at all four confidence thresholds, separately for the two scoring
windows. Only the generated detection threshold varies; the GT references,
cohort, and geometric filters remain fixed.
\input{sec/action_threshold_table}

%% file: sec/evidence_action_plots.tex
\FloatBarrier
\subsection{Diagnostics and Visual Summaries}
\label{app:action_diagnostics}

\begin{figure}[tp]
\centering
\includegraphics[width=0.92\linewidth]{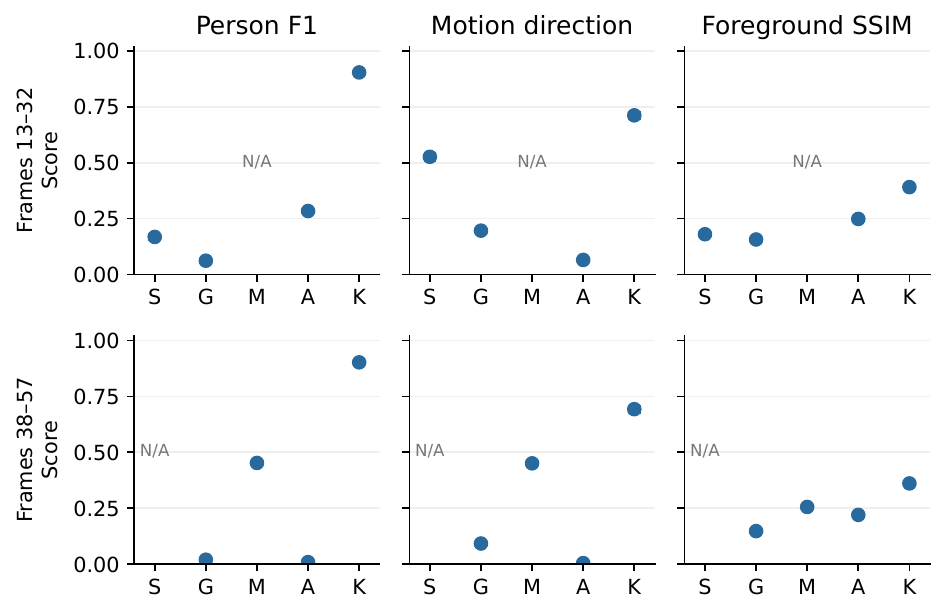}
\caption{\textbf{Action consistency with episode-cluster uncertainty.}
S: Solaris; G: Gamma-World; M: MIRA; A: Khora w/o Cross-Agent
State; K: Khora. Fixed model positions follow publication order, with
N/A marking unavailable model/window combinations. Dots show estimates; bars show 95\% percentile intervals from
2,000 paired episode bootstrap replicates. Higher values are better for all
displayed metrics. Exact values, the other main metrics, and paired differences
appear in Tables~\ref{tab:action_ci} and~\ref{tab:action_paired_ci}.}
\label{fig:action_intervals_plot}
\end{figure}

Figure~\ref{fig:action_intervals_plot} visualizes three complementary
quantities from the complete confidence-interval table: person localization,
image-motion direction, and appearance in the fixed GT foreground. Each
window uses the same 1,643 selected cases. The two windows remain separate,
and absent model/window combinations are not imputed. Confidence intervals are computed by resampling
source episodes, preserving within-episode dependence; they address
sampling uncertainty rather than detector bias.

\subsubsection{Detector sensitivity}
\begin{figure}[tp]
\centering
\includegraphics[width=\linewidth]{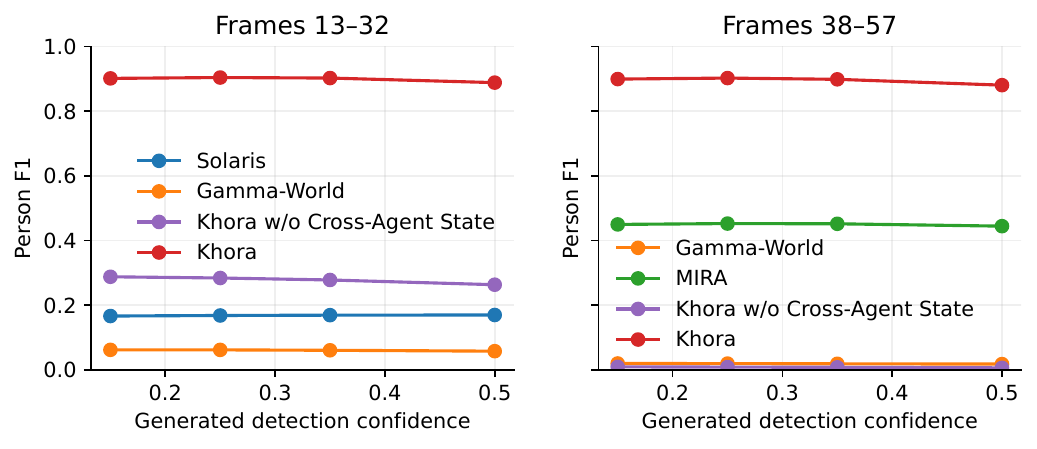}
\caption{\textbf{Sensitivity of person F1 to detection confidence.}
The two panels use different physical frame windows. Lines join the four
recomputed thresholds, 0.15, 0.25, 0.35, and 0.50; the main evaluation uses
0.25. These curves reuse all 1,643 cases and do not reselect the cohort.}
\label{fig:action_threshold_plot}
\end{figure}

Figure~\ref{fig:action_threshold_plot} varies only the confidence threshold
applied to generated detections. GT pseudo-labels, case selection, and
geometric filtering remain fixed. All four thresholds preserve the
within-window F1 ordering, but threshold stability does not validate every
individual detection. In particular, the visible player in the low-F1
example can be missed by the detector. The complete threshold table reports
all TP, FP, FN, precision, recall, and F1 entries.

\subsubsection{Pixel-score controls}
We use one diagnostic clip for identity, black, white, time reversal,
frozen-first-frame, 100-pixel shift, and GT-actor-inpainting controls.
Table~\ref{tab:pixel_controls} reports appearance scores
for both the GT box and GT foreground. All control regions are fixed from
GT; no generated region is selected or aligned. These controls test metric
behavior on one clip and are not extra model baselines or population means.

\input{sec/evidence_pixel_controls}

The controls motivate joint reporting rather than treating color similarity
as a detection accuracy. Pixel metrics deliberately retain frames where the
generated person is absent; localization, pose, and motion metrics expose
errors that a background-dominated or brightness-sensitive average may hide.

PSNR averages the per-frame logarithmic scores, whereas MSE is averaged
in linear RGB-error space. Consequently, the printed mean PSNR need not
equal $-10\log_{10}$ of the printed mean MSE. Exact-match frames use the
same 100 dB cap as the full model evaluation.

\begin{figure}[tp]
\centering
\includegraphics[width=\linewidth]{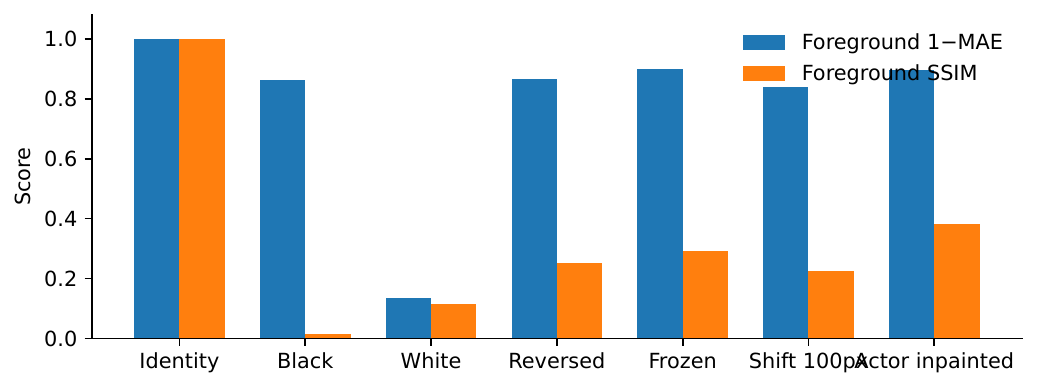}
\caption{\textbf{Why fixed-region RGB similarity needs structural checks.}
All bars use the same GT foreground. A black prediction or an inpainted
actor can retain a high $1-\mathrm{MAE}$ score. SSIM is more sensitive to
local structure but also cannot certify actor presence by itself. The
identity control reaches one on both scores.}
\label{fig:pixel_controls}
\end{figure}

%% file: sec/evidence_pixel_controls.tex
\begin{table}[tp]
\centering\setlength{\tabcolsep}{3pt}
\caption{\textbf{All fixed-region appearance controls.} One diagnostic clip, 20 frames; 313,284 box pixels and 126,508 foreground pixels. All regions are valid in all frames. Identity has 20 exact matches; frozen-first-frame has one; other controls have zero. PSNR is capped at 100 dB.}
\label{tab:pixel_controls}
\begin{tabularx}{\linewidth}{LcN{9.999999}N{9.999999}N{999.999999}N{9.999999}N{9.999999}}
\toprule
\tableheadleft{Control} & \tablehead{Region} & \tablehead{MAE $\downarrow$} & \tablehead{MSE $\downarrow$} & \tablehead{PSNR $\uparrow$} & \tablehead{$1-$MAE $\uparrow$} & \tablehead{SSIM $\uparrow$} \\
\midrule
identity & box & 0.000000 & 0.000000 & 100.000000 & 1.000000 & 1.000000 \\
\rowcolor{tableStripe}
identity & foreground & 0.000000 & 0.000000 & 100.000000 & 1.000000 & 1.000000 \\
black & box & 0.193999 & 0.052088 & 12.939865 & 0.806001 & 0.009516 \\
\rowcolor{tableStripe}
black & foreground & 0.135521 & 0.024501 & 16.161390 & 0.864479 & 0.013803 \\
white & box & 0.806001 & 0.664090 & 1.779099 & 0.193999 & 0.184863 \\
\rowcolor{tableStripe}
white & foreground & 0.864479 & 0.753459 & 1.231538 & 0.135521 & 0.113299 \\
reverse time & box & 0.121000 & 0.027501 & 16.333985 & 0.879000 & 0.310654 \\
\rowcolor{tableStripe}
reverse time & foreground & 0.132385 & 0.027853 & 16.037519 & 0.867615 & 0.253468 \\
frozen first frame & box & 0.094805 & 0.018044 & 22.386295 & 0.905195 & 0.366631 \\
\rowcolor{tableStripe}
frozen first frame & foreground & 0.101878 & 0.017520 & 22.081798 & 0.898122 & 0.293297 \\
shift100px & box & 0.153501 & 0.043562 & 14.393723 & 0.846499 & 0.264591 \\
\rowcolor{tableStripe}
shift100px & foreground & 0.159509 & 0.045927 & 14.203725 & 0.840491 & 0.225933 \\
Actor inpainted & box & 0.042571 & 0.006973 & 21.947406 & 0.957429 & 0.713602 \\
\rowcolor{tableStripe}
Actor inpainted & foreground & 0.103889 & 0.017032 & 18.020472 & 0.896111 & 0.383361 \\
\bottomrule
\end{tabularx}
\end{table}

%% file: sec/action_full_tables.tex
\begin{table}[!tp]
\centering
\setlength{\tabcolsep}{3pt}
\caption{\textbf{Complete cross-view action consistency results, frames 13--32.} All scalar metrics, counts, valid-case denominators, and sample standard deviations. Six decimal places are shown; integer counts are exact. Baselines: Solaris~\citep{savva2026solaris} and Gamma-World~\citep{liu2026gammaworld}.}\label{tab:action_full_early}
\begin{tabularx}{\linewidth}{LN{99.999999}N{99.999999}N{99.999999}N{99.999999}}
\toprule
\tableheadleft{Statistic} & \tablehead{Solaris} & \tablehead{Gamma-\\World} & \tablehead{Khora w/o\\Cross-Agent\\State} & \tablehead{Khora} \\ \midrule
\multicolumn{5}{l}{\textit{Detection, position, and trajectory}}\\
Available-track ADE ($\%D$) $\downarrow$ & 5.155349 & 7.332127 & \secondscore{1.510989} & \bestscore{0.532387} \\
\rowcolor{tableStripe}
Available-track ADE (pixels) $\downarrow$ & 37.855851 & 53.839983 & \secondscore{11.095231} & \bestscore{3.909328} \\
Missing-penalized ADE ($\%D$) $\downarrow$ & \secondscore{20.379866} & 53.844661 & 74.050573 & \bestscore{13.032861} \\
\rowcolor{tableStripe}
Direction accuracy (4 frames) $\uparrow$ & \secondscore{0.526670} & 0.196322 & 0.065353 & \bestscore{0.712070} \\
Available-track direction cosine $\uparrow$ & \secondscore{0.515515} & 0.278088 & 0.313948 & \bestscore{0.830552} \\
\rowcolor{tableStripe}
Extra detections per frame $\downarrow$ & 0.041570 & 0.085027 & \bestscore{0.012203} & \secondscore{0.039623} \\
Detection F1 $\uparrow$ & 0.168277 & 0.061932 & \secondscore{0.284192} & \bestscore{0.904063} \\
\rowcolor{tableStripe}
Available-track FDE (pixels) $\downarrow$ & 52.059019 & 67.112213 & \secondscore{42.719948} & \bestscore{3.987530} \\
Missing-penalized FDE ($\%D$) $\downarrow$ & \secondscore{23.941264} & 58.745059 & 91.860086 & \bestscore{14.526335} \\
\rowcolor{tableStripe}
Mean best IoU, all frames $\uparrow$ & \secondscore{0.204078} & 0.084616 & 0.158711 & \bestscore{0.702175} \\
Motion-pair coverage $\uparrow$ & \secondscore{0.801430} & 0.398509 & 0.165056 & \bestscore{0.809457} \\
\rowcolor{tableStripe}
Image move/stop accuracy $\uparrow$ & \secondscore{0.645123} & 0.311967 & 0.115110 & \bestscore{0.722573} \\
Center PCK ($0.05D$) $\uparrow$ & \secondscore{0.532988} & 0.215307 & 0.241996 & \bestscore{0.871029} \\
\rowcolor{tableStripe}
Position AUC ($0$--$0.1D$) $\uparrow$ & \secondscore{0.463675} & 0.204899 & 0.216242 & \bestscore{0.828912} \\
Detection precision $\uparrow$ & 0.176077 & 0.075568 & \secondscore{0.612741} & \bestscore{0.938466} \\
\rowcolor{tableStripe}
Detection recall $\uparrow$ & 0.161138 & 0.052465 & \secondscore{0.184997} & \bestscore{0.872094} \\
Target-track frame coverage $\uparrow$ & \secondscore{0.839166} & 0.496287 & 0.264699 & \bestscore{0.874254} \\
\rowcolor{tableStripe}
Track breaks per case $\downarrow$ & \bestscore{0.468655} & 1.294583 & 0.914790 & \secondscore{0.768107} \\
\addlinespace
\multicolumn{5}{l}{\textit{Pose scores}}\\
Absolute pose PCK ($0.2h$) $\uparrow$ & \secondscore{0.211109} & 0.049990 & 0.172258 & \bestscore{0.743765} \\
\rowcolor{tableStripe}
Generated joint coverage $\uparrow$ & \secondscore{0.719683} & 0.360884 & 0.225433 & \bestscore{0.772633} \\
Root-normalized shape PCK $\uparrow$ & \secondscore{0.679222} & 0.326775 & 0.211646 & \bestscore{0.744580} \\
\addlinespace
\multicolumn{5}{l}{\textit{Fixed GT box and foreground appearance}}\\
Box frame coverage & 1.000000 & 1.000000 & 1.000000 & 1.000000 \\
\rowcolor{tableStripe}
Box MAE $\downarrow$ & 0.137810 & 0.174213 & \secondscore{0.100830} & \bestscore{0.069708} \\
Box $1-\mathrm{MAE}$ $\uparrow$ & 0.862190 & 0.825787 & \secondscore{0.899170} & \bestscore{0.930292} \\
\rowcolor{tableStripe}
Box MSE $\downarrow$ & 0.040059 & 0.059326 & \secondscore{0.026387} & \bestscore{0.013350} \\
Box PSNR (dB, cap 100) $\uparrow$ & 15.899162 & 14.136474 & \secondscore{17.852195} & \bestscore{20.411451} \\
\rowcolor{tableStripe}
Box SSIM $\uparrow$ & 0.248882 & 0.212112 & \secondscore{0.388850} & \bestscore{0.473400} \\
Box $1-\mathrm{MAE}$, frame-weighted $\uparrow$ & 0.862190 & 0.825787 & \secondscore{0.899170} & \bestscore{0.930292} \\
\rowcolor{tableStripe}
Box SSIM, frame-weighted $\uparrow$ & 0.248882 & 0.212112 & \secondscore{0.388850} & \bestscore{0.473400} \\
Foreground frame coverage & 0.996774 & 0.996774 & 0.996774 & 0.996774 \\
\rowcolor{tableStripe}
Foreground MAE $\downarrow$ & 0.164680 & 0.199953 & \secondscore{0.141726} & \bestscore{0.083418} \\
Foreground $1-\mathrm{MAE}$ $\uparrow$ & 0.835320 & 0.800047 & \secondscore{0.858274} & \bestscore{0.916582} \\
\rowcolor{tableStripe}
Foreground MSE $\downarrow$ & 0.050716 & 0.072776 & \secondscore{0.040470} & \bestscore{0.016864} \\
Foreground PSNR (dB, cap 100) $\uparrow$ & 15.055761 & 13.481165 & \secondscore{16.134487} & \bestscore{19.359696} \\
\rowcolor{tableStripe}
Foreground SSIM $\uparrow$ & 0.180287 & 0.156796 & \secondscore{0.248728} & \bestscore{0.391012} \\
Foreground $1-\mathrm{MAE}$, frame-weighted $\uparrow$ & 0.835348 & 0.800036 & \secondscore{0.858299} & \bestscore{0.916589} \\
\rowcolor{tableStripe}
Foreground SSIM, frame-weighted $\uparrow$ & 0.180345 & 0.156823 & \secondscore{0.248720} & \bestscore{0.390966} \\
\bottomrule
\end{tabularx}
\end{table}

\begin{table}[!tp]
\centering
\setlength{\tabcolsep}{3pt}
\ContinuedFloat
\caption[]{\textbf{Continued.} Complete cross-view action consistency results, frames 13--32.}
\begin{tabularx}{\linewidth}{LN{999999999}N{999999999}N{999999999}N{999999999}}
\toprule
\tableheadleft{Statistic} & \tablehead{Solaris} & \tablehead{Gamma-\\World} & \tablehead{Khora w/o\\Cross-Agent\\State} & \tablehead{Khora} \\ \midrule
\multicolumn{5}{l}{\textit{Counts and coverage denominators}}\\
Cases & 1643 & 1643 & 1643 & 1643 \\
\rowcolor{tableStripe}
False negatives (IoU $\geq 0.5$) & 27565 & 31136 & 26781 & 4203 \\
False positives (IoU $\geq 0.5$) & 24777 & 21090 & 3842 & 1879 \\
\rowcolor{tableStripe}
Correct-direction pairs & 12056 & 4494 & 1496 & 16300 \\
Correct move/stop pairs & 16959 & 8201 & 3026 & 18995 \\
\rowcolor{tableStripe}
Scored frames & 32860 & 32860 & 32860 & 32860 \\
GT moving pairs & 22891 & 22891 & 22891 & 22891 \\
\rowcolor{tableStripe}
All motion pairs & 26288 & 26288 & 26288 & 26288 \\
Retained generated detections & 30072 & 22814 & 9921 & 30536 \\
\rowcolor{tableStripe}
True positives (IoU $\geq 0.5$) & 5295 & 1724 & 6079 & 28657 \\
Correct absolute body joints & 51643 & 12229 & 42139 & 181945 \\
\rowcolor{tableStripe}
Confident GT body joints & 244627 & 244627 & 244627 & 244627 \\
Frames with GT pose & 23028 & 23028 & 23028 & 23028 \\
\rowcolor{tableStripe}
Available generated body joints & 176054 & 88282 & 55147 & 189007 \\
Frames with paired generated pose & 17897 & 9383 & 5664 & 18849 \\
\rowcolor{tableStripe}
Correct normalized shape joints & 166152 & 79936 & 51773 & 182140 \\
GT joints eligible for shape & 244621 & 244621 & 244621 & 244621 \\
\rowcolor{tableStripe}
Box valid frames & 32860 & 32860 & 32860 & 32860 \\
Box candidate frames & 32860 & 32860 & 32860 & 32860 \\
\rowcolor{tableStripe}
Box scored pixels & 139719152 & 139719152 & 139719152 & 139719152 \\
Box exact-match frames & 0 & 0 & 0 & 0 \\
\rowcolor{tableStripe}
Foreground valid frames & 32754 & 32754 & 32754 & 32754 \\
Foreground candidate frames & 32860 & 32860 & 32860 & 32860 \\
\rowcolor{tableStripe}
Foreground scored pixels & 62380972 & 62380972 & 62380972 & 62380972 \\
Foreground exact-match frames & 0 & 0 & 0 & 0 \\
\addlinespace
\multicolumn{5}{l}{\textit{Valid-case counts for each statistic}}\\
Available-track ADE ($\%D$): valid cases & 1525 & 1259 & 1241 & 1634 \\
\rowcolor{tableStripe}
Available-track ADE (pixels): valid cases & 1525 & 1259 & 1241 & 1634 \\
Missing-penalized ADE ($\%D$): valid cases & 1643 & 1643 & 1643 & 1643 \\
\rowcolor{tableStripe}
Direction accuracy (4 frames): valid cases & 1638 & 1638 & 1638 & 1638 \\
Available-track direction cosine: valid cases & 1477 & 1063 & 596 & 1587 \\
\rowcolor{tableStripe}
Extra detections per frame: valid cases & 1643 & 1643 & 1643 & 1643 \\
Detection F1: valid cases & 1643 & 1643 & 1643 & 1643 \\
\rowcolor{tableStripe}
Available-track FDE (pixels): valid cases & 1345 & 746 & 142 & 1412 \\
Missing-penalized FDE ($\%D$): valid cases & 1643 & 1643 & 1643 & 1643 \\
\rowcolor{tableStripe}
False negatives (IoU $\geq 0.5$): valid cases & 1643 & 1643 & 1643 & 1643 \\
False positives (IoU $\geq 0.5$): valid cases & 1643 & 1643 & 1643 & 1643 \\
\rowcolor{tableStripe}
Mean best IoU, all frames: valid cases & 1643 & 1643 & 1643 & 1643 \\
Motion-pair coverage: valid cases & 1643 & 1643 & 1643 & 1643 \\
\rowcolor{tableStripe}
Image move/stop accuracy: valid cases & 1643 & 1643 & 1643 & 1643 \\
Correct-direction pairs: valid cases & 1643 & 1643 & 1643 & 1643 \\
\rowcolor{tableStripe}
Correct move/stop pairs: valid cases & 1643 & 1643 & 1643 & 1643 \\
Scored frames: valid cases & 1643 & 1643 & 1643 & 1643 \\
\rowcolor{tableStripe}
GT moving pairs: valid cases & 1643 & 1643 & 1643 & 1643 \\
\bottomrule
\end{tabularx}
\end{table}

\begin{table}[!tp]
\centering
\setlength{\tabcolsep}{3pt}
\ContinuedFloat
\caption[]{\textbf{Continued.} Complete cross-view action consistency results, frames 13--32.}
\begin{tabularx}{\linewidth}{LN{99.999999}N{99.999999}N{99.999999}N{9.999999}}
\toprule
\tableheadleft{Statistic} & \tablehead{Solaris} & \tablehead{Gamma-\\World} & \tablehead{Khora w/o\\Cross-Agent\\State} & \tablehead{Khora} \\ \midrule
\multicolumn{5}{l}{\textit{Valid-case counts for each statistic (continued)}}\\
All motion pairs: valid cases & 1643 & 1643 & 1643 & 1643 \\
\rowcolor{tableStripe}
Retained generated detections: valid cases & 1643 & 1643 & 1643 & 1643 \\
Center PCK ($0.05D$): valid cases & 1643 & 1643 & 1643 & 1643 \\
\rowcolor{tableStripe}
Position AUC ($0$--$0.1D$): valid cases & 1643 & 1643 & 1643 & 1643 \\
Detection precision: valid cases & 1643 & 1643 & 1643 & 1643 \\
\rowcolor{tableStripe}
Detection recall: valid cases & 1643 & 1643 & 1643 & 1643 \\
Target-track frame coverage: valid cases & 1643 & 1643 & 1643 & 1643 \\
\rowcolor{tableStripe}
True positives (IoU $\geq 0.5$): valid cases & 1643 & 1643 & 1643 & 1643 \\
Track breaks per case: valid cases & 1643 & 1643 & 1643 & 1643 \\
\rowcolor{tableStripe}
Absolute pose PCK ($0.2h$): valid cases & 1288 & 1288 & 1288 & 1288 \\
Correct absolute body joints: valid cases & 1643 & 1643 & 1643 & 1643 \\
\rowcolor{tableStripe}
Confident GT body joints: valid cases & 1643 & 1643 & 1643 & 1643 \\
Frames with GT pose: valid cases & 1643 & 1643 & 1643 & 1643 \\
\rowcolor{tableStripe}
Generated joint coverage: valid cases & 1288 & 1288 & 1288 & 1288 \\
Available generated body joints: valid cases & 1643 & 1643 & 1643 & 1643 \\
\rowcolor{tableStripe}
Frames with paired generated pose: valid cases & 1643 & 1643 & 1643 & 1643 \\
Root-normalized shape PCK: valid cases & 1288 & 1288 & 1288 & 1288 \\
\rowcolor{tableStripe}
Correct normalized shape joints: valid cases & 1643 & 1643 & 1643 & 1643 \\
GT joints eligible for shape: valid cases & 1643 & 1643 & 1643 & 1643 \\
\rowcolor{tableStripe}
Box MAE: valid cases & 1643 & 1643 & 1643 & 1643 \\
Box $1-\mathrm{MAE}$: valid cases & 1643 & 1643 & 1643 & 1643 \\
\rowcolor{tableStripe}
Box MSE: valid cases & 1643 & 1643 & 1643 & 1643 \\
Box PSNR (dB, cap 100): valid cases & 1643 & 1643 & 1643 & 1643 \\
\rowcolor{tableStripe}
Box SSIM: valid cases & 1643 & 1643 & 1643 & 1643 \\
Foreground MAE: valid cases & 1643 & 1643 & 1643 & 1643 \\
\rowcolor{tableStripe}
Foreground $1-\mathrm{MAE}$: valid cases & 1643 & 1643 & 1643 & 1643 \\
Foreground MSE: valid cases & 1643 & 1643 & 1643 & 1643 \\
\rowcolor{tableStripe}
Foreground PSNR (dB, cap 100): valid cases & 1643 & 1643 & 1643 & 1643 \\
Foreground SSIM: valid cases & 1643 & 1643 & 1643 & 1643 \\
\addlinespace
\multicolumn{5}{l}{\textit{Across-case sample standard deviations}}\\
Available-track ADE (pixels): case SD & 31.217285 & 35.289713 & 14.888488 & 4.970599 \\
\rowcolor{tableStripe}
Mean best IoU, all frames: case SD & 0.209103 & 0.133646 & 0.172150 & 0.174980 \\
Position AUC ($0$--$0.1D$): case SD & 0.298586 & 0.242170 & 0.235268 & 0.205931 \\
\rowcolor{tableStripe}
Absolute pose PCK ($0.2h$): case SD & 0.275559 & 0.123267 & 0.202844 & 0.302854 \\
Box MAE: case SD & 0.069088 & 0.081146 & 0.045313 & 0.027811 \\
\rowcolor{tableStripe}
Box $1-\mathrm{MAE}$: case SD & 0.069088 & 0.081146 & 0.045313 & 0.027811 \\
Box MSE: case SD & 0.034486 & 0.046501 & 0.021108 & 0.009809 \\
\rowcolor{tableStripe}
Box PSNR (dB, cap 100): case SD & 4.341592 & 4.299445 & 4.276162 & 3.827186 \\
Box SSIM: case SD & 0.135028 & 0.125549 & 0.111220 & 0.095963 \\
\rowcolor{tableStripe}
Foreground MAE: case SD & 0.086697 & 0.099556 & 0.068595 & 0.032630 \\
Foreground $1-\mathrm{MAE}$: case SD & 0.086697 & 0.099556 & 0.068595 & 0.032630 \\
\rowcolor{tableStripe}
Foreground MSE: case SD & 0.047088 & 0.060868 & 0.034412 & 0.012490 \\
Foreground PSNR (dB, cap 100): case SD & 4.430126 & 4.455401 & 4.326201 & 3.726357 \\
\rowcolor{tableStripe}
Foreground SSIM: case SD & 0.112952 & 0.102911 & 0.111444 & 0.098158 \\
\bottomrule
\end{tabularx}
\end{table}

\begin{table}[!tp]
\centering
\setlength{\tabcolsep}{3pt}
\caption{\textbf{Complete cross-view action consistency results, frames 38--57.} All scalar metrics, counts, valid-case denominators, and sample standard deviations. Six decimal places are shown; integer counts are exact. Baselines: Gamma-World~\citep{liu2026gammaworld} and MIRA~\citep{hu2026mira}.}\label{tab:action_full_late}
\begin{tabularx}{\linewidth}{LN{99.999999}N{99.999999}N{99.999999}N{99.999999}}
\toprule
\tableheadleft{Statistic} & \tablehead{Gamma-\\World} & \tablehead{MIRA} & \tablehead{Khora w/o\\Cross-Agent\\State} & \tablehead{Khora} \\ \midrule
\multicolumn{5}{l}{\textit{Detection, position, and trajectory}}\\
Available-track ADE ($\%D$) $\downarrow$ & 8.480216 & \secondscore{3.085277} & 6.758419 & \bestscore{0.572149} \\
\rowcolor{tableStripe}
Available-track ADE (pixels) $\downarrow$ & 62.270429 & \secondscore{22.655263} & 49.627232 & \bestscore{4.201300} \\
Missing-penalized ADE ($\%D$) $\downarrow$ & 73.272929 & \secondscore{17.046396} & 96.600790 & \bestscore{13.700354} \\
\rowcolor{tableStripe}
Direction accuracy (4 frames) $\uparrow$ & 0.092087 & \secondscore{0.450706} & 0.004259 & \bestscore{0.692939} \\
Available-track direction cosine $\uparrow$ & 0.195565 & \secondscore{0.361216} & 0.158781 & \bestscore{0.806403} \\
\rowcolor{tableStripe}
Extra detections per frame $\downarrow$ & 0.067985 & \secondscore{0.033262} & \bestscore{0.002161} & 0.035180 \\
Detection F1 $\uparrow$ & 0.019482 & \secondscore{0.452510} & 0.008690 & \bestscore{0.902400} \\
\rowcolor{tableStripe}
Available-track FDE (pixels) $\downarrow$ & 66.919357 & \secondscore{41.816554} & 55.790681 & \bestscore{4.219938} \\
Missing-penalized FDE ($\%D$) $\downarrow$ & 76.987914 & \secondscore{23.430781} & 97.637923 & \bestscore{19.515723} \\
\rowcolor{tableStripe}
Mean best IoU, all frames $\uparrow$ & 0.035263 & \secondscore{0.391557} & 0.007622 & \bestscore{0.685430} \\
Motion-pair coverage $\uparrow$ & 0.221432 & \secondscore{0.799871} & 0.025829 & \bestscore{0.804664} \\
\rowcolor{tableStripe}
Image move/stop accuracy $\uparrow$ & 0.165969 & \secondscore{0.624011} & 0.014227 & \bestscore{0.703971} \\
Center PCK ($0.05D$) $\uparrow$ & 0.091023 & \secondscore{0.702648} & 0.019781 & \bestscore{0.865520} \\
\rowcolor{tableStripe}
Position AUC ($0$--$0.1D$) $\uparrow$ & 0.096553 & \secondscore{0.623204} & 0.017384 & \bestscore{0.819846} \\
Detection precision $\uparrow$ & 0.027567 & \secondscore{0.477649} & 0.065023 & \bestscore{0.943756} \\
\rowcolor{tableStripe}
Detection recall $\uparrow$ & 0.015064 & \secondscore{0.429884} & 0.004656 & \bestscore{0.864516} \\
Target-track frame coverage $\uparrow$ & 0.290809 & \secondscore{0.856086} & 0.036062 & \bestscore{0.867833} \\
\rowcolor{tableStripe}
Track breaks per case $\downarrow$ & 0.926354 & \secondscore{0.763238} & \bestscore{0.121120} & 0.810712 \\
\addlinespace
\multicolumn{5}{l}{\textit{Pose scores}}\\
Absolute pose PCK ($0.2h$) $\uparrow$ & 0.018057 & \secondscore{0.412928} & 0.004485 & \bestscore{0.738481} \\
\rowcolor{tableStripe}
Generated joint coverage $\uparrow$ & 0.200740 & \secondscore{0.692774} & 0.021243 & \bestscore{0.767578} \\
Root-normalized shape PCK $\uparrow$ & 0.177811 & \secondscore{0.652417} & 0.017866 & \bestscore{0.738853} \\
\addlinespace
\multicolumn{5}{l}{\textit{Fixed GT box and foreground appearance}}\\
Box frame coverage & 1.000000 & 1.000000 & 1.000000 & 1.000000 \\
\rowcolor{tableStripe}
Box MAE $\downarrow$ & 0.201603 & \secondscore{0.108183} & 0.109529 & \bestscore{0.073891} \\
Box $1-\mathrm{MAE}$ $\uparrow$ & 0.798397 & \secondscore{0.891817} & 0.890471 & \bestscore{0.926109} \\
\rowcolor{tableStripe}
Box MSE $\downarrow$ & 0.075867 & \secondscore{0.026792} & 0.030498 & \bestscore{0.014456} \\
Box PSNR (dB, cap 100) $\uparrow$ & 13.167113 & \secondscore{17.715944} & 17.079508 & \bestscore{19.967869} \\
\rowcolor{tableStripe}
Box SSIM $\uparrow$ & 0.195440 & 0.329751 & \secondscore{0.381442} & \bestscore{0.445497} \\
Box $1-\mathrm{MAE}$, frame-weighted $\uparrow$ & 0.798397 & \secondscore{0.891817} & 0.890471 & \bestscore{0.926109} \\
\rowcolor{tableStripe}
Box SSIM, frame-weighted $\uparrow$ & 0.195440 & 0.329751 & \secondscore{0.381442} & \bestscore{0.445497} \\
Foreground frame coverage & 0.997505 & 0.997505 & 0.997505 & 0.997505 \\
\rowcolor{tableStripe}
Foreground MAE $\downarrow$ & 0.226105 & \secondscore{0.126754} & 0.162544 & \bestscore{0.086796} \\
Foreground $1-\mathrm{MAE}$ $\uparrow$ & 0.773895 & \secondscore{0.873246} & 0.837456 & \bestscore{0.913204} \\
\rowcolor{tableStripe}
Foreground MSE $\downarrow$ & 0.091118 & \secondscore{0.032994} & 0.049062 & \bestscore{0.017780} \\
Foreground PSNR (dB, cap 100) $\uparrow$ & 12.664245 & \secondscore{16.916130} & 15.030659 & \bestscore{19.040465} \\
\rowcolor{tableStripe}
Foreground SSIM $\uparrow$ & 0.147564 & \secondscore{0.255510} & 0.220065 & \bestscore{0.360863} \\
Foreground $1-\mathrm{MAE}$, frame-weighted $\uparrow$ & 0.773969 & \secondscore{0.873271} & 0.837491 & \bestscore{0.913202} \\
\rowcolor{tableStripe}
Foreground SSIM, frame-weighted $\uparrow$ & 0.147587 & \secondscore{0.255570} & 0.220087 & \bestscore{0.360813} \\
\bottomrule
\end{tabularx}
\end{table}

\begin{table}[!tp]
\centering
\setlength{\tabcolsep}{3pt}
\ContinuedFloat
\caption[]{\textbf{Continued.} Complete cross-view action consistency results, frames 38--57.}
\begin{tabularx}{\linewidth}{LN{999999999}N{999999999}N{999999999}N{999999999}}
\toprule
\tableheadleft{Statistic} & \tablehead{Gamma-\\World} & \tablehead{MIRA} & \tablehead{Khora w/o\\Cross-Agent\\State} & \tablehead{Khora} \\ \midrule
\multicolumn{5}{l}{\textit{Counts and coverage denominators}}\\
Cases & 1643 & 1643 & 1643 & 1643 \\
\rowcolor{tableStripe}
False negatives (IoU $\geq 0.5$) & 32365 & 18734 & 32707 & 4452 \\
False positives (IoU $\geq 0.5$) & 17461 & 15448 & 2200 & 1693 \\
\rowcolor{tableStripe}
Correct-direction pairs & 2054 & 10053 & 95 & 15456 \\
Correct move/stop pairs & 4363 & 16404 & 374 & 18506 \\
\rowcolor{tableStripe}
Scored frames & 32860 & 32860 & 32860 & 32860 \\
GT moving pairs & 22305 & 22305 & 22305 & 22305 \\
\rowcolor{tableStripe}
All motion pairs & 26288 & 26288 & 26288 & 26288 \\
Retained generated detections & 17956 & 29574 & 2353 & 30101 \\
\rowcolor{tableStripe}
True positives (IoU $\geq 0.5$) & 495 & 14126 & 153 & 28408 \\
Correct absolute body joints & 4421 & 101101 & 1098 & 180809 \\
\rowcolor{tableStripe}
Confident GT body joints & 244839 & 244839 & 244839 & 244839 \\
Frames with GT pose & 23005 & 23005 & 23005 & 23005 \\
\rowcolor{tableStripe}
Available generated body joints & 49149 & 169618 & 5201 & 187933 \\
Frames with paired generated pose & 5343 & 17099 & 651 & 18763 \\
\rowcolor{tableStripe}
Correct normalized shape joints & 43533 & 159730 & 4374 & 180892 \\
GT joints eligible for shape & 244828 & 244828 & 244828 & 244828 \\
\rowcolor{tableStripe}
Box valid frames & 32860 & 32860 & 32860 & 32860 \\
Box candidate frames & 32860 & 32860 & 32860 & 32860 \\
\rowcolor{tableStripe}
Box scored pixels & 141430664 & 141430664 & 141430664 & 141430664 \\
Box exact-match frames & 0 & 0 & 0 & 0 \\
\rowcolor{tableStripe}
Foreground valid frames & 32778 & 32778 & 32778 & 32778 \\
Foreground candidate frames & 32860 & 32860 & 32860 & 32860 \\
\rowcolor{tableStripe}
Foreground scored pixels & 63335210 & 63335210 & 63335210 & 63335210 \\
Foreground exact-match frames & 0 & 0 & 0 & 0 \\
\addlinespace
\multicolumn{5}{l}{\textit{Valid-case counts for each statistic}}\\
Available-track ADE ($\%D$): valid cases & 847 & 1617 & 145 & 1631 \\
\rowcolor{tableStripe}
Available-track ADE (pixels): valid cases & 847 & 1617 & 145 & 1631 \\
Missing-penalized ADE ($\%D$): valid cases & 1643 & 1643 & 1643 & 1643 \\
\rowcolor{tableStripe}
Direction accuracy (4 frames): valid cases & 1624 & 1624 & 1624 & 1624 \\
Available-track direction cosine: valid cases & 650 & 1537 & 75 & 1562 \\
\rowcolor{tableStripe}
Extra detections per frame: valid cases & 1643 & 1643 & 1643 & 1643 \\
Detection F1: valid cases & 1643 & 1643 & 1643 & 1643 \\
\rowcolor{tableStripe}
Available-track FDE (pixels): valid cases & 416 & 1334 & 42 & 1330 \\
Missing-penalized FDE ($\%D$): valid cases & 1643 & 1643 & 1643 & 1643 \\
\rowcolor{tableStripe}
False negatives (IoU $\geq 0.5$): valid cases & 1643 & 1643 & 1643 & 1643 \\
False positives (IoU $\geq 0.5$): valid cases & 1643 & 1643 & 1643 & 1643 \\
\rowcolor{tableStripe}
Mean best IoU, all frames: valid cases & 1643 & 1643 & 1643 & 1643 \\
Motion-pair coverage: valid cases & 1643 & 1643 & 1643 & 1643 \\
\rowcolor{tableStripe}
Image move/stop accuracy: valid cases & 1643 & 1643 & 1643 & 1643 \\
Correct-direction pairs: valid cases & 1643 & 1643 & 1643 & 1643 \\
\rowcolor{tableStripe}
Correct move/stop pairs: valid cases & 1643 & 1643 & 1643 & 1643 \\
Scored frames: valid cases & 1643 & 1643 & 1643 & 1643 \\
\rowcolor{tableStripe}
GT moving pairs: valid cases & 1643 & 1643 & 1643 & 1643 \\
\bottomrule
\end{tabularx}
\end{table}

\begin{table}[!tp]
\centering
\setlength{\tabcolsep}{3pt}
\ContinuedFloat
\caption[]{\textbf{Continued.} Complete cross-view action consistency results, frames 38--57.}
\begin{tabularx}{\linewidth}{LN{99.999999}N{99.999999}N{99.999999}N{9.999999}}
\toprule
\tableheadleft{Statistic} & \tablehead{Gamma-\\World} & \tablehead{MIRA} & \tablehead{Khora w/o\\Cross-Agent\\State} & \tablehead{Khora} \\ \midrule
\multicolumn{5}{l}{\textit{Valid-case counts for each statistic (continued)}}\\
All motion pairs: valid cases & 1643 & 1643 & 1643 & 1643 \\
\rowcolor{tableStripe}
Retained generated detections: valid cases & 1643 & 1643 & 1643 & 1643 \\
Center PCK ($0.05D$): valid cases & 1643 & 1643 & 1643 & 1643 \\
\rowcolor{tableStripe}
Position AUC ($0$--$0.1D$): valid cases & 1643 & 1643 & 1643 & 1643 \\
Detection precision: valid cases & 1643 & 1643 & 1643 & 1643 \\
\rowcolor{tableStripe}
Detection recall: valid cases & 1643 & 1643 & 1643 & 1643 \\
Target-track frame coverage: valid cases & 1643 & 1643 & 1643 & 1643 \\
\rowcolor{tableStripe}
True positives (IoU $\geq 0.5$): valid cases & 1643 & 1643 & 1643 & 1643 \\
Track breaks per case: valid cases & 1643 & 1643 & 1643 & 1643 \\
\rowcolor{tableStripe}
Absolute pose PCK ($0.2h$): valid cases & 1311 & 1311 & 1311 & 1311 \\
Correct absolute body joints: valid cases & 1643 & 1643 & 1643 & 1643 \\
\rowcolor{tableStripe}
Confident GT body joints: valid cases & 1643 & 1643 & 1643 & 1643 \\
Frames with GT pose: valid cases & 1643 & 1643 & 1643 & 1643 \\
\rowcolor{tableStripe}
Generated joint coverage: valid cases & 1311 & 1311 & 1311 & 1311 \\
Available generated body joints: valid cases & 1643 & 1643 & 1643 & 1643 \\
\rowcolor{tableStripe}
Frames with paired generated pose: valid cases & 1643 & 1643 & 1643 & 1643 \\
Root-normalized shape PCK: valid cases & 1311 & 1311 & 1311 & 1311 \\
\rowcolor{tableStripe}
Correct normalized shape joints: valid cases & 1643 & 1643 & 1643 & 1643 \\
GT joints eligible for shape: valid cases & 1643 & 1643 & 1643 & 1643 \\
\rowcolor{tableStripe}
Box MAE: valid cases & 1643 & 1643 & 1643 & 1643 \\
Box $1-\mathrm{MAE}$: valid cases & 1643 & 1643 & 1643 & 1643 \\
\rowcolor{tableStripe}
Box MSE: valid cases & 1643 & 1643 & 1643 & 1643 \\
Box PSNR (dB, cap 100): valid cases & 1643 & 1643 & 1643 & 1643 \\
\rowcolor{tableStripe}
Box SSIM: valid cases & 1643 & 1643 & 1643 & 1643 \\
Foreground MAE: valid cases & 1643 & 1643 & 1643 & 1643 \\
\rowcolor{tableStripe}
Foreground $1-\mathrm{MAE}$: valid cases & 1643 & 1643 & 1643 & 1643 \\
Foreground MSE: valid cases & 1643 & 1643 & 1643 & 1643 \\
\rowcolor{tableStripe}
Foreground PSNR (dB, cap 100): valid cases & 1643 & 1643 & 1643 & 1643 \\
Foreground SSIM: valid cases & 1643 & 1643 & 1643 & 1643 \\
\addlinespace
\multicolumn{5}{l}{\textit{Across-case sample standard deviations}}\\
Available-track ADE (pixels): case SD & 36.328662 & 23.289835 & 36.659425 & 4.773047 \\
\rowcolor{tableStripe}
Mean best IoU, all frames: case SD & 0.082983 & 0.215464 & 0.043863 & 0.181199 \\
Position AUC ($0$--$0.1D$): case SD & 0.176210 & 0.250192 & 0.090624 & 0.215206 \\
\rowcolor{tableStripe}
Absolute pose PCK ($0.2h$): case SD & 0.070866 & 0.313627 & 0.036918 & 0.308651 \\
Box MAE: case SD & 0.098199 & 0.048697 & 0.051581 & 0.030293 \\
\rowcolor{tableStripe}
Box $1-\mathrm{MAE}$: case SD & 0.098199 & 0.048697 & 0.051581 & 0.030293 \\
Box MSE: case SD & 0.061671 & 0.020717 & 0.025337 & 0.011033 \\
\rowcolor{tableStripe}
Box PSNR (dB, cap 100): case SD & 4.490447 & 4.081494 & 4.383445 & 3.776860 \\
Box SSIM: case SD & 0.121931 & 0.129327 & 0.113693 & 0.097813 \\
\rowcolor{tableStripe}
Foreground MAE: case SD & 0.118242 & 0.059475 & 0.078319 & 0.034689 \\
Foreground $1-\mathrm{MAE}$: case SD & 0.118242 & 0.059475 & 0.078319 & 0.034689 \\
\rowcolor{tableStripe}
Foreground MSE: case SD & 0.078751 & 0.027252 & 0.040515 & 0.013399 \\
Foreground PSNR (dB, cap 100): case SD & 4.717251 & 4.057525 & 4.398476 & 3.688559 \\
\rowcolor{tableStripe}
Foreground SSIM: case SD & 0.101376 & 0.114871 & 0.108171 & 0.095651 \\
\bottomrule
\end{tabularx}
\end{table}

%% file: sec/action_ci_tables.tex
\begingroup
\setlength{\tabcolsep}{4pt}
\begin{table}[!tp]
\centering
\caption{\textbf{Episode-cluster bootstrap intervals for all main-table scores.} Two thousand paired replicates over 1,466 source episodes; percentile 95\% intervals, conditional on the selected evaluation cohort. Baselines: Solaris~\citep{savva2026solaris}, Gamma-World~\citep{liu2026gammaworld}, and MIRA~\citep{hu2026mira}.}\label{tab:action_ci}
\begin{tabularx}{\linewidth}{lLN{9.999999}N{9.999999}N{9.999999}}
\toprule
\tableheadleft{Method} & \tableheadleft{Metric} & \tablehead{Estimate} & \tablehead{Lower 95\%} & \tablehead{Upper 95\%} \\ \midrule
\multicolumn{5}{l}{\textit{Relative frames 13--32}}\\
 & Detection F1 $\uparrow$ & 0.168277 & 0.155148 & 0.181531 \\
\rowcolor{tableStripe}
 & Detection recall $\uparrow$ & 0.161138 & 0.148115 & 0.174070 \\
 & Direction accuracy (4 frames) $\uparrow$ & 0.526670 & 0.510304 & 0.543214 \\
\rowcolor{tableStripe}
 & Absolute pose PCK ($0.2h$) $\uparrow$ & 0.211109 & 0.195324 & 0.227955 \\
 & Foreground SSIM $\uparrow$ & 0.180287 & 0.175056 & 0.186040 \\
\rowcolor{tableStripe}
\multirow{-6}{*}{Solaris} & Foreground MAE $\downarrow$ & 0.164680 & 0.160421 & 0.168658 \\
\addlinespace
 & Detection F1 $\uparrow$ & 0.061932 & 0.053986 & 0.070030 \\
\rowcolor{tableStripe}
 & Detection recall $\uparrow$ & 0.052465 & 0.045684 & 0.059479 \\
 & Direction accuracy (4 frames) $\uparrow$ & 0.196322 & 0.183887 & 0.208817 \\
\rowcolor{tableStripe}
 & Absolute pose PCK ($0.2h$) $\uparrow$ & 0.049990 & 0.042895 & 0.057416 \\
 & Foreground SSIM $\uparrow$ & 0.156796 & 0.151948 & 0.161934 \\
\rowcolor{tableStripe}
\multirow{-6}{*}{Gamma-World} & Foreground MAE $\downarrow$ & 0.199953 & 0.194890 & 0.204760 \\
\addlinespace
 & Detection F1 $\uparrow$ & 0.284192 & 0.271254 & 0.297233 \\
\rowcolor{tableStripe}
 & Detection recall $\uparrow$ & 0.184997 & 0.175137 & 0.194768 \\
 & Direction accuracy (4 frames) $\uparrow$ & 0.065353 & 0.058684 & 0.072398 \\
\rowcolor{tableStripe}
 & Absolute pose PCK ($0.2h$) $\uparrow$ & 0.172258 & 0.160885 & 0.183308 \\
 & Foreground SSIM $\uparrow$ & 0.248728 & 0.243653 & 0.254287 \\
\rowcolor{tableStripe}
\multirow{-6}{*}{\makecell[l]{Khora w/o\\Cross-Agent State}} & Foreground MAE $\downarrow$ & 0.141726 & 0.138325 & 0.145015 \\
\addlinespace
 & Detection F1 $\uparrow$ & 0.904063 & 0.897255 & 0.910779 \\
\rowcolor{tableStripe}
 & Detection recall $\uparrow$ & 0.872094 & 0.862477 & 0.881669 \\
 & Direction accuracy (4 frames) $\uparrow$ & 0.712070 & 0.697602 & 0.725972 \\
\rowcolor{tableStripe}
 & Absolute pose PCK ($0.2h$) $\uparrow$ & 0.743765 & 0.730095 & 0.756911 \\
 & Foreground SSIM $\uparrow$ & 0.391012 & 0.386251 & 0.395688 \\
\rowcolor{tableStripe}
\multirow{-6}{*}{\textbf{Khora}} & Foreground MAE $\downarrow$ & 0.083418 & 0.081792 & 0.084973 \\
\bottomrule
\end{tabularx}
\end{table}

\begin{table}[!tp]
\centering
\ContinuedFloat
\caption[]{\textbf{Continued.} Episode-cluster bootstrap intervals, frames 38--57.}
\begin{tabularx}{\linewidth}{lLN{9.999999}N{9.999999}N{9.999999}}
\toprule
\tableheadleft{Method} & \tableheadleft{Metric} & \tablehead{Estimate} & \tablehead{Lower 95\%} & \tablehead{Upper 95\%} \\ \midrule
\multicolumn{5}{l}{\textit{Relative frames 38--57}}\\
 & Detection F1 $\uparrow$ & 0.019482 & 0.015037 & 0.024414 \\
\rowcolor{tableStripe}
 & Detection recall $\uparrow$ & 0.015064 & 0.011583 & 0.018869 \\
 & Direction accuracy (4 frames) $\uparrow$ & 0.092087 & 0.083380 & 0.101183 \\
\rowcolor{tableStripe}
 & Absolute pose PCK ($0.2h$) $\uparrow$ & 0.018057 & 0.013859 & 0.022148 \\
 & Foreground SSIM $\uparrow$ & 0.147564 & 0.142932 & 0.152811 \\
\rowcolor{tableStripe}
\multirow{-6}{*}{Gamma-World} & Foreground MAE $\downarrow$ & 0.226105 & 0.220287 & 0.231622 \\
\addlinespace
 & Detection F1 $\uparrow$ & 0.452510 & 0.438689 & 0.467047 \\
\rowcolor{tableStripe}
 & Detection recall $\uparrow$ & 0.429884 & 0.416224 & 0.444601 \\
 & Direction accuracy (4 frames) $\uparrow$ & 0.450706 & 0.435808 & 0.466345 \\
\rowcolor{tableStripe}
 & Absolute pose PCK ($0.2h$) $\uparrow$ & 0.412928 & 0.395976 & 0.430564 \\
 & Foreground SSIM $\uparrow$ & 0.255510 & 0.250030 & 0.261062 \\
\rowcolor{tableStripe}
\multirow{-6}{*}{MIRA} & Foreground MAE $\downarrow$ & 0.126754 & 0.123871 & 0.129498 \\
\addlinespace
 & Detection F1 $\uparrow$ & 0.008690 & 0.004995 & 0.012845 \\
\rowcolor{tableStripe}
 & Detection recall $\uparrow$ & 0.004656 & 0.002663 & 0.006895 \\
 & Direction accuracy (4 frames) $\uparrow$ & 0.004259 & 0.002759 & 0.005863 \\
\rowcolor{tableStripe}
 & Absolute pose PCK ($0.2h$) $\uparrow$ & 0.004485 & 0.002416 & 0.006921 \\
 & Foreground SSIM $\uparrow$ & 0.220065 & 0.215129 & 0.225481 \\
\rowcolor{tableStripe}
\multirow{-6}{*}{\makecell[l]{Khora w/o\\Cross-Agent State}} & Foreground MAE $\downarrow$ & 0.162544 & 0.158687 & 0.166021 \\
\addlinespace
 & Detection F1 $\uparrow$ & 0.902400 & 0.895124 & 0.909580 \\
\rowcolor{tableStripe}
 & Detection recall $\uparrow$ & 0.864516 & 0.853689 & 0.875000 \\
 & Direction accuracy (4 frames) $\uparrow$ & 0.692939 & 0.677961 & 0.706945 \\
\rowcolor{tableStripe}
 & Absolute pose PCK ($0.2h$) $\uparrow$ & 0.738481 & 0.724728 & 0.752238 \\
 & Foreground SSIM $\uparrow$ & 0.360863 & 0.356268 & 0.365589 \\
\rowcolor{tableStripe}
\multirow{-6}{*}{\textbf{Khora}} & Foreground MAE $\downarrow$ & 0.086796 & 0.085012 & 0.088439 \\
\bottomrule
\end{tabularx}
\end{table}
\endgroup
\begin{table}[!tp]
\centering\caption{\textbf{Paired Khora minus ablation differences.} Signed differences in original metric units with episode-cluster bootstrap 95\% intervals. Negative MAE differences favor Khora; positive differences favor Khora for the other metrics.}\label{tab:action_paired_ci}
\begin{tabularx}{\linewidth}{cLN{-9.999999}N{-9.999999}N{-9.999999}}\toprule
\tablehead{Frames} & \tableheadleft{Metric} & \tablehead{Difference} & \tablehead{Lower 95\%} & \tablehead{Upper 95\%} \\ \midrule
 & Detection F1 $\uparrow$ & 0.619872 & 0.607150 & 0.631774 \\
\rowcolor{tableStripe}
 & Detection recall $\uparrow$ & 0.687097 & 0.675364 & 0.698097 \\
 & Direction accuracy (4 frames) $\uparrow$ & 0.646717 & 0.631147 & 0.661243 \\
\rowcolor{tableStripe}
 & Absolute pose PCK ($0.2h$) $\uparrow$ & 0.571507 & 0.557994 & 0.585445 \\
 & Foreground SSIM $\uparrow$ & 0.142285 & 0.138408 & 0.146243 \\
\rowcolor{tableStripe}
\multirow{-6}{*}{13--32} & Foreground MAE $\downarrow$ & -0.058308 & -0.060516 & -0.056113 \\
 & Detection F1 $\uparrow$ & 0.893710 & 0.885384 & 0.901474 \\
\rowcolor{tableStripe}
 & Detection recall $\uparrow$ & 0.859860 & 0.849057 & 0.870239 \\
 & Direction accuracy (4 frames) $\uparrow$ & 0.688680 & 0.673438 & 0.702869 \\
\rowcolor{tableStripe}
 & Absolute pose PCK ($0.2h$) $\uparrow$ & 0.733997 & 0.720478 & 0.747374 \\
 & Foreground SSIM $\uparrow$ & 0.140798 & 0.136622 & 0.145054 \\
\rowcolor{tableStripe}
\multirow{-6}{*}{38--57} & Foreground MAE $\downarrow$ & -0.075748 & -0.078307 & -0.073170 \\
\bottomrule
\end{tabularx}
\end{table}

%% file: sec/action_threshold_table.tex
\begingroup\setlength{\tabcolsep}{3pt}
\begin{table}[!tp]
\centering
\caption{\textbf{Detection-threshold sensitivity on all 1,643 cases.} Raw detections, GT boxes, and geometric filtering are fixed. Only the generated detection confidence threshold changes. P, R, and F1 pool the displayed counts; 0.25 reproduces the main evaluation. Ranks compare models at the same threshold and window. Baselines: Solaris~\citep{savva2026solaris}, Gamma-World~\citep{liu2026gammaworld}, and MIRA~\citep{hu2026mira}.}\label{tab:action_thresholds}
\begin{tabularx}{\linewidth}{LcN{99999}N{99999}N{99999}N{9.999999}N{9.999999}N{9.999999}}
\toprule
\tableheadleft{Method} & \tablehead{Conf.} & \tablehead{TP} & \tablehead{FP} & \tablehead{FN} & \tablehead{Precision $\uparrow$} & \tablehead{Recall $\uparrow$} & \tablehead{F1 $\uparrow$} \\ \midrule
\multicolumn{8}{l}{\textit{Relative frames 13--32}}\\
 & 0.15 & 5341 & 25917 & 27519 & 0.170868 & 0.162538 & 0.166599 \\
\rowcolor{tableStripe}
 & 0.25 & 5295 & 24777 & 27565 & 0.176077 & 0.161138 & 0.168277 \\
 & 0.35 & 5255 & 23966 & 27605 & 0.179836 & 0.159921 & 0.169295 \\
\rowcolor{tableStripe}
\multirow{-4}{*}{Solaris} & 0.50 & 5135 & 22478 & 27725 & 0.185963 & 0.156269 & 0.169828 \\
\addlinespace
 & 0.15 & 1859 & 25219 & 31001 & 0.068654 & 0.056573 & 0.062031 \\
\rowcolor{tableStripe}
 & 0.25 & 1724 & 21090 & 31136 & 0.075568 & 0.052465 & 0.061932 \\
 & 0.35 & 1607 & 18486 & 31253 & 0.079978 & 0.048904 & 0.060695 \\
\rowcolor{tableStripe}
\multirow{-4}{*}{Gamma-World} & 0.50 & 1439 & 15354 & 31421 & 0.085690 & 0.043792 & 0.057962 \\
\addlinespace
 & 0.15 & 6354 & 4918 & 26506 & \secondscore{0.563698} & \secondscore{0.193366} & \secondscore{0.287954} \\
\rowcolor{tableStripe}
 & 0.25 & 6079 & 3842 & 26781 & \secondscore{0.612741} & \secondscore{0.184997} & \secondscore{0.284192} \\
 & 0.35 & 5815 & 3145 & 27045 & \secondscore{0.648996} & \secondscore{0.176963} & \secondscore{0.278097} \\
\rowcolor{tableStripe}
\multirow{-4}{*}{\makecell[l]{Khora w/o\\Cross-Agent State}} & 0.50 & 5346 & 2376 & 27514 & \secondscore{0.692308} & \secondscore{0.162690} & \secondscore{0.263467} \\
\addlinespace
 & 0.15 & 29261 & 2802 & 3599 & \bestscore{0.912610} & \bestscore{0.890475} & \bestscore{0.901406} \\
\rowcolor{tableStripe}
 & 0.25 & 28657 & 1879 & 4203 & \bestscore{0.938466} & \bestscore{0.872094} & \bestscore{0.904063} \\
 & 0.35 & 28102 & 1315 & 4758 & \bestscore{0.955298} & \bestscore{0.855204} & \bestscore{0.902484} \\
\rowcolor{tableStripe}
\multirow{-4}{*}{\textbf{Khora}} & 0.50 & 26857 & 748 & 6003 & \bestscore{0.972903} & \bestscore{0.817316} & \bestscore{0.888349} \\
\bottomrule
\end{tabularx}
\end{table}

\begin{table}[!tp]
\centering
\ContinuedFloat
\caption[]{\textbf{Continued.} Detection-threshold sensitivity, frames 38--57.}
\begin{tabularx}{\linewidth}{LcN{99999}N{99999}N{99999}N{9.999999}N{9.999999}N{9.999999}}
\toprule
\tableheadleft{Method} & \tablehead{Conf.} & \tablehead{TP} & \tablehead{FP} & \tablehead{FN} & \tablehead{Precision $\uparrow$} & \tablehead{Recall $\uparrow$} & \tablehead{F1 $\uparrow$} \\ \midrule
\multicolumn{8}{l}{\textit{Relative frames 38--57}}\\
 & 0.15 & 547 & 21626 & 32313 & 0.024670 & 0.016646 & 0.019879 \\
\rowcolor{tableStripe}
 & 0.25 & 495 & 17461 & 32365 & 0.027567 & 0.015064 & 0.019482 \\
 & 0.35 & 456 & 14978 & 32404 & 0.029545 & 0.013877 & 0.018884 \\
\rowcolor{tableStripe}
\multirow{-4}{*}{Gamma-World} & 0.50 & 411 & 12230 & 32449 & 0.032513 & 0.012508 & 0.018066 \\
\addlinespace
 & 0.15 & 14459 & 16962 & 18401 & \secondscore{0.460170} & \secondscore{0.440018} & \secondscore{0.449869} \\
\rowcolor{tableStripe}
 & 0.25 & 14126 & 15448 & 18734 & \secondscore{0.477649} & \secondscore{0.429884} & \secondscore{0.452510} \\
 & 0.35 & 13806 & 14449 & 19054 & \secondscore{0.488621} & \secondscore{0.420146} & \secondscore{0.451804} \\
\rowcolor{tableStripe}
\multirow{-4}{*}{MIRA} & 0.50 & 13092 & 12948 & 19768 & \secondscore{0.502765} & \secondscore{0.398418} & \secondscore{0.444550} \\
\addlinespace
 & 0.15 & 170 & 3083 & 32690 & 0.052259 & 0.005173 & 0.009415 \\
\rowcolor{tableStripe}
 & 0.25 & 153 & 2200 & 32707 & 0.065023 & 0.004656 & 0.008690 \\
 & 0.35 & 139 & 1677 & 32721 & 0.076542 & 0.004230 & 0.008017 \\
\rowcolor{tableStripe}
\multirow{-4}{*}{\makecell[l]{Khora w/o\\Cross-Agent State}} & 0.50 & 109 & 1157 & 32751 & 0.086098 & 0.003317 & 0.006388 \\
\addlinespace
 & 0.15 & 29016 & 2654 & 3844 & \bestscore{0.916198} & \bestscore{0.883019} & \bestscore{0.899303} \\
\rowcolor{tableStripe}
 & 0.25 & 28408 & 1693 & 4452 & \bestscore{0.943756} & \bestscore{0.864516} & \bestscore{0.902400} \\
 & 0.35 & 27784 & 1199 & 5076 & \bestscore{0.958631} & \bestscore{0.845526} & \bestscore{0.898533} \\
\rowcolor{tableStripe}
\multirow{-4}{*}{\textbf{Khora}} & 0.50 & 26419 & 728 & 6441 & \bestscore{0.973183} & \bestscore{0.803987} & \bestscore{0.880531} \\
\bottomrule
\end{tabularx}
\end{table}
\endgroup

%% file: sec/evidence_runtime.tex
\section{Runtime and Scalability}
\label{app:runtime}

\subsection{Compute-Only Measurements}

Figure~\ref{fig:scalability_runtime} reports compute-only scaling
measurements with populations 1, 2, 4, 8, 16, 32, 48, 64, and 80.
One GPU is assigned per agent. Each rollout step emits four frames, so
per-view FPS is $4000/L$ for step latency $L$ in ms, and distributed
throughput is $N\cdot4000/L$.
Compute-only latency sums STBoard update, projection, and view synthesis
(model sampling plus VAE decoding). Image encoding, network transport,
and client delivery are excluded.

\input{sec/evidence_runtime_means}

\subsection{Measurement Scope and Timing Ranges}
We exclude deadline-fallback worker results from
view-synthesis compute timing. Table~\ref{tab:runtime_delivery} reports
the mean and maximum fallback-worker counts per step.
Per-GPU memory is the mean of measured GPU peaks. Logical total memory is
calculated by multiplying this value by the active agent count.

Table~\ref{tab:runtime_ranges} reports observed minimum and maximum timings.
Projection accounts for 4.56--6.97\% of mean compute-step latency across
the measured populations.

\input{sec/evidence_runtime_ranges}

\begin{figure}[tp]
\centering
\includegraphics[width=\linewidth]{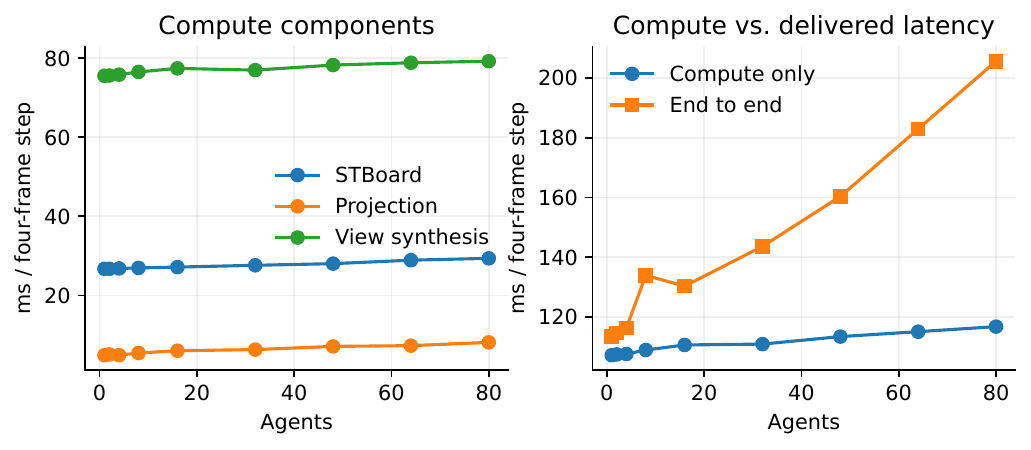}
\caption{\textbf{Measured runtime components and delivery overhead.}
Left: STBoard update, projection, and view-synthesis latency. Right:
compute-only versus end-to-end latency, including delivery in the latter.
Points show measured populations from the same experiment.}
\label{fig:runtime_detail}
\end{figure}

\subsection{End-to-End Delivery Performance}
As the population increases from 1 to 80 agents, end-to-end latency grows
from 113.5055 to 205.6232 ms;
corresponding per-view throughput decreases from 35.2406 to 19.4531 FPS.
In addition to model computation, these measurements include image
encoding, network transport, and client delivery.

\input{sec/evidence_runtime_delivery}

\subsection{Extrapolation Beyond Measured Populations}
We extrapolate compute-only latency to unmeasured populations of 128, 256,
512, and 1,024 agents using the fitted linear model
$L(N)=107.516194+0.117777N$ ms, with $R^2=0.975822$ on the measured
populations. Table~\ref{tab:runtime_projections} lists every projected
point used by the figure. Projections hold memory per GPU at the measured
80-agent value and assume one GPU per agent. Network costs are excluded.

\input{sec/evidence_runtime_projections}

%% file: sec/evidence_runtime_means.tex
\begin{table}[tp]
\centering\setlength{\tabcolsep}{3pt}
\caption{\textbf{All measured compute-only scaling points.} Times are ms per four-frame step; memory is GiB per GPU. Throughput is view-frames/s, distributed over one GPU per agent.}
\label{tab:runtime_means}
\begin{spreadtabular}{8}{cN{99.9999}N{9.9999}N{99.9999}N{999.9999}N{99.9999}N{9.9999}N{9999.9999}}
\toprule
\tablehead{Agents} & \tablehead{STBoard $\downarrow$} & \tablehead{Project $\downarrow$} & \tablehead{Synthesis $\downarrow$} & \tablehead{Total $\downarrow$} & \tablehead{FPS $\uparrow$} & \tablehead{GiB/GPU $\downarrow$} & \tablehead{View-fps $\uparrow$} \\
\midrule
1 & 26.7359 & 4.8839 & 75.5396 & 107.1594 & 37.3276 & 8.1621 & 37.3276 \\
\rowcolor{tableStripe}
2 & 26.7343 & 5.0808 & 75.6049 & 107.4200 & 37.2370 & 8.0117 & 74.4740 \\
4 & 26.8281 & 4.9044 & 75.8272 & 107.5597 & 37.1886 & 7.9365 & 148.7545 \\
\rowcolor{tableStripe}
8 & 26.9591 & 5.4382 & 76.5093 & 108.9066 & 36.7287 & 7.8989 & 293.8299 \\
16 & 27.1625 & 6.0219 & 77.4040 & 110.5884 & 36.1702 & 7.8801 & 578.7225 \\
\rowcolor{tableStripe}
32 & 27.6174 & 6.3001 & 76.9649 & 110.8824 & 36.0742 & 7.8707 & 1154.3757 \\
48 & 28.0366 & 7.1105 & 78.2479 & 113.3950 & 35.2749 & 7.8676 & 1693.1964 \\
\rowcolor{tableStripe}
64 & 28.9105 & 7.2967 & 78.8271 & 115.0343 & 34.7722 & 7.8660 & 2225.4233 \\
80 & 29.3795 & 8.1325 & 79.2210 & 116.7330 & 34.2662 & 7.8651 & 2741.2984 \\
\bottomrule
\end{spreadtabular}
\end{table}

%% file: sec/evidence_runtime_ranges.tex
\begin{table}[tp]
\centering\setlength{\tabcolsep}{4pt}
\caption{\textbf{Observed compute-only timing ranges.} Each entry is minimum--maximum in ms; these ranges are not confidence intervals and component extrema need not co-occur.}
\label{tab:runtime_ranges}
\begin{spreadtabular}{5}{cI{99.999}{99.999}I{9.999}{99.999}I{99.999}{99.999}I{999.999}{999.999}}
\toprule
\tablehead{Agents} & \tablehead{STBoard $\downarrow$} & \tablehead{Projection $\downarrow$} & \tablehead{Synthesis $\downarrow$} & \tablehead{Total $\downarrow$} \\
\midrule
1 & 26.488--27.476 & 4.310--8.740 & 75.040--77.740 & 105.838--113.956 \\
\rowcolor{tableStripe}
2 & 26.506--27.172 & 4.310--7.870 & 74.980--77.980 & 105.796--113.022 \\
4 & 26.584--27.341 & 4.370--6.610 & 75.150--79.600 & 106.104--113.551 \\
\rowcolor{tableStripe}
8 & 26.692--27.439 & 4.600--8.710 & 75.190--84.180 & 106.482--120.329 \\
16 & 26.960--27.710 & 4.590--29.390 & 75.520--92.210 & 107.070--149.310 \\
\rowcolor{tableStripe}
32 & 27.378--28.277 & 4.690--12.270 & 75.400--80.580 & 107.468--121.127 \\
48 & 27.773--29.507 & 4.930--12.100 & 75.590--91.400 & 108.293--133.007 \\
\rowcolor{tableStripe}
64 & 28.549--30.024 & 4.920--17.820 & 75.660--91.260 & 109.129--139.104 \\
80 & 28.950--30.923 & 5.080--28.840 & 75.700--96.750 & 109.730--156.513 \\
\bottomrule
\end{spreadtabular}
\end{table}

%% file: sec/evidence_runtime_delivery.tex
\begin{table}[tp]
\centering\setlength{\tabcolsep}{3pt}
\caption{\textbf{End-to-end latency and fallback counts.} Times are ms; FPS includes delivery. Fallback columns report the mean and maximum number of fallback workers per measured step.}
\label{tab:runtime_delivery}
\begin{spreadtabular}{7}{cN{99.9999}N{999.9999}N{999.9999}N{99.9999}N{9.9999}N{99}}
\toprule
\tablehead{Agents} & \tablehead{Image transport $\downarrow$} & \tablehead{Server $\downarrow$} & \tablehead{End-to-end $\downarrow$} & \tablehead{FPS $\uparrow$} & \tablehead{Fallback mean $\downarrow$} & \tablehead{Max $\downarrow$} \\
\midrule
1 & 2.4563 & 111.8901 & 113.5055 & 35.2406 & 0.0000 & 0 \\
\rowcolor{tableStripe}
2 & 3.1265 & 112.5721 & 114.6192 & 34.8982 & 0.0000 & 0 \\
4 & 4.0102 & 113.3196 & 116.2761 & 34.4009 & 0.0000 & 0 \\
\rowcolor{tableStripe}
8 & 5.7167 & 129.0565 & 133.8820 & 29.8771 & 0.4100 & 1 \\
16 & 10.3416 & 122.0269 & 130.3112 & 30.6957 & 0.0000 & 0 \\
\rowcolor{tableStripe}
32 & 19.1814 & 128.4396 & 143.6130 & 27.8526 & 0.0000 & 0 \\
48 & 29.2574 & 137.9087 & 160.3278 & 24.9489 & 0.0000 & 0 \\
\rowcolor{tableStripe}
64 & 42.8465 & 150.2562 & 182.9297 & 21.8663 & 0.0000 & 0 \\
80 & 55.4797 & 163.0364 & 205.6232 & 19.4531 & 1.2800 & 53 \\
\bottomrule
\end{spreadtabular}
\end{table}

%% file: sec/evidence_runtime_projections.tex
\begin{table}[tp]
\centering\setlength{\tabcolsep}{4pt}
\caption{\textbf{Compute-only projections, not measurements.} Values extend the measured linear latency fit. Logical memory and throughput assume one allocated GPU per agent.}
\label{tab:runtime_projections}
\begin{spreadtabular}{6}{cN{999.9999}N{99.9999}N{9.9999}N{9999.9999}N{99999.9999}}
\toprule
\tablehead{Agents} & \tablehead{Total ms $\downarrow$} & \tablehead{FPS $\uparrow$} & \tablehead{GiB/GPU $\downarrow$} & \tablehead{Logical GiB $\downarrow$} & \tablehead{View-fps $\uparrow$} \\
\midrule
128 & 122.5916 & 32.6287 & 7.8651 & 1006.7313 & 4176.4678 \\
\rowcolor{tableStripe}
256 & 137.6671 & 29.0556 & 7.8651 & 2013.4625 & 7438.2348 \\
512 & 167.8179 & 23.8354 & 7.8651 & 4026.9250 & 12203.7007 \\
\rowcolor{tableStripe}
1024 & 228.1197 & 17.5347 & 7.8651 & 8053.8500 & 17955.4856 \\
\bottomrule
\end{spreadtabular}
\end{table}

%% file: sec/evidence_qualitative.tex
\section{Additional Qualitative Results}
\label{app:visual_coverage}

\subsection{Dynamic Population Examples}
\begin{figure}[tp]
\centering
\begin{tabular}{@{}cc@{}}
\includegraphics[width=.43\linewidth]{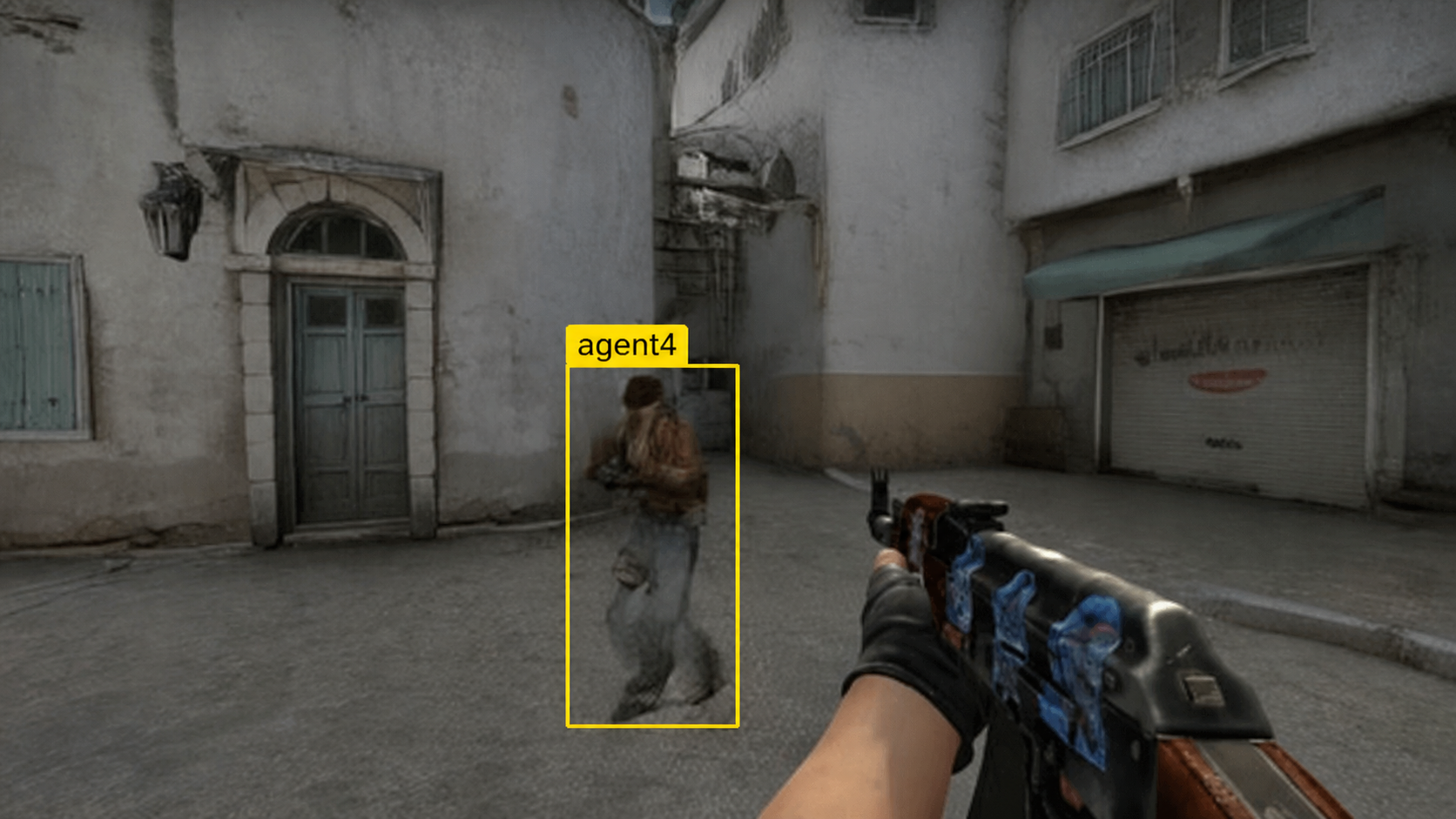} &
\includegraphics[width=.43\linewidth]{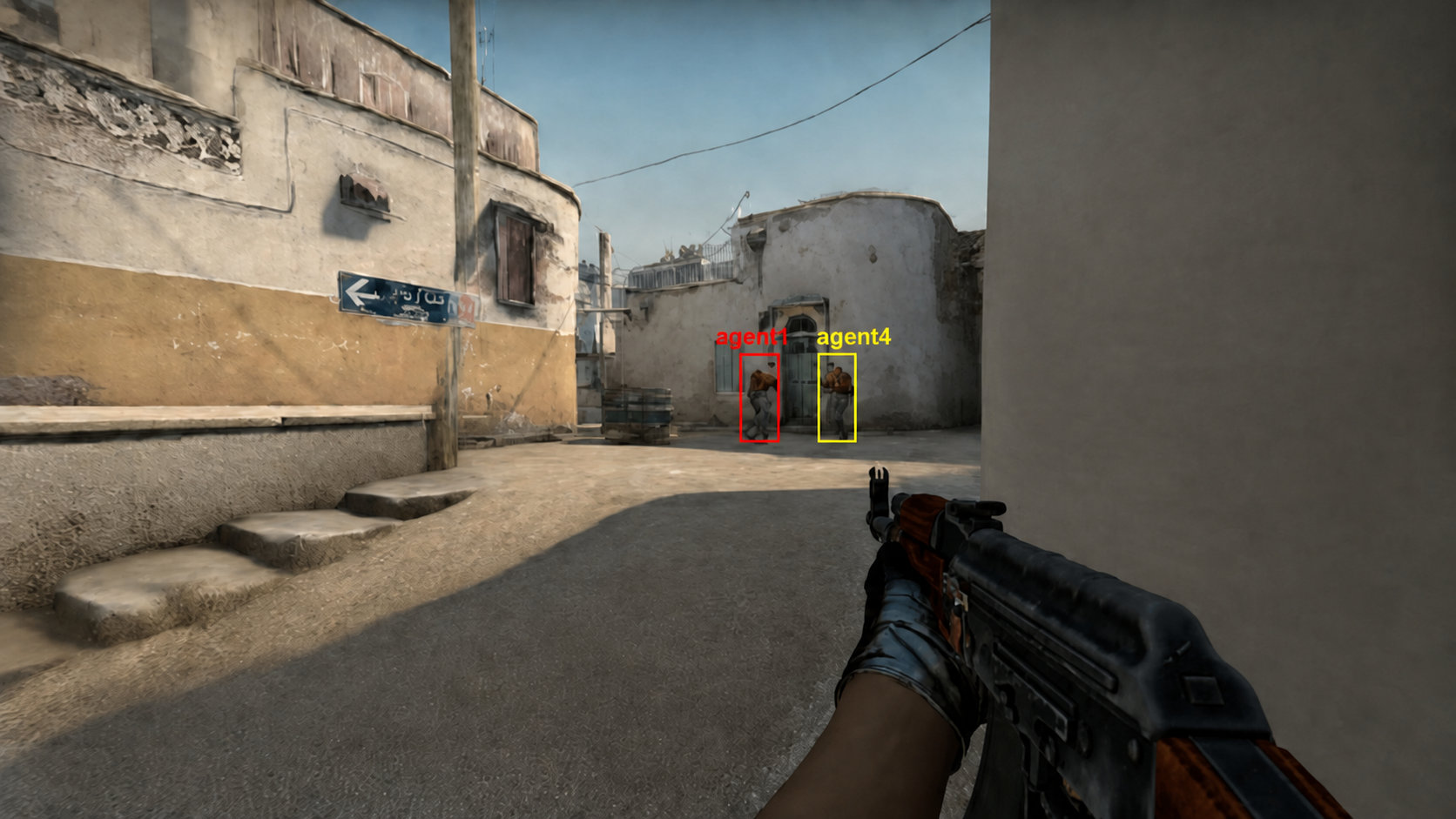}\\
Agent 1, $t_1$ & Agent 2, $t_1$\\[4pt]
\includegraphics[width=.43\linewidth]{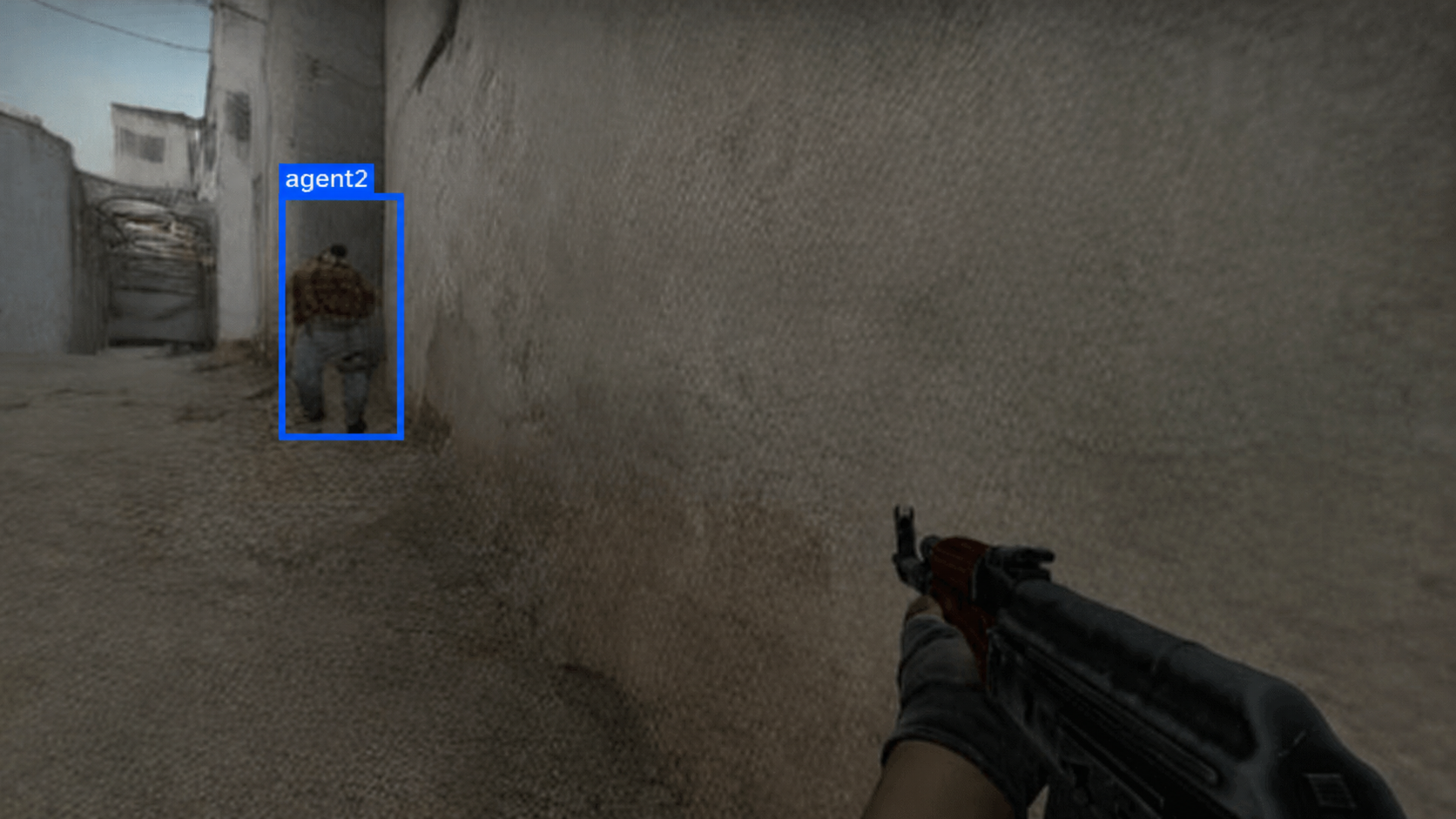} &
\includegraphics[width=.43\linewidth]{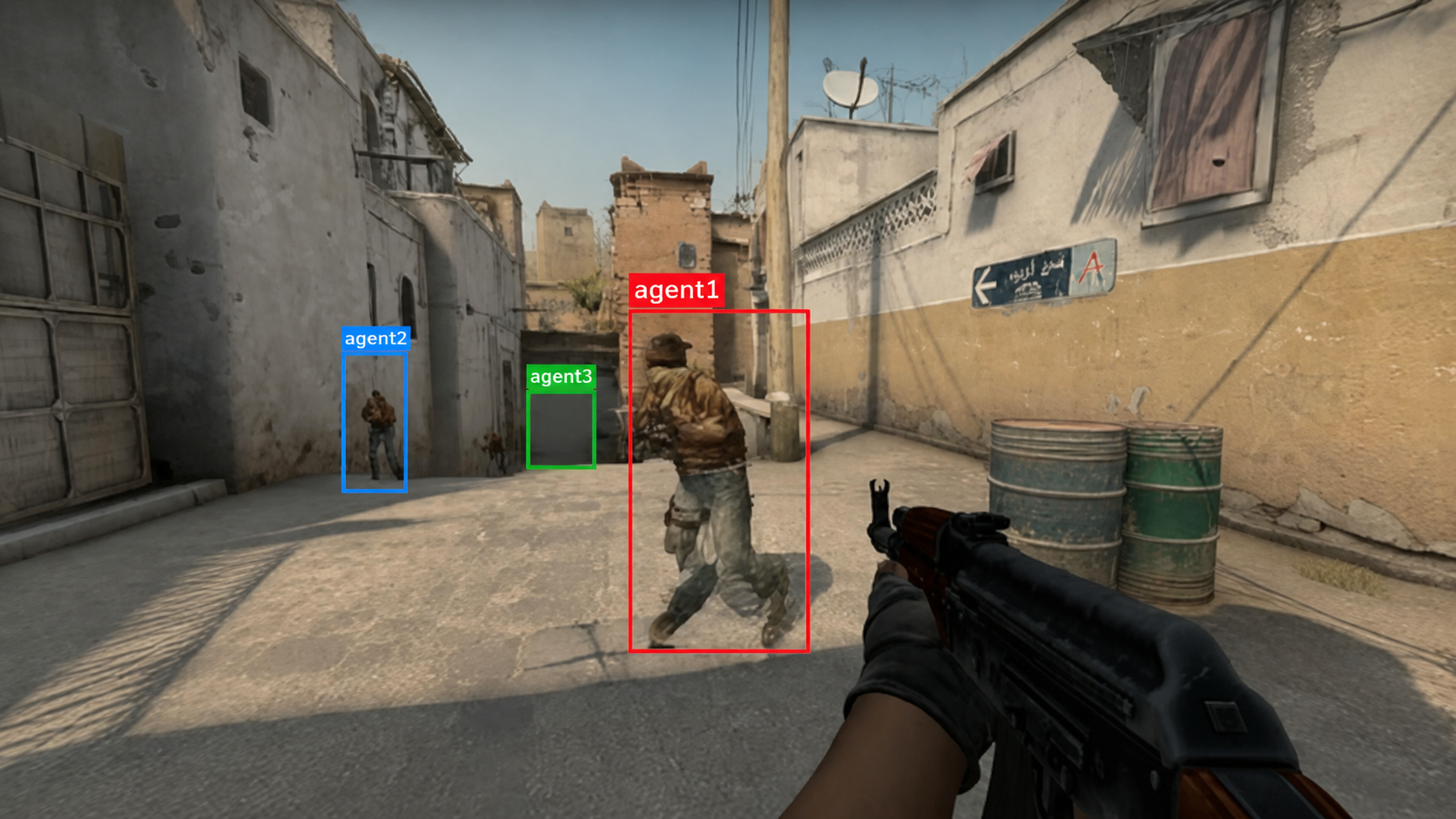}\\
Agent 3, $t_1$ & Agent 4, $t_1$
\end{tabular}
\caption{\textbf{Dynamic population: four-agent stage.}
Enlarged views of all four active agents at $t_1$ in
Figure~\ref{fig:dynamic_population}. Agent numbering and stage labels
are consistent with the main figure.}
\label{fig:dynamic_detail_join}
\end{figure}

\begin{figure}[tp]
\centering
\begin{tabular}{@{}cc@{}}
\includegraphics[width=.43\linewidth]{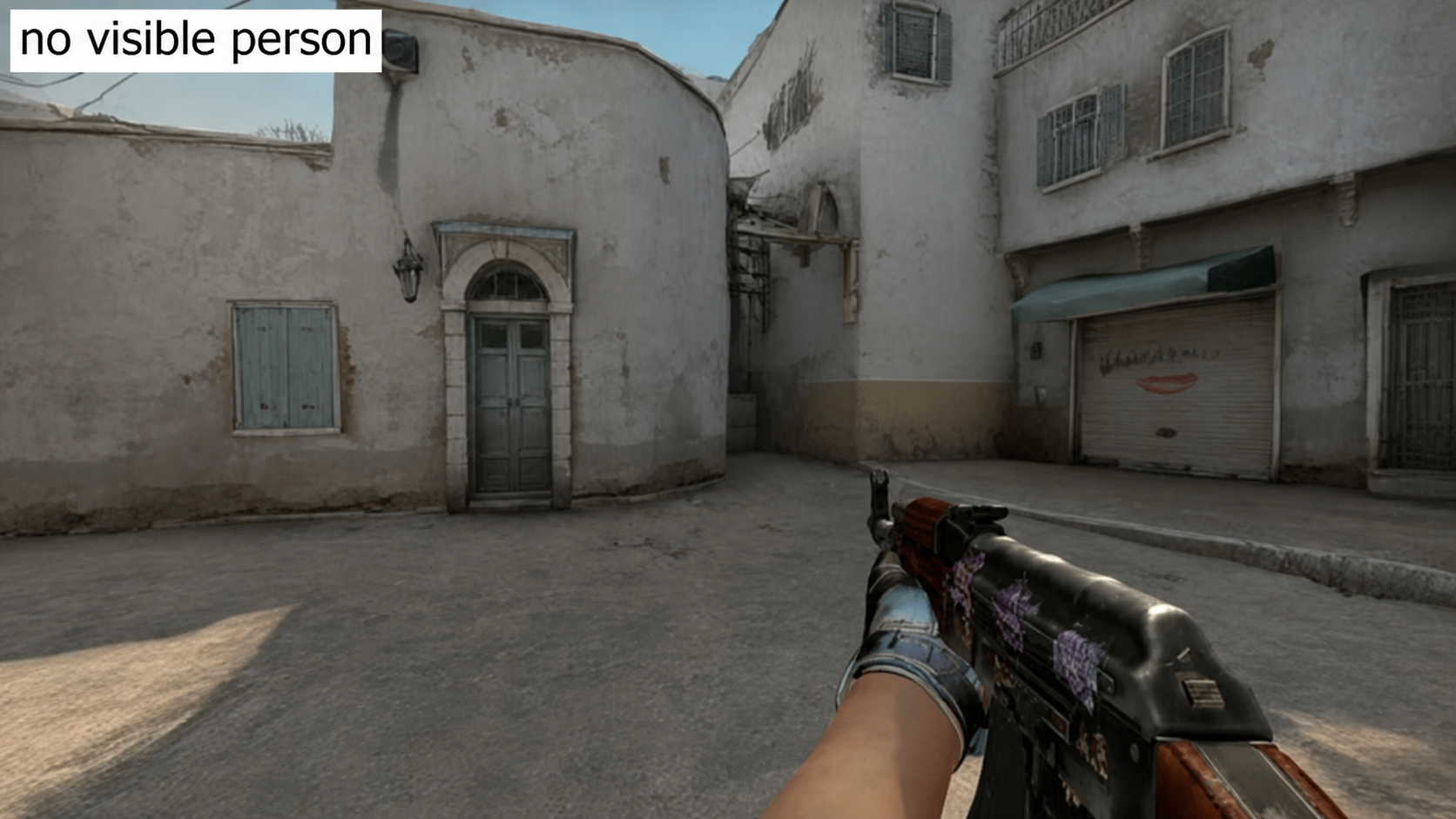} &
\includegraphics[width=.43\linewidth]{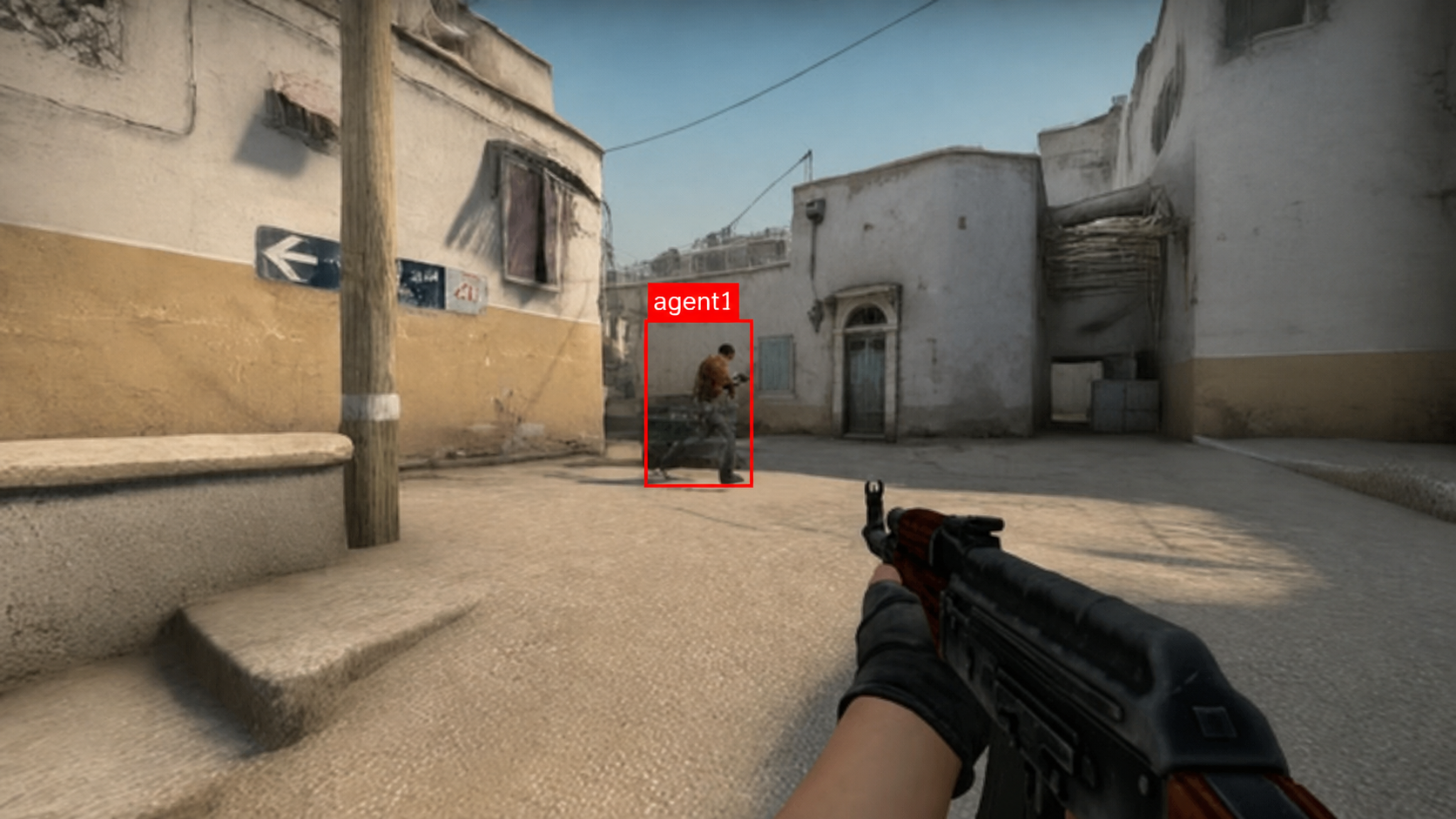}\\
Agent 1, $t_0$ & Agent 2, $t_0$\\[4pt]
\includegraphics[width=.43\linewidth]{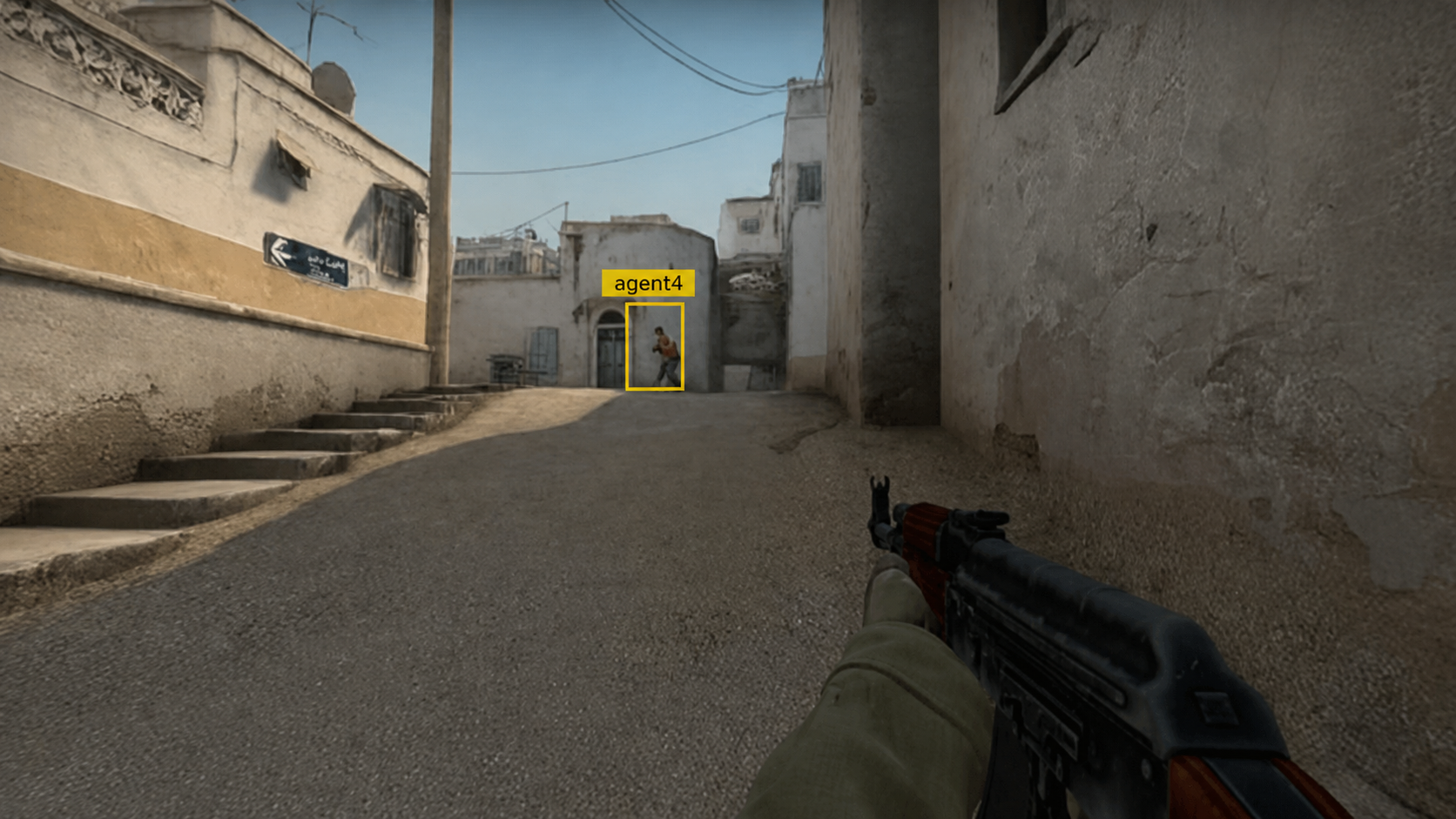} &
\includegraphics[width=.43\linewidth]{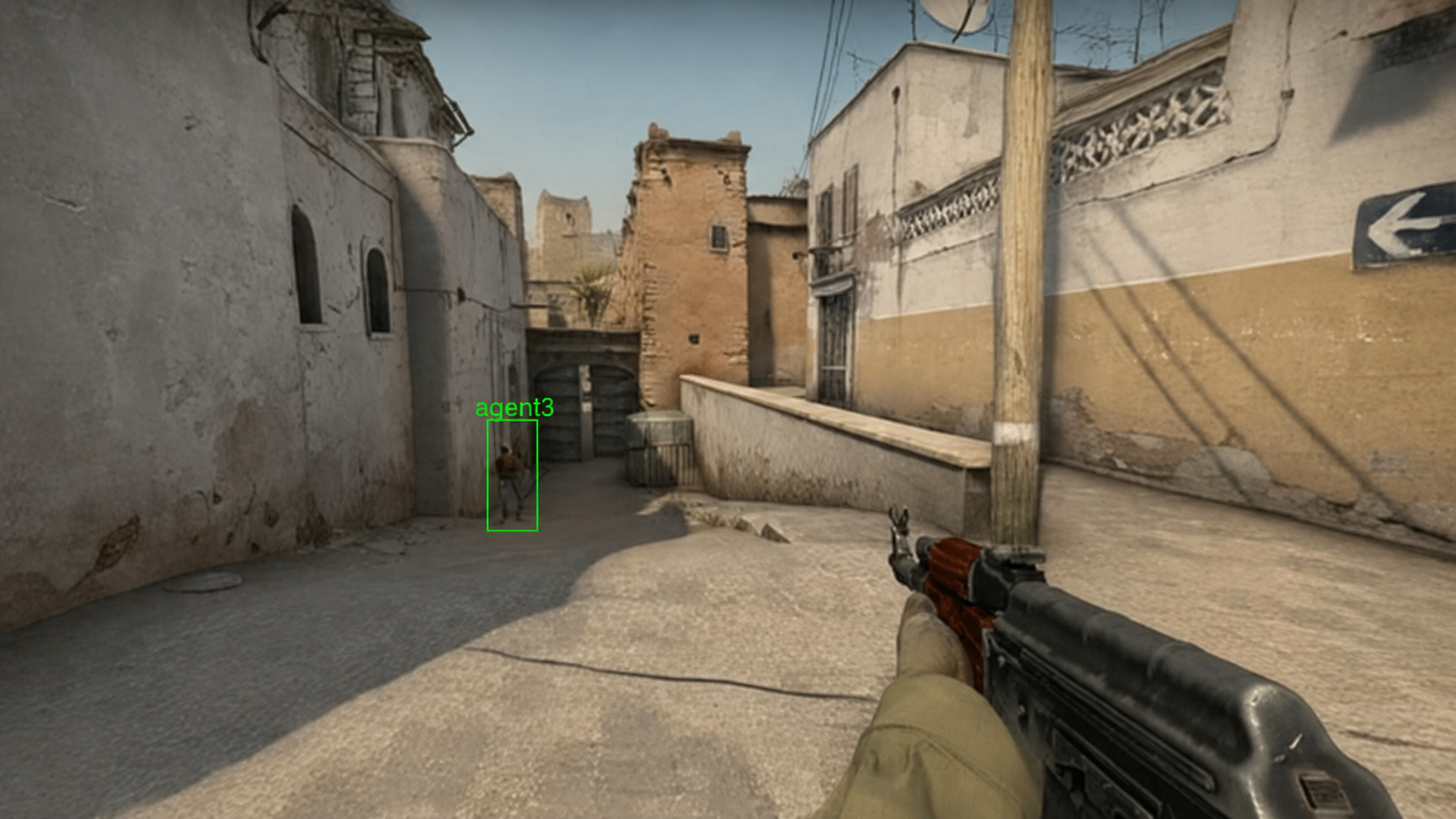}\\
Agent 3, $t_2$ & Agent 4, $t_2$
\end{tabular}
\caption{\textbf{Dynamic population: initial and final stages.}
Top: the two initially active agents. Bottom: the two remaining agents after
Agents 1 and 2 leave. Agent numbering follows
Figure~\ref{fig:dynamic_population}.}
\label{fig:dynamic_detail_endpoints}
\end{figure}

Figure~\ref{fig:dynamic_detail_join} enlarges all four active views from the
middle stage of Figure~\ref{fig:dynamic_population}.
Figure~\ref{fig:dynamic_detail_endpoints} enlarges the initial and final
two-agent stages. Together, these panels show the
$2\rightarrow4\rightarrow2$ population-change sequence at a larger scale,
complementing the main figure rather than adding independent test cases.

\subsection{Full-Frame Context for the Actor Comparison}
\begin{figure}[tp]
\centering
\includegraphics[width=\linewidth]{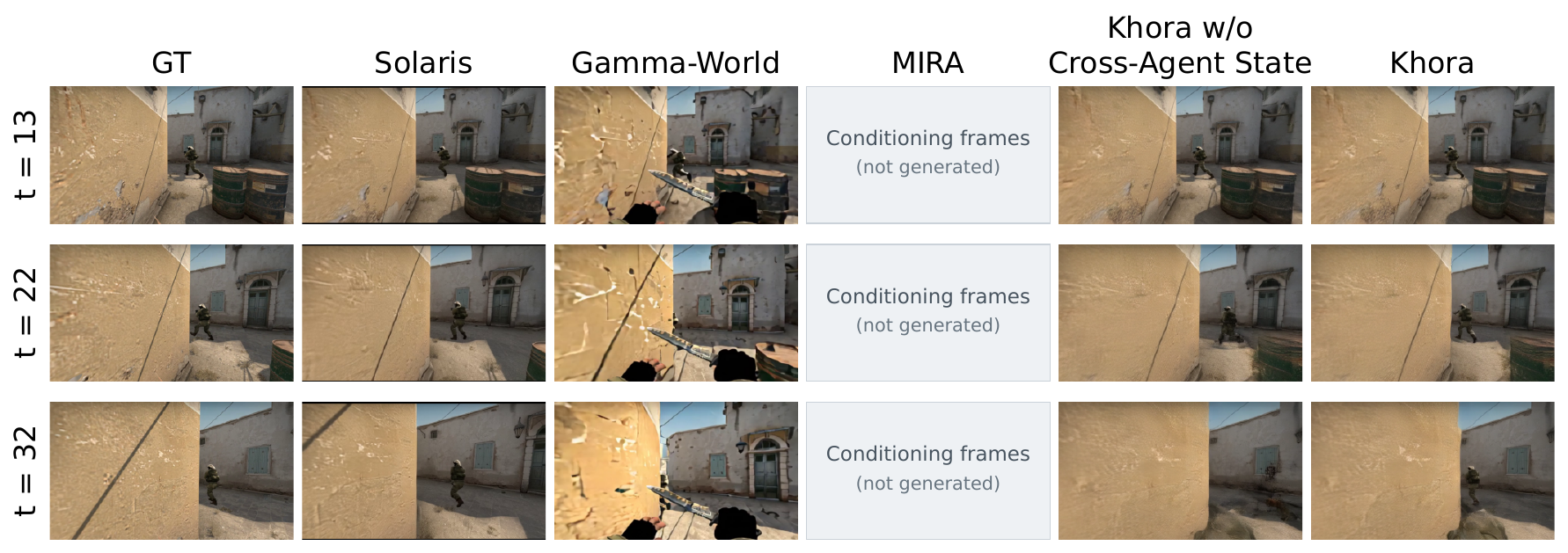}
\caption{\textbf{Full-frame generated context, early window.} Case 79,
source frames 13, 22, and 32 from top to bottom. Columns share the same
GT camera view and physical time. Images use the scoring resize/padding
protocol, with no learned enhancement, registration, or crop selection.
The six columns keep the same model order in both windows; MIRA
conditioning frames are explicitly marked rather than shown as generated results.}
\label{fig:context_early}
\end{figure}

\begin{figure}[tp]
\centering
\includegraphics[width=\linewidth]{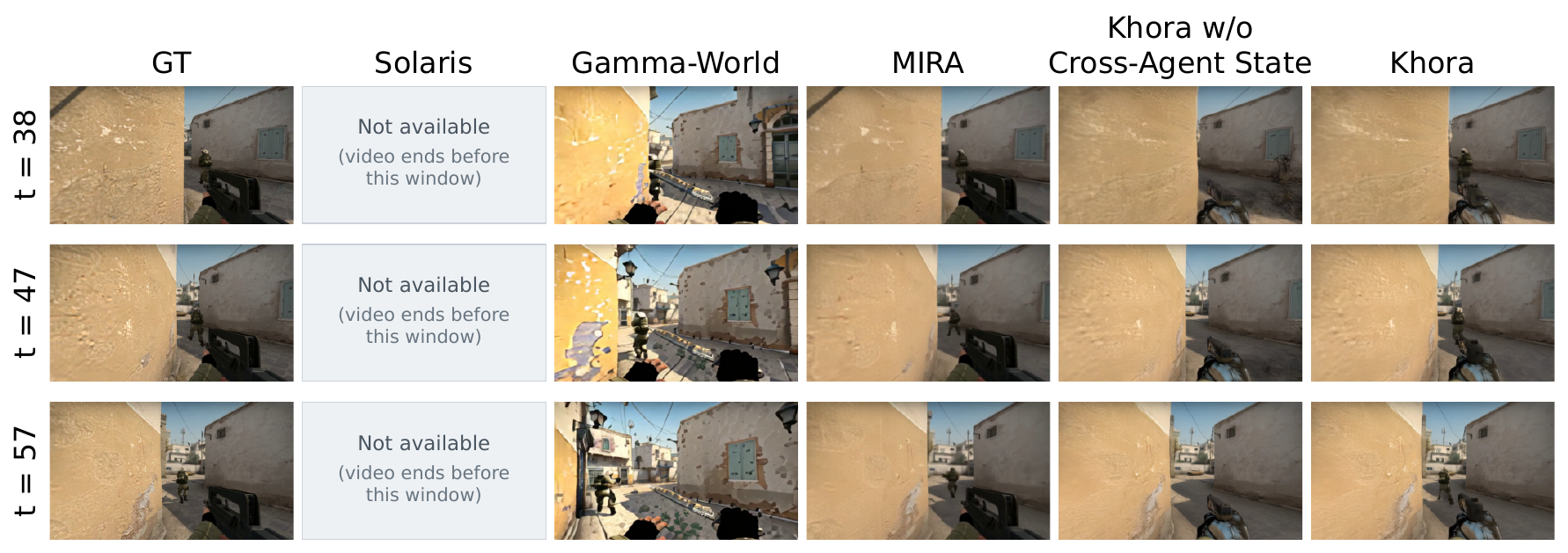}
\caption{\textbf{Full-frame generated context, late window.} The same case
and ego view at source frames 38, 47, and 57. MIRA is available in this
window; Solaris is not. No single purely generated window exists across all
five models, so early and late examples are shown separately. The Solaris
column remains visible and is marked unavailable.}
\label{fig:context_late}
\end{figure}

Figures~\ref{fig:context_early} and~\ref{fig:context_late} show the complete
scoring images for case 79 from the sampled inspection set. The early panel
uses source frames 13, 22, and 32; the late
panel uses 38, 47, and 57. The same ego view and physical frames are used
within each panel, including the Solaris one-frame offset. These examples
complement the enlarged actor crops rather than replacing full-cohort
metrics. They are not selected to maximize any model's score.

These full-frame panels show the surrounding scene, first-person body parts,
and any missing or misplaced players. Such context is useful when assessing
whether a high scene-level score reflects the target character. Quantitative
world and action scores remain distinct because their reference objects,
regions, and temporal windows differ.

